\documentclass{article}
\usepackage[utf8]{inputenc}
\usepackage[letterpaper, margin=3cm]{geometry}
\usepackage{alltt,amsmath}
\usepackage{dingbat}
\usepackage{multirow}
\usepackage{url}
\usepackage{float}
\usepackage{graphicx}
\usepackage[table,xcdraw]{xcolor}
\usepackage{caption}
\usepackage{tabularx}
\usepackage{subcaption}
\usepackage{multicol}
\usepackage{pgfplots}
\usepackage[many]{tcolorbox}
\usepackage{amssymb}
\usepackage[round,sort]{natbib}
\usepackage[UKenglish]{babel}
\usepackage{hhline}
\usepackage{bigstrut}
\usepackage{array}
\usepackage{natbib}
\usepackage{graphicx}
\usepackage{color,soul}
\usepackage{pifont}% http://ctan.org/pkg/pifont
\newcommand{\cmark}{\ding{51}}%
\newcommand{\xmark}{\ding{55}}%

\DeclareGraphicsExtensions{.pdf,.png,.jpg}

\title{A Technique Based on Trade-off Maps to Visualise and Analyse Relationships Between Objectives in Optimisation Problems}
\author{
  Rodrigo Lankaites Pinheiro \\
  ASAP Research Group\\School of Computer Science\\The University of Nottingham, UK\\
  \texttt{rodrigo.lankaitespinheiro@nottingham.ac.uk}
  \and
  Dario Landa-Silva\\
  ASAP Research Group\\School of Computer Science\\The University of Nottingham, UK\\
  \texttt{dario.landasilva@nottingham.ac.uk}
  \and
  Jason Atkin\\
  ASAP Research Group\\School of Computer Science\\The University of Nottingham, UK\\
  \texttt{jason.atkin@nottingham.ac.uk}
}
\date{}

\begin{document}

\maketitle

\begin{abstract}
Understanding the relationships between objectives in a multiobjective optimisation problem is important for developing tailored and efficient solving techniques. In particular, when tackling combinatorial optimisation problems with many objectives that arise in real-world logistic scenarios, better support for the decision maker can be achieved through better understanding of the often complex fitness landscape. This paper makes a contribution in this direction by presenting a technique that allows a visualisation and analysis of the local and global relationships between objectives in optimisation problems with many objectives. The proposed technique uses four steps: first the global pairwise relationships are analysed using the Kendall correlation method; then the ranges of the values found on the given Pareto front are estimated and assessed; next these ranges are used to plot a map using Gray code, similar to Karnaugh maps, that has the ability to highlight the trade-offs between multiple objectives; and finally local relationships are identified using scatter-plots. Experiments are presented for three different combinatorial optimisation problems: multiobjective multidimensional knapsack problem, multiobjective nurse scheduling problem and multiobjective vehicle routing problem with time windows. Results show that the proposed technique helps in the gaining of insights into the problem difficulty arising  from the relationships between objectives. 
\end{abstract}

\textbf{Keywords:} multiobjective fitness landscape analysis; trade-off region maps; fitness landscape visualisation; multiobjective combinatorial problems 

\section{Introduction}
\label{sIntro}

The development of solution techniques for multiobjective optimisation problems (MOPs) has witnessed many improvements in recent years, partly prompted by real-world applications, but also due to the increasing computational power available; affordable computers are now much more capable of performing the complex computations which are required to solve such problems. Within this field, the study of problems with many-objectives (more than three) calls for special attention: when the number of objectives grows, the number of trade-off solutions may grow exponentially, hence designing algorithms to effectively solve these problems can be very challenging \citep{Fleming2005,Purshouse2007}. One option to tackle this problem is found in the single-objective problem literature: the design of algorithms that include domain knowledge and are tailored to the specific problem. This approach requires a deeper understanding of the optimisation problem under consideration. In the presence of multiple objectives, it is important to understand the relationships between objectives and the fitness landscape of a given problem. \citet{Knowles2002} discussed this in the context of the multiobjective quadratic assignment problem, and \citet{Castro-Gutierrez2011} studied the relationships between objectives in the multiobjective vehicle routing problem with time windows.

The multiobjective optimisation literature often focuses on problems in which the multiple objectives exhibit strongly conflicting natures, as these present an increased challenge for many algorithms, hence being a motivation for applying multiobjective techniques. However, when the number of objectives grows, conflicts could be presented locally rather than globally. A \emph{global relationship} holds throughout the majority of the search space and is often easy to identify. A \emph{local relationship} exists in a restricted region of the search space and can be difficult to spot. Understanding local relationships is a useful tool to design tailored algorithms \citep{Garret2008b}. Additionally, with an increased number of objectives, composite relationships may emerge, i.e. relationships between three or more objectives that occur locally.

It is difficult to assess the multiobjective nature of MOPs and identifying trade-offs is very important for this assessment. However, the existence of multiple objectives may not guarantee the occurrence of trade-offs. Additionally, when tackling multiobjective scenarios, information about trade-offs and relationships between objectives can impact on the choice of solution techniques and algorithmic design -- objectives can be grouped or removed, trade-offs can be systematically explored, or the search could be biased towards regions of interest. Additionally, most common benchmark MOPs found in the literature are generated to emulate some sort of behaviour (usually conflicting), hence the fitness landscape is often known beforehand. Examples of this are the ZTD family of functions \citep{Zitzler2000} and DTLZ test suite \citep{Deb2002}. However, for real-world optimisation problems, especially combinatorial ones, the fitness landscape is usually unknown and learning its nature may potentially give designers substantial advantages for developing effective specialised algorithms.

\citet{Purshouse2003a} discussed some common techniques for analysing and visualising relationships between objectives in MOPs, including scatter plots and parallel coordinates (graphical representations), Kendall correlation \citep{kendall1938measure} (a quantitative metric), amongst others. \citet{Khabzaoui2004} presented statistical measures. Previous work has also applied some of these techniques to improve the understanding of relationships in specific problems, such as \citet{Castro-Gutierrez2011} who used the Kendall correlation metric to assess the multiobjective vehicle routing problem with time windows and \citet{Ishibuchi2011} who assessed many-objective problems with correlated objectives. However, these techniques are best suited for problems with two objectives and fail to discover insights into problems with a larger number of objectives.

In \citep{Pinheiro2015}, a technique to analyse and visualise relationships (both local and global) between objectives in multiobjective problems was introduced. This technique aims to improve the understanding of the fitness landscape of MOPs by assessing different perspectives of the problem objectives. In the present work, the technique is revisited and is both extended and further investigated, applying it to new problems. In order to be applied, the technique requires an approximation set of non-dominated solutions to be supplied by some means. A novel visualisation tool based on Karnaugh maps \citep{k:map:1953} is proposed to visualise the relationships between the many objectives. The analysis technique is performed in four steps: first the global pairwise relationships are analysed using the Kendall correlation method; then the ranges of the values found on the given Pareto front are estimated and assessed; next these ranges are used to plot a map using Gray code, similar to Karnaugh maps, that have the ability to highlight the trade-offs between multiple objectives; and finally local relationships are identified using scatter-plots.

% Rodrigo: I changed sets to instances. Did you mean sets of instances?
% Jason: I meant set of instances. We have different sets of instances for each problem. Each set have its own characteristics.
It was suggested in \citep{Pinheiro2015} that the technique could be used to compare the multiobjective natures of different problems and that the information obtained could be used in the design of algorithms. The analysis is extended in this work to two different problems. The technique is first applied to different instances of a multiobjective multidimensional knapsack problem to demonstrate how it can be useful to highlight different multiobjective natures of variations of the same problem. Next, a new set of multiobjective nurse scheduling problem instances is designed, based on the NSPLib \citep{maenhout2005} and the technique is applied to show that it can be used to assess the multiobjective aspects of previously unseen instances. The technique is then applied to an existing set of instances of the multiobjective vehicle routing problem with time windows \citep{Castro-Gutierrez2011}, to demonstrate that the technique can help to expand the understanding of that existing dataset. Finally, different problems are compared, to draw conclusions regarding their common features.

In summary, the contributions of this work are threefold:
\begin{enumerate}
\item The usefulness of the analysis technique presented in \citep{Pinheiro2015} is improved by reducing its reliability on domain knowledge. One weakness of the previous work was the need to define a threshold for the third step of the analysis, because it was exclusively based on domain knowledge or input from the decision-maker. A novel method to calculate better thresholds for the region maps is proposed here, along with a consideration of how to use this information to draw richer conclusions about the multiobjective nature of the given MOP.

\item The technique is shown to facilitate a comparison between the fitness landscapes of different problems, and the usefulness of this information for the crafting of solution algorithms is observed -- given two problems with similar fitness landscapes, an efficient solution algorithm for the first problem is more likely to also be efficient for the second one, given that it has similar exploratory capability on both problems.

\item Improved insights on the generation of multiobjective benchmark scenarios are presented. It is shown that in order to generate a MOP benchmark dataset, it may not be sufficient to parametrise constraints to produce scenarios with distinct fitness landscapes. Instead, a combination of parametrised constraints and correlated data is shown to be more effective.
\end{enumerate}

Section \ref{sRelated} surveys related work. Section \ref{sMop} provides the motivation for this work. Section \ref{sProposal} describes the proposed technique while experimental results applying the proposed technique to three distinct combinatorial problems are presented in Section \ref{sResults}. Finally, Section \ref{sConclusion} concludes the paper.

\section{Related Work}
\label{sRelated}

The topic of analysing fitness landscapes of MOPs is not new to the literature and important progress has been made in the past. Most previous work focuses on adapting single-objective techniques and methods to multiobjective scenarios. Some of the main contributions in this field are outlined here.

\citet{Huband2006} identified the need for properly evaluating MOPs to validate algorithm development. In their work, they extend the knowledge of several benchmark problems. They assess the problems regarding multiple criteria, including the need for external or medial parameters, the scalable number of parameters and objectives, the problems being dissimilar with regard to parameter domains and trade-off ranges, the knowledge of the Pareto optima, the optimal geometry, the parameter dependencies, the bias, the mappings and the modality. Although they provide a comprehensive assessment of the problems studied, the evaluation is specifically useful for the generation of benchmarking scenarios.

%Rodrigo - did you mean what I changed the following to, or did I change the meaning?
 \citet{Korhonen2008} surveyed the most common methods used in multiobjective decision-making frameworks, including multivariate statistical methods. They suggest the use of charts and graphs as visual aids in the decision-making process, including bar and line charts to visualise multiple solution vectors. In their work they also raise the importance of developing advanced techniques to further aid the decision-making process. In the same book, \citet{Lotov2008} described techniques for visualising the Pareto optimal set of many-objective problems, including scatterplots and heatmaps.

\citet{Garrett2008} and \citet{Garret2008b} adapted single-objective landscape analysis techniques such as analysis of the distribution of Pareto optima, fitness distance correlation, ruggedness, random walk and the geometry of the objective space to obtain insights useful to the tailoring of multiobjective evolutionary algorithms. They apply the proposed techniques to two-objective quadratic and generalised assignment problems. They conclude that using hybrid algorithms that explore knowledge about the landscape provides performance gains compared to more general approaches.

A graphical approach to evaluate the quality of a Pareto front based on ranks of objectives was proposed by \citet{Brownlee2012}. According to their assessment, the technique may not be suitable for large solution sets as it is based on an individual analysis of solutions. 

\citet{Verel2011a} adapted single-objective landscape analysis techniques to set-based multiobjective problems with objective correlation. Later, \citet{Verel2011b} conducted a study of the landscape of local optima in such problems. \citet{Verel2013331} proposed to carry out \textit{a priori} analysis of a problem by evaluating the problem size, its epistasis, the number of objectives and the correlation values between objectives, to suggest the best way to tackle it. They concluded that, depending on the problem features, different types of algorithms (scalar or Pareto approaches) and sizes of the solution archive should be employed.

Recently, different methods to visualise solutions, such as scatter plots, parallel coordinates and heat maps were reviewed by \citet{Walker2013}. They proposed two techniques: a data mining visualisation tool to plot a convex graph, and a new similarity measure of solutions to plot them in a two-dimensional space. Moreover, multiple visualisation techniques for many-objective approximation sets were surveyed by \citet{Tusar2015}. They also proposed a visualisation method based on orthogonal projections of a section and employed it on four-dimensional approximation sets. Other visualisation techniques include objective wheels, bar graphs and colour stacks, as explored by \citet{Anderson:Dror2001}. 

\citet{Fleming2014} proposed a technique to estimate the Pareto front of continuous optimisation problems, and then used the estimated front to obtain values for the decision variables of interesting solutions. They applied it to convex benchmark problems. The concept involves obtaining an initial solution set (using any multiobjective algorithm) and then calculating a projection matrix of the optimal Pareto set.

Finally, \citet{Castro-Gutierrez2011} assessed the well-known Solomon \citep{Solomon1987} dataset of the Multiobjective Vehicle Routing Problem with Time-Windows (MVRPTW) using the objectives pairwise correlation analysis proposed by \citet{Purshouse2003a} and concluded that those instances are not suited for multiobjective benchmarking purposes. They then proposed a new set of benchmark instances for the MVRPTW and showed that those instances exhibit an interesting multiobjective nature. These instances are further evaluated in this paper, using the new analysis technique proposed here, showing that the results are consistent with their claim and providing further insights into those datasets. 

Therefore, it can be observed that an improved understanding of fitness landscapes of MOPs can help in the development of better solution methods. Additionally, analysis and visualisation of fitness landscapes and relationships in MOPs is a topic of significant interest for researchers. The technique proposed in this paper seeks to make a contribution in this area.

\section{Objective Relationships in Multiobjective Optimisation}
\label{sMop}

This research focuses on the investigation of relationships between objectives in MOPs by analysing non-dominated approximation solution sets and their coverage of the objective space. Additionally, the concepts proposed by \citet{Purshouse2003a}, of conflicting, harmonious and independent objective relationships, are used.

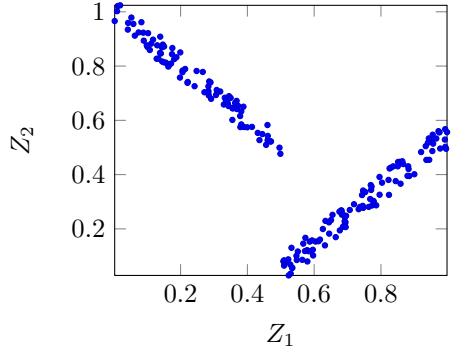
\begin{figure}[!h]
\centering
\begin{tikzpicture}
\begin{axis}[
    domain=0:1,
	width=6cm,
	xlabel={$Z_1$},
	ylabel={$Z_2$},
    enlargelimits=false,
]
\addplot+[
    only marks,
    scatter,
    mark size=1pt,
    scatter src=\thisrow{class},]
table
{figures/chart1.dat};
\end{axis}
\end{tikzpicture}
\caption[]{Example of complex relationship between two objectives $Z_1$ and $Z_2$ in a 3-objectives optimisation problem, $Z_3$ is omitted for simplicity.}
\label{notIndependent}
\end{figure}

Much of the work in the literature relies on the study of pairwise relationships between objectives \citep{Castro-Gutierrez2011,Ishibuchi2011,Guo2013}. However, results obtained solely from this technique can be misleading. Take for instance Figure \ref{notIndependent}, which shows the scatter plot of two minimisation objectives $Z_1$ and $Z_2$ for a scenario with three objectives ($Z_3$ is omitted for simplicity). 
%Rodrigo - I may be missing something here, but I do not believe that your argument is correct, even though I don't disagree with the correlation. The same would apply is they were totally correlatd- hld would be below 0.5 and half above.
%Jason - They are totally correlated, but the correlation value is zero, which by definition means independence. The point here is to show a counterexample where using correlations only can be misleading.
%I think this arises because the paper is not defining some key concepts that although may be know to MOO researchers, nevertheless should be defined: MOP problem, Pareto dominance, Pareto front, Pareto set, and conditions for conflicting, harmonious and independence according to Purshouse and Flemming. Also it is not mentioned that Z1 and Z2 are both minimisation or both maximisation, otherwise the argument does not make sense. (Dario) 
The Kendall correlation value can be calculated as follows: given an approximation solution set and any two objectives $Z_i$ and $Z_j$ where $i \neq j$, calculate the number of \textit{concordant} pairs $\mu_c$ and \textit{discordant} pairs $\mu_d$. A pair of solutions $(a,b)$ is concordant when $Z_i^a > Z_i^b$ and $Z_j^a > Z_j^b$ (where $Z_x^y$ is the value of objective $x$ for solution $y$), or when $Z_i^a < Z_i^b$ and $Z_j^a < Z_j^b$. A pair of solutions is discordant when $Z_i^a > Z_i^b$ and $Z_j^a < Z_j^b$, or when $Z_i^a < Z_i^b$ and $Z_j^a > Z_j^b$. If $\mu$ is the total number of solutions in the approximation set, then the Kendall correlation coefficient $\tau$ can be calculated as follows: 
$$\tau = \frac{\mu_c - \mu_d}{\frac{1}{2}\mu(\mu-1)}$$

%Given that the number of solutions where $Z_1 < 0.5$ is equal to the number of solutions where  \geq 0.5$, then the correlation value of $(Z_1, Z_2)$ is zero, hence one could conclude that the objectives are independent. 
Considering the global objective space, using techniques such as the Kendall correlation technique, it would be easy to conclude that the objectives $Z_1$ and $Z_2$ are independent.
%Rodrigo: JA notes that kendall correlation could have been used on only parts of the objective space, so it's not an inate flaw of the technique, but of the application of it.
%Jason: JA? Well, ypu can apply that in that fashion, although even so it would be difficult to spot the local relationships. Also, in the literature we can find some works that purely rely on global pairwise correlation values.
However, it is clear that, in fact, the objectives are conflicting when $Z_1 < 0.5$ and harmonious when $Z_1 \geq 0.5$. 
This misleading loss of information is a consequence of considering the global rather than local objective space, thus identifying only \emph{global relationships} between objectives.
%This misleading information (or loss of information) happens because the Kendall correlation technique identifies only \emph{global relationships} between objectives and does not present the ability to highlight local relationships. 

\begin{figure}[!h]
\centering
\begin{subfigure}[b]{0.48\textwidth}
\begin{equation*}\fontsize{8}{8}
\begin{aligned}
& ({\color[HTML]{FE0000}45}, {\color[HTML]{32CB00}68}, {\color[HTML]{32CB00}85})
,({\color[HTML]{FE0000}6},{\color[HTML]{32CB00}63},{\color[HTML]{32CB00}99})
,({\color[HTML]{FE0000}34},{\color[HTML]{32CB00}64},{\color[HTML]{32CB00}95}),\\
& ({\color[HTML]{FE0000}28},{\color[HTML]{32CB00}100},{\color[HTML]{FE0000}48})
,({\color[HTML]{32CB00}98},{\color[HTML]{FE0000}47},{\color[HTML]{32CB00}69})
,({\color[HTML]{FE0000}48},{\color[HTML]{32CB00}62},{\color[HTML]{32CB00}79}),\\
& ({\color[HTML]{32CB00}72},{\color[HTML]{FE0000}24},{\color[HTML]{32CB00}90}),
({\color[HTML]{32CB00}82},{\color[HTML]{32CB00}79},{\color[HTML]{FE0000}16}),({\color[HTML]{FE0000}36},{\color[HTML]{32CB00}100},{\color[HTML]{32CB00}97}),\\
& ({\color[HTML]{32CB00}100},{\color[HTML]{32CB00}87},{\color[HTML]{FE0000}41}),
({\color[HTML]{32CB00}98},{\color[HTML]{FE0000}19},{\color[HTML]{32CB00}87}),
({\color[HTML]{32CB00}85},{\color[HTML]{32CB00}57},{\color[HTML]{FE0000}50}),\\
& ({\color[HTML]{32CB00}88},{\color[HTML]{FE0000}20},{\color[HTML]{32CB00}73}),
({\color[HTML]{32CB00}91},{\color[HTML]{FE0000}48},{\color[HTML]{32CB00}99}),
({\color[HTML]{32CB00}94},{\color[HTML]{FE0000}31},{\color[HTML]{32CB00}70}),\\
& ({\color[HTML]{32CB00}56},{\color[HTML]{FE0000}49},{\color[HTML]{32CB00}59}),
({\color[HTML]{32CB00}75},{\color[HTML]{32CB00}93},{\color[HTML]{FE0000}1}),
({\color[HTML]{FE0000}38},{\color[HTML]{32CB00}84},{\color[HTML]{32CB00}85}),\\
& ({\color[HTML]{FE0000}45},{\color[HTML]{32CB00}78},{\color[HTML]{FE0000}47})
\end{aligned}
\end{equation*}
\caption{Non-dominated points.}
\label{hiddentradeoffA}
\end{subfigure}%
\hfill
\begin{subfigure}[b]{0.50\textwidth}
\centering
\begin{subfigure}[b]{0.25\textwidth}
\begin{tcolorbox}[fit,width=2.01cm,height=1.54cm,blank,
  borderline={0.4pt}{0pt},valign=center,
  nobeforeafter]
  \centering
 $Z_1$
\end{tcolorbox}
\end{subfigure}
\begin{subfigure}[b]{0.25\textwidth}
\begin{tcolorbox}[fit,width=2.01cm,height=1.54cm,blank,
  borderline={0.4pt}{0pt},valign=center,
  nobeforeafter]
  \centering
 -0.30
\end{tcolorbox}
\end{subfigure}
\begin{subfigure}[b]{0.25\textwidth}
\begin{tcolorbox}[fit,width=2.01cm,height=1.54cm,blank,
  borderline={0.4pt}{0pt},valign=center,
  nobeforeafter]
  \centering
 -0.33
\end{tcolorbox}
\end{subfigure}
\\
\begin{subfigure}[b]{0.25\textwidth}
\begin{tikzpicture}
\begin{axis}[
	ticks=none,
	width=3.59cm,
	ymin=-5,
	xmin=-5,
		xmax=105,
	ymax=105,
    enlargelimits=false,
]
\addplot[
    only marks,
    scatter,
    mark size=1pt,
    scatter src=\thisrow{class},]
table[x={A}, y={B}]
{figures/chart2.dat};
\end{axis}
\end{tikzpicture}
\end{subfigure}
\begin{subfigure}[b]{0.25\textwidth}
\begin{tcolorbox}[fit,width=2.01cm,height=1.54cm,blank,
  borderline={0.4pt}{0pt},valign=center,
  nobeforeafter]
  \centering
 $Z_2$
\end{tcolorbox}
\end{subfigure}
\begin{subfigure}[b]{0.25\textwidth}
\begin{tcolorbox}[fit,width=2.01cm,height=1.54cm,blank,
  borderline={0.4pt}{0pt},valign=center,
  nobeforeafter]
  \centering
 -0.33
\end{tcolorbox}
\end{subfigure}
\\
\begin{subfigure}[b]{0.25\textwidth}
\begin{tikzpicture}
\begin{axis}[
	ticks=none,
	width=3.59cm,
	ymin=-5,
	xmin=-5,
		xmax=105,
	ymax=105,
    enlargelimits=false,
]
\addplot[
    only marks,
    scatter,
    mark size=1pt,
    scatter src=\thisrow{class},]
table[x={A}, y={C}]
{figures/chart2.dat};
\end{axis}
\end{tikzpicture}
\end{subfigure}
\begin{subfigure}[b]{0.25\textwidth}
\begin{tikzpicture}
\begin{axis}[
	ticks=none,
	width=3.58cm,
	ymin=-5,
	xmin=-5,
	xmax=105,
	ymax=105,
    enlargelimits=false,
]
\addplot[
    only marks,
    scatter,
    mark size=1pt,
    scatter src=\thisrow{class},]
table[x={B}, y={C}]
{figures/chart2.dat};
\end{axis}
\end{tikzpicture}
\end{subfigure}
\begin{subfigure}[b]{0.25\textwidth}
\begin{tcolorbox}[fit,width=2.01cm,height=1.54cm,blank,
  borderline={0.4pt}{0pt},valign=center,
  nobeforeafter]
  \centering
 $Z_3$
\end{tcolorbox}
\end{subfigure}
\caption[]{Scatter-plot matrix and correlation coefficients.}
\label{hiddentradeoffB}
\end{subfigure}
\caption{Three-way conflicting objectives}
\label{hiddentradeoff}
\end{figure}

Additionally, common representations fail to highlight \emph{composite relationships} between more than two objectives. Figure \ref{hiddentradeoff} shows evidence of that. Figure \ref{hiddentradeoffA} shows a set of 19 non-dominated points where green values are above 50 and red values are equal to or below 50. It is clear that no point has three values simultaneously above 50. Figure  \ref{hiddentradeoffB} shows the scatter plot and the Kendall correlation coefficients for each pair of objectives. The scatter plot and correlation values do not help to identify the three-way trade-off. Likewise, it is clear that the correlation values do not indicate any strong pairwise correlation. 

To better analyse and visualise the multiobjective nature of optimisation problems, techniques are needed that will help in identifying global, local and composite relationships between objectives as well as interesting trade-offs in the fitness landscape.

In order to illustrate the analysis technique we apply it to different scenarios of the multiobjective multidimensional knapsack problem (MOMKP) \citep{Lust2010}. We aim to show that within the same problem, the proposed technique can identify multiple scenarios with distinct multiobjective natures.

In the MOMKP, we have \textit{n} items $(i=1,\dots,n)$ with \textit{m} weights $w^i_j$ $(j=1,\dots,m)$ and \textit{p} profits $c^i_k$ $(k=1,\dots,p)$. A set of items must be selected to maximise the \textit{p} profits while not exceeding the capacities $W_j$ of the knapsack. This problem can be formulated as follows:
\begin{equation*}
\begin{aligned}
& \text{maximise}
& & \sum_{i=1}^{n} c^i_k x_i  
& & & k=1,\dots,p\\
& \text{subject to}
& & \sum_{i=1}^{n} w^i_j x_i \leq W_j
& & & j=1,\dots, m \\
& & & x_i \in {0,1}
& & & i=1,\dots, n \\
\end{aligned}
\end{equation*}
\section{A Four Step Analysis and Visualisation Technique}
\label{sProposal}

The analysis technique which was originally proposed in \citep{Pinheiro2015} is now revisited, refined and enhanced. This is a four-step method to analyse and visualise objectives relationships in MOPs. 
The following requirements must be met in order to apply the technique:
\begin{enumerate}
\item Obtain an approximation to the Pareto optimal set for a subset of instances of the multiobjective problem in hand. This can be obtained by any multiobjective algorithm (\textit{MOA}), provided that the quality of the solutions is good enough. A combination of different methods is suggested here, to increase the likelihood of obtaining a good quality approximation set. Multiple MOAs are publicly available and frameworks like JMetal \citep{Durillo2011}, ParadiseEO \citep{Cahon2004} and MOEA Framework \citep{moeaf} provide efficient implementations of several state of the art algorithms that may be applied to different problem domains. 
\item  Obtain knowledge of the problem domain and the importance of different objective values. Steps two and three particularly require this, when defining meaningful ranges, quality thresholds and regions of interest. 
\end{enumerate}

The aim of the proposed technique is to aid the study of subsets of problem instances, to help the design of more effective tailored algorithms for solving other instances of the same problem. The focus of this work is to investigate the suitability of the technique for that purpose, hence it is applied to three distinct MOPs. 
%It might be argued that having to solve the problem beforehand is a weakness of the method, however, the improved knowledge may be used to design algorithms that present better performance on the same or unknown instances. 
Although the need to solve the problem first could be seen as a weakness, there are often similarities between some problem instances, especially for real-world problems, thus insights into one set of instances can give insights into many other instances.
Solving many-objective problems is computationally expensive, thus any help in tailoring fast techniques that can provide `good-enough' results is important, and this technique may allow a user to identify similarities between instances which could be exploited. 
Moreover, the increased understanding of the problem could help to identify strengths and weaknesses in algorithms.

Each step in the analysis brings to light more information regarding any relationships between objectives, the coverage of the feasible region and the trade-off landscapes. The details of the four steps in the proposed technique are presented in the remainder of this section, the results of the analysis of three distinct problems are presented in the next section: a multiobjective multidimensional knapsack problem, a multiobjective nurse scheduling problem and a multiobjective vehicle routing problem with time windows.

\subsection{Step 1 -- Global Pairwise Relationship Analysis:}

The available approximation set of non-dominated solutions is first analysed using Kendall correlation coefficients \citep{kendall1938measure}, to identify global pairwise relationships, as proposed by \cite{Purshouse2003a}. The existence of strongly conflicting correlations (values $<-0.5$) immediately indicates that a trade-off surface exists. Analogously, the existence of strongly harmonious correlations (values $>0.5$) indicates that objectives could be aggregated. Objectives identified as independent only prove that the objectives are not globally dependent, but do not imply the absence of local trade-offs in some areas of the fitness landscape.

If independence is detected, the problem could be assessed for the possibility of decomposing the decision variables according to the objectives in order to solve each objective (or groups of objectives) separately, as such an approach has been observed to improve performance \citep{Purshouse2003b}.

\subsection{Step 2 -- Objective Range Analysis:}

Next, the range (the difference between best and worst values observed) of each objective in the approximation set is calculated. The problem domain knowledge or expertise of the decision-maker is then used to assess how to approach each objective; e.g. assign it to a cluster, remove it, etc.
This approach assumes that there is some way to categorise solutions into high vs low quality, or acceptable vs unacceptable to the decision-maker. Hence, the availability of a domain-expert or sufficient domain knowledge to determine this, is important.

%Rodrigo : this had a + at the start of paragraph, was the + deliberate? It looked odd so I thought it's probably a typo and removed it but feel free to add it back in if it was deliberate.
A \textit{meaningful objective} is an objective in which the range is large enough such that solutions can be classified into multiple quality categories (good to bad, high to low, acceptable to unacceptable, etc) depending on the problem domain. The existence of this type of objective indicates the presence of a trade-off in the fitness landscape. Analogously, a \textit{non-meaningful objective} is an objective in which the range is small to the point that its variability can be considered negligible according to the problem domain, thus it is not worth exploring further. 
Again, the categorisation of solution quality is important, to avoid issues of scaling factors, for example. An objective with a large range of values across the non-dominated solutions, but where all such values are considered by decision-makers to be good (if such is possible), would be a non-meaningful objective. Conversely, even a binary objective where it is important for the objective to have the value of 1 would be meaningful if some of the non-dominated solutions had a value of 0 for the objective.
%Rodrigo - feel free to trim this part or change it - I wanted to get something in here for you. J.

One way to tackle non-meaningful objectives is to ignore them. With a low enough variability across the approximation set, it is possible that optimising for the other objectives will still result in acceptable values for the non-meaningful objectives.
This may be the result of composite relationships or of the characteristics of the data in the scenario. 
Note that having a small range across the initial solution set does not imply that the range is negligible for the entire objective space. To assess whether this is the case, remove this objective from the MOA, determine a new approximation set, and calculate the ranges of the excluded objective in the new solution set. If the new range is also negligible, then it is likely that the objective can be ignored. 
% Rodrigo - I added a 'likely' since you only obtain an approximation set so it is possible that the optimal values that you miss have poor values for this objective, but are hard to find for the algorithm that you used.
% I also just removed the recommend. In what basis is the recommendation being made? Is there evidence?

Another option to tackle non-meaningful objectives is to cluster them \citep{Guo2013}. 
Clustering the objectives will reduce the size of the objective space, but the MOA will still keep some pressure towards optimising these objectives. Arguably the overall results for these objectives may not be as good as solving them separately. However, considering that they all presented small ranges, it is possible that the new solutions will present similar ranges as before.
%Rodrigo - if you cluster, e.g. single objective as weighted sum of the others maybe? then surely pushing that sum to its extreme value will still give as good a value? Presumably the issue is that the weight may mean that a specific one may not take its optimal value?
%Jason: That is correct, you may not get the optimal value (although usually in many-objectives literature, we aim for 'good-enough' due to difficulties to explore the objective space), however if the range is small the impact of clustering may not be too bad and by grouping objectives you will be eliminating a lot of complexity and reducing the objective space considerably, which could make your algorithms perform better overall.
% Rodrigo - my comment is about "the overall results for these objectives may not be as good as solving them separately". I am not convinced that this is true for the objectives as a whole apart from for the heuristic nature of the process, but on those grounds one could equally argue that, "separating the objectives again, the overall results for these objectives may not be as good as solving them together". You can make the argument for a single objective, because you may have chosen a bad weight for it, but how can you make the argument for all of them?
%Jason: That would be true ("separating the objectives..."), especially if the MOA has the ability to reach the optimal. Thing is on many-objectives problems that is rarely the case. In case of clustering and assigning a bad weight, them impact of this wouldn't be too heavy because the objectives in the cluster have low ranges

\subsection{Step 3 -- Trade-off Regions Analysis:}

The third step consists of a quantitative method based on Karnaugh maps \citep{k:map:1953} to classify the objective space into regions of interest to aid in identifying trade-offs and complex relationships between objectives. A Karnaugh map is a method to visualise and simplify boolean algebra expressions using a truth table. A map for $i$ variables has $2^i$ cells. The cells' labelling follows the Gray code, therefore any two adjacent cells differ in only one bit. For example, in a three variables scenario, the cells adjacent to cell 0 ($000_2$) are cells 1($001_2$), 2($010_2$) and 4($100_2$). Karnaugh maps are used to group variables because they make it easy to visualise patterns.

Given an approximation set, to build its region map it is necessary to first define, for each objective $Z_i$ where $(i=1,2,\dots m)$, a threshold  $t_i$ such that values above (assuming a maximisation problem) $t_i$ are considered good or acceptable, and values below $t_i$ are considered bad or inadequate. 
The threshold can be obtained using domain knowledge or empirically -- for example, using the average value for each objective as its threshold.
Since the threshold value may be important for problems, it can be worth considering alternative values for the threshold, as discussed later.

Next, each objective value in each solution is classified as good (\cmark) or bad (\xmark) according to the objective's threshold.
For example, a solution $(Z_1=\text{\cmark},Z_2=\text{\xmark},Z_3=\text{\cmark})$ in a three-objective scenario indicates that for $Z_1$ and $Z_3$ the objective values are better than their respective thresholds, while $Z_2$ presents a value which is worse than its threshold.

A \textit{region map} is then drawn. This is similar to a Karnaugh map, but instead of boolean variables, the objectives of the MOP are used. These are arranged such that the cells of the map follow the Gray code (replacing 0's and 1's by \cmark and \xmark). The map is built with $2^m$ regions where each region represents a single combination of good and/or bad objectives. Each cell in the map represents a region $r_k$ using a binary encoding such that the least significant digit represents objective $Z_1$ and the most significant digit is objective $Z_m$. Finally, the region number is identified for each solution (considering $\text{\cmark}=0$ and $\text{\xmark}=1$) and the number of solutions in each region is determined. The solution of the previous example, $(Z_1=\text{\cmark},Z_2=\text{\xmark},Z_1=\text{\cmark})$, falls into region $r_{2}$ because $2_{10}= 010_2$. Region $r_0$ represents solutions with good values in all objectives

\begin{figure}[h]
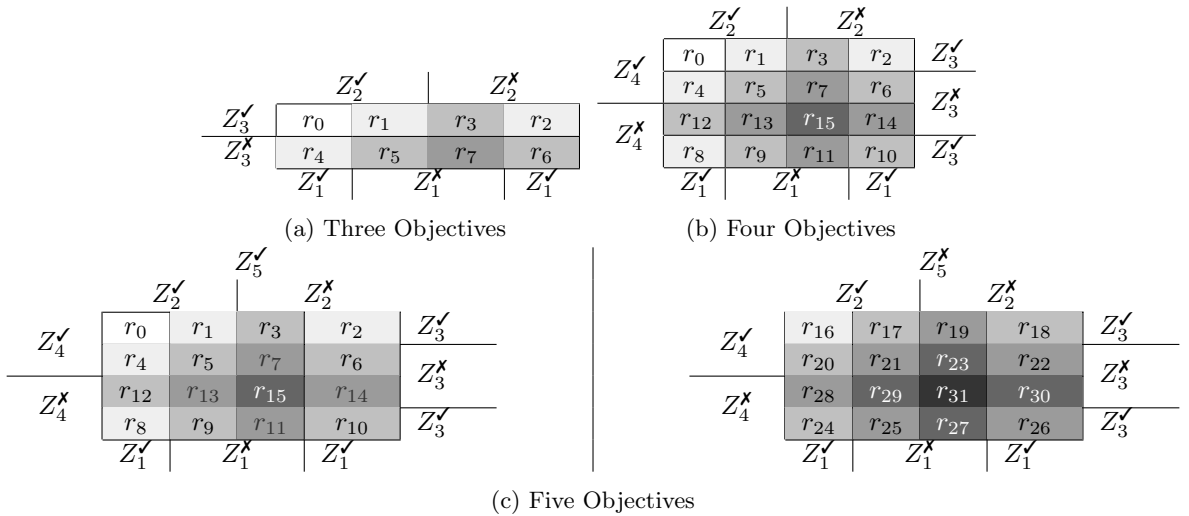

\centering
\begin{subfigure}[b]{0.33\textwidth}
\resizebox{\textwidth}{!}{%
\begin{tabularx}{\textwidth}{XX|X|X|X}
 & \multicolumn{2}{c|}{$Z_2^\text{\cmark}$} & \multicolumn{2}{c}{$Z_2^\text{\xmark}$} \\ 
 \hhline{~----}
\multicolumn{1}{c|}{$Z_3^\text{\cmark}$} & \centering $r_0$ & {\centering \cellcolor[HTML]{EFEFEF}$r_1$} & \centering \cellcolor[HTML]{C0C0C0}$r_3$ & \multicolumn{1}{c|}{\cellcolor[HTML]{EFEFEF}$r_2$} \\ \hhline{-----}
\multicolumn{1}{c|}{$Z_3^\text{\xmark}$} & \centering \cellcolor[HTML]{EFEFEF}$r_4$ & \centering \cellcolor[HTML]{C0C0C0}$r_5$ & \centering \cellcolor[HTML]{9B9B9B}$r_7$ & \multicolumn{1}{c|}{\cellcolor[HTML]{C0C0C0}$r_6$} \\ \cline{2-5} 
 & \centering $Z_1^\text{\cmark}$ & \multicolumn{2}{c|}{\centering $Z_1^\text{\xmark}$} & \centering $Z_1^\text{\cmark}$
\end{tabularx}
}
\caption{Three Objectives}
\label{rma}
\end{subfigure}
\begin{subfigure}[b]{0.33\textwidth}
\resizebox{\textwidth}{!}{%
\begin{tabularx}{\textwidth}{XX|X|X|XX}
 & \multicolumn{2}{c|}{\textbf{$Z_2^\text{\cmark}$}} & \multicolumn{2}{c}{\textbf{$Z_2^\text{\xmark}$}} &  \\ \hhline{~----}
\multicolumn{1}{c|}{} &  \centering $r_0$ &  \centering \cellcolor[HTML]{EFEFEF}$r_1$ &  \centering \cellcolor[HTML]{C0C0C0}$r_3$ & \multicolumn{1}{c|}{ \centering \cellcolor[HTML]{EFEFEF}$r_2$} & \textbf{$Z_3^\text{\cmark}$} \\ \hhline{~-----} 
\multicolumn{1}{c|}{\multirow{-2}{*}{\textbf{$Z_4^\text{\cmark}$}}} &  \centering \cellcolor[HTML]{EFEFEF}$r_4$ &  \centering \cellcolor[HTML]{C0C0C0}$r_5$ &  \centering \cellcolor[HTML]{9B9B9B}$r_7$ & \multicolumn{1}{c|}{ \centering \cellcolor[HTML]{C0C0C0}$r_6$} &  \\ \hhline{-----}
\multicolumn{1}{c|}{} &  \centering \cellcolor[HTML]{C0C0C0}$r_{12}$ &  \centering \cellcolor[HTML]{9B9B9B}$r_{13}$ &  \centering {\cellcolor[HTML]{656565}\textcolor{white}{$r_{15}$}} & \multicolumn{1}{c|}{ \centering \cellcolor[HTML]{9B9B9B}$r_{14}$} & \multirow{-2}{*}{\textbf{$Z_3^\text{\xmark}$}} \\ \hhline{~-----} 
\multicolumn{1}{c|}{\multirow{-2}{*}{\textbf{$Z_4^\text{\xmark}$}}} &  \centering \cellcolor[HTML]{EFEFEF}$r_8$ &  \centering \cellcolor[HTML]{C0C0C0}$r_{9}$ &  \centering \cellcolor[HTML]{9B9B9B}$r_{11}$ & \multicolumn{1}{c|}{ \centering \cellcolor[HTML]{C0C0C0}$r_{10}$} & \textbf{$Z_3^\text{\cmark}$} \\ \hhline{~----}
 & \textbf{$Z_1^\text{\cmark}$} & \multicolumn{2}{c|}{\textbf{$Z_1^\text{\xmark}$}} & \textbf{$Z_1^\text{\cmark}$} & 
\end{tabularx}
}
\caption{Four Objectives}
\label{rmb}
\end{subfigure}
\hfill
\\
\begin{subfigure}[b]{\textwidth}
\resizebox{\linewidth}{!}{%
\begin{tabularx}{\textwidth}{XX|X|X|XXX|XXX|X|X|XX}
\multicolumn{6}{c}{$Z_5^\text{\cmark}$} &  & \multicolumn{1}{c}{} & \multicolumn{6}{c}{$Z_5^\text{\xmark}$} \\
 & \multicolumn{2}{c|}{$Z_2^\text{\cmark}$} & \multicolumn{2}{c}{$Z_2^\text{\xmark}$} &  &  &  &  & \multicolumn{2}{c|}{$Z_2^\text{\cmark}$} & \multicolumn{2}{c}{$Z_2^\text{\xmark}$} &  \\ \cline{2-5} \cline{10-13}
\multicolumn{1}{c|}{} & \multicolumn{1}{c|}{$r_0$} & \multicolumn{1}{c|}{\cellcolor[HTML]{EFEFEF}$r_1$} & \multicolumn{1}{c|}{\cellcolor[HTML]{C0C0C0}$r_3$} & \multicolumn{1}{c|}{\cellcolor[HTML]{EFEFEF}$r_2$} & $Z_3^\text{\cmark}$ &  &  & \multicolumn{1}{c|}{} & \multicolumn{1}{c|}{\cellcolor[HTML]{EFEFEF}$r_{16}$} & \multicolumn{1}{c|}{\cellcolor[HTML]{C0C0C0}$r_{17}$} & \multicolumn{1}{c|}{\cellcolor[HTML]{9B9B9B}$r_{19}$} & \multicolumn{1}{c|}{\cellcolor[HTML]{C0C0C0}$r_{18}$} & $Z_3^\text{\cmark}$ \\ \cline{2-6} \cline{10-14} 
\multicolumn{1}{c|}{\multirow{-2}{*}{$Z_4^\text{\cmark}$}} & \multicolumn{1}{c|}{\cellcolor[HTML]{EFEFEF}$r_4$} & \multicolumn{1}{c|}{\cellcolor[HTML]{C0C0C0}$r_5$} & \multicolumn{1}{c|}{\cellcolor[HTML]{9B9B9B}{\color[HTML]{333333} $r_7$}} & \multicolumn{1}{c|}{\cellcolor[HTML]{C0C0C0}$r_6$} &  &  &  & \multicolumn{1}{c|}{\multirow{-2}{*}{$Z_4^\text{\cmark}$}} & \multicolumn{1}{c|}{\cellcolor[HTML]{C0C0C0}$r_{20}$} & \multicolumn{1}{c|}{\cellcolor[HTML]{9B9B9B}$r_{21}$} & \multicolumn{1}{c|}{\cellcolor[HTML]{656565}{\color[HTML]{FFFFFF} $r_{23}$}} & \multicolumn{1}{c|}{\cellcolor[HTML]{9B9B9B}$r_{22}$} &  \\ \cline{1-5} \cline{9-13}
\multicolumn{1}{c|}{} & \multicolumn{1}{c|}{\cellcolor[HTML]{C0C0C0}$r_{12}$} & \multicolumn{1}{c|}{\cellcolor[HTML]{9B9B9B}{\color[HTML]{333333} $r_{13}$}} & \multicolumn{1}{c|}{\cellcolor[HTML]{656565}{\color[HTML]{FFFFFF} $r_{15}$}} & \multicolumn{1}{c|}{\cellcolor[HTML]{9B9B9B}{\color[HTML]{333333} $r_{14}$}} & \multirow{-2}{*}{$Z_3^\text{\xmark}$} &  &  & \multicolumn{1}{c|}{} & \multicolumn{1}{c|}{\cellcolor[HTML]{9B9B9B}$r_{28}$} & \multicolumn{1}{c|}{\cellcolor[HTML]{656565}{\color[HTML]{FFFFFF} $r_{29}$}} & \multicolumn{1}{c|}{\cellcolor[HTML]{343434}{\color[HTML]{FFFFFF} $r_{31}$}} & \multicolumn{1}{c|}{\cellcolor[HTML]{656565}{\color[HTML]{FFFFFF} $r_{30}$}} & \multirow{-2}{*}{$Z_3^\text{\xmark}$} \\ \cline{2-6} \cline{10-14} 
\multicolumn{1}{c|}{\multirow{-2}{*}{$Z_4^\text{\xmark}$}} & \multicolumn{1}{c|}{\cellcolor[HTML]{EFEFEF}$r_8$} & \multicolumn{1}{c|}{\cellcolor[HTML]{C0C0C0}$r_9$} & \multicolumn{1}{c|}{\cellcolor[HTML]{9B9B9B}{\color[HTML]{333333} $r_{11}$}} & \multicolumn{1}{c|}{\cellcolor[HTML]{C0C0C0}$r_{10}$} & $Z_3^\text{\cmark}$ &  &  & \multicolumn{1}{c|}{\multirow{-2}{*}{$Z_4^\text{\xmark}$}} & \multicolumn{1}{c|}{\cellcolor[HTML]{C0C0C0}$r_{24}$} & \multicolumn{1}{c|}{\cellcolor[HTML]{9B9B9B}$r_{25}$} & \multicolumn{1}{c|}{\cellcolor[HTML]{656565}{\color[HTML]{FFFFFF} $r_{27}$}} & \multicolumn{1}{c|}{\cellcolor[HTML]{9B9B9B}$r_{26}$} & $Z_3^\text{\cmark}$ \\ \cline{2-5} \cline{10-13}
 & \multicolumn{1}{c|}{$Z_1^\text{\cmark}$} & \multicolumn{2}{c|}{$Z_1^\text{\xmark}$} & $Z_1^\text{\cmark}$ &  &  &  &  & \multicolumn{1}{c|}{$Z_1^\text{\cmark}$} & \multicolumn{2}{c|}{$Z_1^\text{\xmark}$} & $Z_1^\text{\cmark}$ & 
\end{tabularx}%
}
\caption{Five Objectives}
\label{rmc}
\end{subfigure}
\caption{Region map schematics, illustrating the use for 3, 4 or 5 objectives.}
\label{rm}
\end{figure}

Figure \ref{rm} shows the region map schematics for 3, 4 and 5 objectives. Each region is labelled according to the Gray code. A shading scheme is used to display the differences between regions, such that regions with lighter tones have a higher number of good objectives than regions with darker tones.

The use of these maps makes it easy to spot where solutions are concentrated in the objective space and the existence of trade-offs. When region $r_0$ is not empty, then there exist solutions with good values in all objectives, meaning that the problem could possibly be tackled with single-objective algorithms. Alternatively, that could mean that the threshold could be increased until trade-offs appear. 
Performing a range analysis in this region could provide further insights into the best approach to take. 
Conversely, if most solutions fall into region $r_{2^m-1}$, then it is possible that the threshold is set too high and should be lowered for more accurate results. 
If there are no solutions in $r_0$, but they are scattered throughout the map, then trade-offs have been identified and the map can be used to visualise them.

\subsection*{Threshold Analysis}

%Rodrigo - how much of this could be automated? Probably the same questions applies to some of the other things too?
%Jason: the entire step 3 could be automated. Actually it was when I performed the experiments. But because we are focused here on the results of the analysis I don't really explore that in the paper, but if we find a way to use this analysis during the optimisation, then it can be integrated into it.

As previously mentioned, moving the threshold for objectives may sometimes help to clarify and further investigate the trade-offs.
Region maps from modified threshold values can be used to determine additional information about the multiobjective nature of the problem, including:
\begin{itemize}
\item The maximum threshold such that all solutions are considered good in all objectives. This information can be useful for assessing whether the problem could be tackled as a single-objective problem. 
% Rodrigo - it seems that this could be automatically worked out without the trial and error of moving thresholds?
% My concern is that I think that this is just a case of setting the value to the lowest value for each objective, to force all solutions into it.
% Thus, do you mean all solutions, or that solutions are good in all objectives? 
% Or do both have value? even though one can be worked out without actually moving thresholds and recalculating?
%Jason: yeah, I wrote it wrong. It should say considered good in all objectives.
\item The minimum threshold to ensure the existence of trade-offs. This information could be used to empirically define a threshold in cases where domain knowledge is not available.
%By using this method to calculate the threshold we reduce the amount of input required from the decision-maker or the amount of domain knowledge required and assumptions.  
%Rodrigo - this seems to be the same point as above? I moved it here then commented out rather than deleted in case you think it adds info?
\item The steepness of the decrease in the number of solutions for all objectives across the thresholds. A higher steepness indicates that the instances evaluated are more alike in their value for that objective.
\end{itemize}

In this paper, the following method is used for this analysis:
First determine the maximum and minimum values for each objective, then determine $\alpha-1$ equally spaced threshold values for the objective which will divide this range into $\alpha$ parts with equal range, where $\alpha$ is a value chosen by the user. 
A region map is then generated for each of these threshold values.
%To do that, it is first important to normalise the objective values of all solutions using the minimum and maximum values found in the range analysis. Therefore thresholds for all criteria are scalars $t_i \in [0,1]$. We then discretise this interval into $k$ equidistant points and, for each value, we generate a region map with that threshold.
% Rodrigo: I don't see why you need to normalise to do this - what have I missed? Maybe I misunderstood?
%Jason: because when you normalise you put all values in the same range, hence you can say "we cannot find solution good in all objectives when we aim for at least 70% (for example) of quality on each objective". It is really just to have them on the same scale.

%Rodrigo - it seems that there should be ways to find meaningful threshold values from the data too. e.g. rather than splitting the ranges, what about splitting the points up equally?
% Also, there may be value in assessing each section using a statistical test? Woudl work well in the initial example scatter plot graph you gave.

%Jason: that could be tested, although I see that trying to split the points equally, the map might not highlight trade-offs. Also, using the suggested way, the threshold graph can tell us a little bit about the instances. Anyway, only way to know is testing this idea.

% And you use a single threshold for high-low. It won't work with K-maps, but what about having multiple points and using statistical tests to determine any relationships between points in that region?
%Jason: K-maps? Not sure how could I do that.

\subsection{Step 4 -- Multiobjective Scatter Plot Analysis:}

Finally, a scatter plot graph of all objectives of the normalised approximation set is assessed. 
The values are normalised and a scatter plot is drawn of all of the solutions, using the selected objective on the $x$ axis and the remaining objectives on the $y$ axis. Ideally, the objective of choice is well spread throughout its range in order to avoid gaps in the graph and the solutions being too concentrated in a small region, which would make the resulting graph difficult to read. 
It may be interesting to test different objectives in order to identify which objective provides more useful information, or whether multiple objectives provide different insights.
% Unnecessar,y I think, since implied:  all of them could be considered instead of only one. 

\begin{figure}[!ht]
\centering
\begin{subfigure}{0.4\linewidth}
\resizebox{\linewidth}{!}{%
\centering
\includegraphics[width=\linewidth]{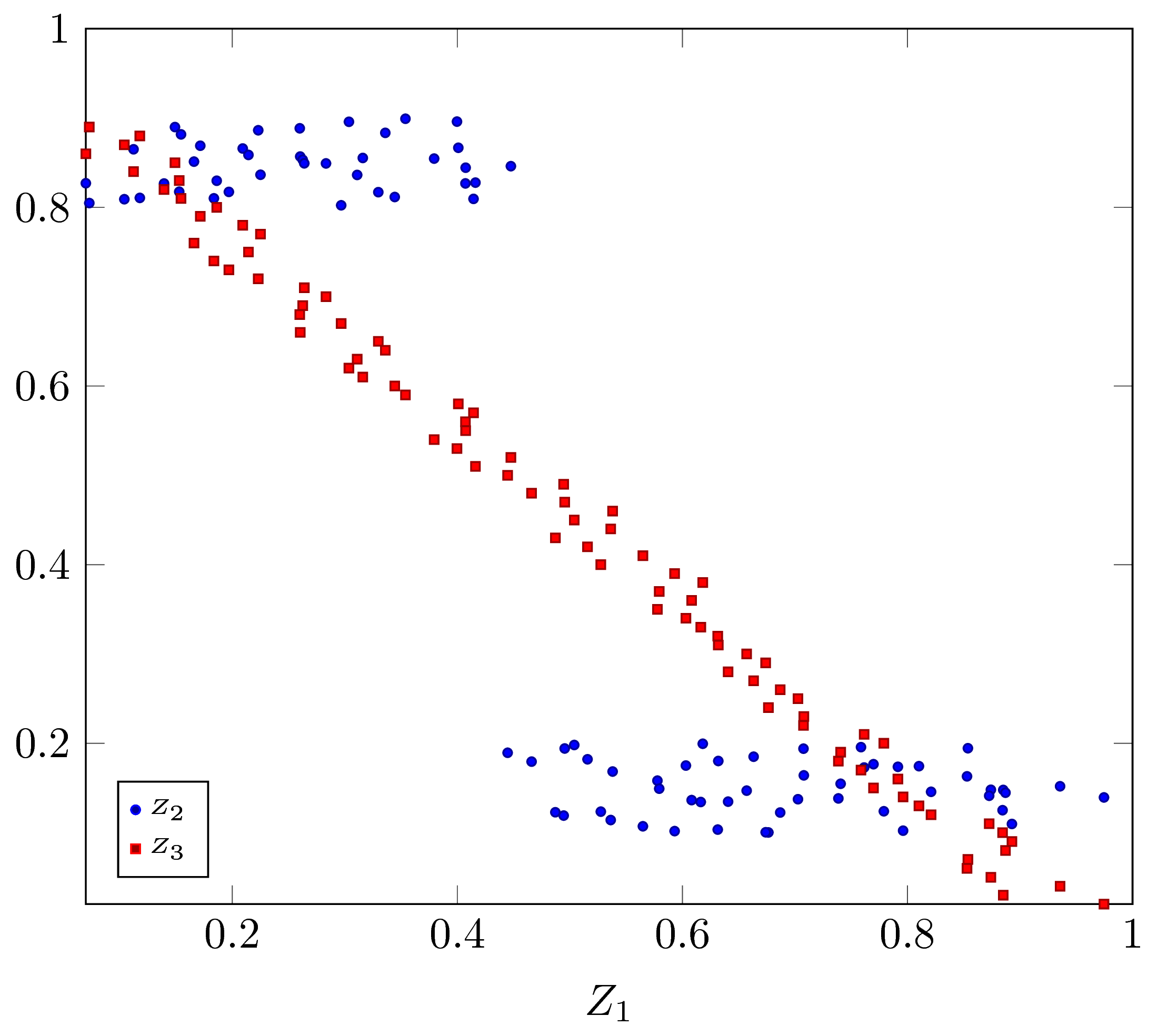}
}
\subcaption{Scatter plot of a three-objective \\ approximation set using $Z_1$ as pivot.}
\end{subfigure}
\begin{subfigure}{0.4\linewidth}
\resizebox{\linewidth}{!}{%
\includegraphics[width=\linewidth]{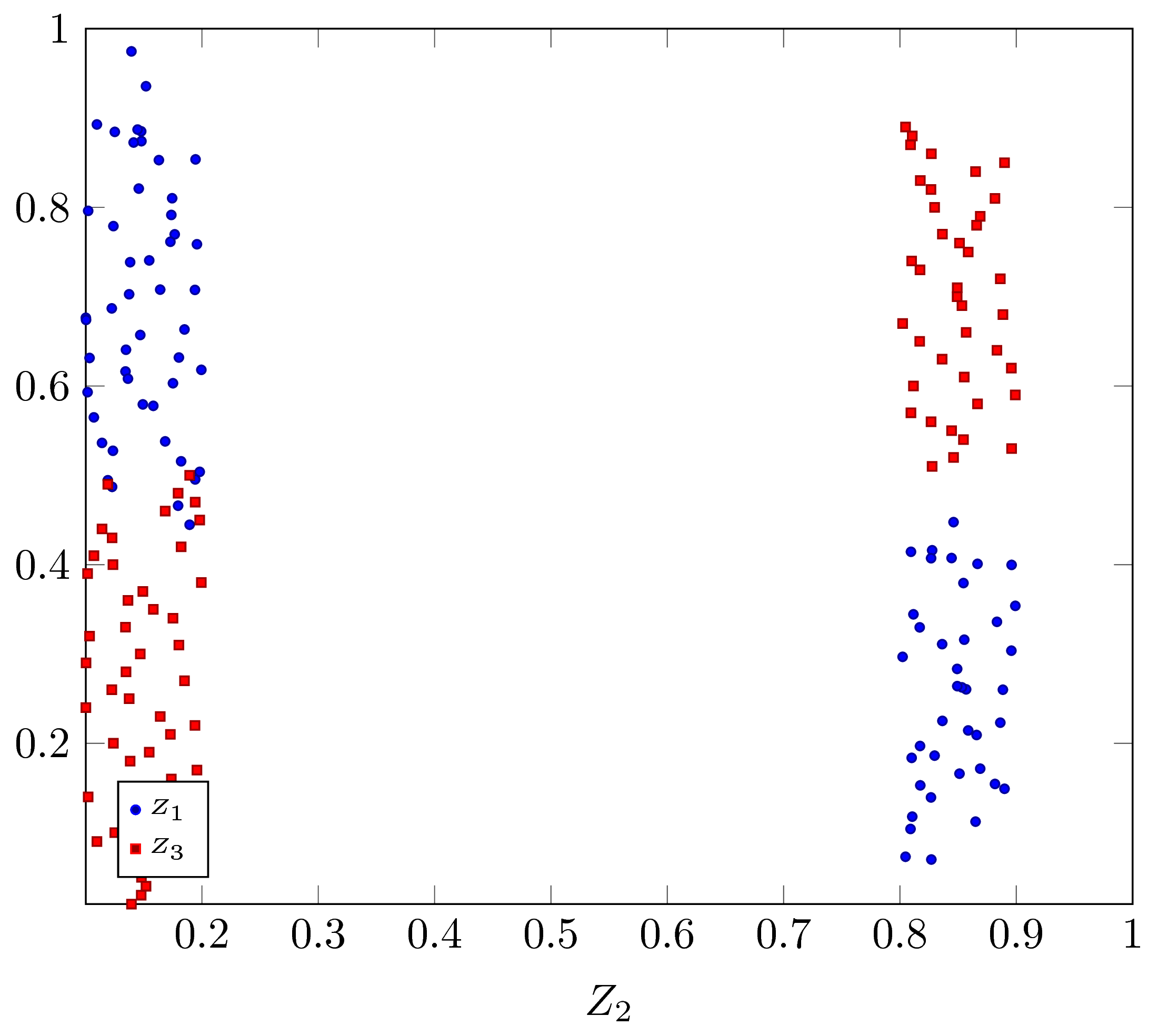}
}
\subcaption{Scatter plot of a three-objective \\ approximation set using $Z_2$ as pivot}
\end{subfigure}
\caption{Importance of choosing the right pivot on a three-objective approximation set: in a) $Z_1$--$Z_3$ the conflicting relationship is easily spotted, while in b) the same relationship can not be visualised.}
\label{fig:chooseobjective}
\end{figure}

Figure \ref{fig:chooseobjective} illustrates the impact of choosing the wrong objective as the pivot. In a) $Z_1$ is chosen and the conflicting relationship between $Z_1$--$Z_3$ can be clearly visualised, as well as the small range of values of $Z_2$. 
In b) $Z_2$ was chosen as the pivot and the conflicting relationship between $Z_1$--$Z_3$ is not prominent anymore. 
Additionally, in b) all values are situated within a small region of the plot space because $Z_2$ has values ranging only from [0.1,0.2] and [0.8,0.9]. 

The well-known pattern recognition capacities of the human brain are now utilised with a visual inspection of the resulting graph(s) to identify local relationships (conflicts and harmony), interesting patterns, gaps in the objective space, and well-spread trade-offs or isolated regions. This analysis can provide useful knowledge that can be used to tailor algorithms for the problem. Additionally, if multiple instances of a problem present consistent landscapes, then the information obtained and the algorithmic designed could be more likely to succeed in solving other previously unseen instances of the problem.

The following sections present experimental results from applying the proposed analysis and visualisation technique to three distinct MOPs: a multiobjective multidimensional knapsack problem, a multiobjective nurse scheduling problem and a multiobjective vehicle routing with time windows. All of the steps for the proposed analysis are performed for each of the datasets for each problem and the new knowledge obtained from doing so is discussed.

\section{Problem Scenarios}
\label{sScenarios}

In order to illustrate the analysis technique and provide a comparison between different problems, it has been applied to different multiobjective combinatorial problems. This will illustrate different benefits and insights from the technique.
%Moreover, we aim to show that within the same problem, the proposed technique identify multiple scenarios with distinct multiobjective natures.

\subsection{Multiobjective Multidimensional Knapsack Problem (MOMKP)}

In the multiobjective multidimensional knapsack problem \citep{Lust2010}, there are \textit{n} items $(i=1,\dots,n)$ with \textit{m} weights $w^i_j$ $(j=1,\dots,m)$ and \textit{p} profits $c^i_k$ $(k=1,\dots,p)$. The aim is to select a set of items to maximise the \textit{p} profits while not exceeding the capacities $W_j$ of the knapsack. Decision variable $x_i$ indicates if item $i$ is selected $(x_i=1)$ or not $(x_i=0)$. This problem can be formulated as follows:
\begin{equation*}
\begin{aligned}
& \text{maximise}
& & \sum_{i=1}^{n} c^i_k x_i  
& & & k=1,\dots,p\\
& \text{subject to}
& & \sum_{i=1}^{n} w^i_j x_i \leq W_j
& & & j=1,\dots, m \\
& & & x_i \in {0,1}
& & & i=1,\dots, n \\
\end{aligned}
\end{equation*}

Five MMKP datasets are considered here, each of which has five instances, all with $m=4$, $p=4$, $n=1000$ and $W_j=50000$. The first four datasets were generated following the guidelines in \citet{Bazgan2007} and are as follows:
\begin{itemize}
	\item \textbf{Set A}: Independent random instances where $w^i_j \in_N [1,1000]$ and $c^i_k \in_N [1,1000]$.
	\item \textbf{Set B}: Uncorrelated harmonious instances where \\ $w^i_j \in_N [1,1000]$, $c^i_1 \in_N [1,1000]$ and $c^i_k \in_N [max$ $\{c^i_{k-1}$ $-100, 1\}$ $,min\{c^i_{k-1}+100, 1000\}]$ for $k=(2,3,4)$.
	\item \textbf{Set C}: Uncorrelated conflicting instances where $w^i_j \in_N [1,1000]$, $c^i_1 \in_N [1,1000]$ and $c^i_k$  $\in_N$ $[max\{900-c^i_{k-1},$  $1\},min\{1100-c^i_{k-1}, 1000\}]$ for $k=(2,3,4)$. 
	\item \textbf{Set D}: Correlated conflicting instances where $w^i_1 \in_N [max\{900-|c^i_1-c^i_4|, 1\}$, $min\{1100-|c^i_1-c^i_4|, 1000\}]$, $w^i_j \in_N [max\{900-|c^i_k-c^i_{k-1}|, 1\}$, $min\{1100-|c^i_k-c^i_{k-1}|, 1000\}]$,  $c^i_1 \in_N [1,1000]$ and $c^i_k \in_N [max\{900-c^i_{k-1}, 1\}$, $min\{1100-c^i_{k-1}, 1000\}]$ for $k=(2,3,4)$.
	\item \textbf{Set X}: Correlated special set where:
	\begin{equation*}
\begin{aligned}
& r^i \in_R [0,1] \\
& c^i_k \begin{cases}
            c^i_1 \in_N [900,1000], c^i_2 \in_N [c^i_1,1000], c^i_k \in_N [0,100]\text{ for }k=(3,4). & \quad \text{if } r^i \leq 0.1 \\
            c^i_3 \in_N [900,1000], c^i_4 \in_N [c^3_1,1000], c^i_k \in_N [0,100]\text{ for }k=(1,2). & \quad \text{if } 0.1 < r^i \leq 0.2 \\
            c^i_1 \in_N [900,1000], c^i_3 \in_N [c^1_1,1000], c^i_k \in_N [0,100]\text{ for }k=(2,4). & \quad \text{if } 0.2 < r^i \leq 0.3 \\
           c^i_2, c^i_3 \in_N [900,1000] \text{ and } c^i_1,c^i_4 \in_N [0,100]. & \quad \text{if } 0.3 < r^i \leq 0.4 \\
           c^i_k \in_N [0,1000]. & \quad \text{if } r^i > 0.4\\
        \end{cases} \\
& w^i_1 = c^i_1+c^i_2+c^i_3 \\
& w^i_2 = c^i_2+c^i_3+c^i_4 \\
& w^i_3 = c^i_1+c^i_3+c^i_4 \\
& w^i_4 = c^i_1+c^i_2+c^i_4 \\
\end{aligned}
\end{equation*}
\end{itemize}

Set $A$ contains only independent objectives. In set $B$ all objectives are harmonious. Set $C$ contains three pairs of conflicting objectives, $(Z_2,Z_1)$, $(Z_3,Z_2)$ and $(Z_3,Z_4)$, while the weights are uncorrelated. Set $D$ has conflicting objectives, like set $C$, but the weights are correlated to the objective values. The fifth set $X$ was tailored to provide a misleading scenario.

\subsection{Multiobjective Nurse Scheduling Problem (MONSP)}

An extension of the nurse scheduling problem (NSP) presented by \citet{maenhout2007} is also considered. 
The original problem consists of creating a work plan for a set of nurses, that must be assigned shifts across a time period (week or month) to cover hospital requirements. The problem is composed of a number of hard and soft constraints to be considered when generating each individual schedule. 
The hard constraints define strict work regulations and the soft constraints define working policies and desirable aspects. For this work the number of objectives is artificially extended using preferences and soft constraint violations as additional objectives. 

\begin{itemize}

\item \textit{Hard Constraints:} The NSP has two hard constraints, both of which aim to avoid certain shift patterns. The first forbids the assignment of an afternoon shift before a morning shift. The second forbids the assignment of a night shift before a morning or an afternoon shift.

\item \textit{Soft Constraints:} There are four soft constraints that represent nurses' preferences. The first defines a minimum and maximum number of working days within the scheduling period. The second defines a minimum and maximum number of consecutive working assignments. The third defines a minimum and maximum number of assignments of each shift type in the scheduling period. The fourth soft constraint defines a minimum and maximum number of consecutive shift assignments of the same type.

\item \textit{Preferences:} The preferences represent the main objectives for the MONSP. Each preference is a numerical value relating to a nurse and a shift. It could represent the nurse preference regarding that shift, the hospital preference to assign that nurse to that shift, the patients' preferences, or a combination of these.
%Rodrigo - is it correct that it could be a combination? I assume that you could have one weight which is a linear sum if you wanted to?
\end{itemize}

The MONSP can be defined as follows. 
Given a set $N$ of nurses, where $N = \{n_1, n_2, \dots, n_{|N|}\}$, a set $D$ of days where $D = \{d_1, d_2, \dots, d_{|D|}\}$ and a set $S$ of shifts where $S = \{s_1, s_2, \dots, s_{|S|}\}$, find an assignment of shifts for each day $d_i \in D$, such that each nurse $n_j \in N$ has been assigned a specific shift $s_k \in S$ and hard constraints are met. 
Shifts refer to either a given working period (early, day or night shift) or a rest period (free shift).
Each tuple (\textit{nurse,day,shift}) has a set of $m$ values representing different preferences regarding that nurse being assigned to that shift on that day. 
The different preferences may represent personal preferences, the hospital preferences, patient preferences, costs, etc. 
The (multiple) objectives to optimise consist of the sum of each type of preference value, and the sum of a set of soft constraints.
% Rodrigo: You said: "that we consider as an extra objective." so I assume that they must be summed otherwise there would be multiple rather than just one. Please clarify.

%The data used in this work was provided by the NSPLib, a library with problem instances for the NSP \citep{maenhout2005}. A duty roster, or roster, is a sequence of $|D|$ shifts assigned to a nurse. We consider $S_d$ to be the set of shifts required for day $d$. The NSPLib contains two types of files: base files and case files. Base files present the shift requirements for each day, number of nurses available and a preferences matrix $n_{max} \times d_{max}$. Case files present different constraint set-ups. We employ the preference matrices of additional base files as extra objectives. Hence, the base files only describe the problem requirements and number of nurses, case files describe the constraints and objective files (which are extracted from the base files) describe the preferences matrices.
% Rodrigo: I think that the previous paragraph could be clearer - but I don't think I udnerstand it enough to clarify it :(  Is it repeating earlier info? Can some be removed? Are all of the details needed? Are you just saying that you take their files and add files for new preference objectives? If so then I think it could be said more concisely. Potentially replace by:

 % and see below - which I put with the table comments.

\begin{table}[!ht]
\caption{Generational patterns for the MONSP using the NSPLib files. Letters represent base files.}
  \centering
  \resizebox{0.8\linewidth}{!}{%
\begin{tabular}{rrrrrrrrrrrrrrrrrrrr}
\cline{1-6}\cline{8-13}\cline{15-20}\multicolumn{6}{c}{\textbf{Independent Set}}  &       & \multicolumn{6}{c}{\textbf{Objectives-Dependent Set}} &       & \multicolumn{6}{c}{\textbf{Requirements-Dependent Set}} \bigstrut\\
\cline{1-6}\cline{8-13}\cline{15-20}\multicolumn{1}{c}{\textbf{Case}} & \multicolumn{1}{c}{\textbf{Base}} & \multicolumn{1}{c}{\boldmath{}\textbf{$Z_1$}\unboldmath{}} & \multicolumn{1}{c}{\boldmath{}\textbf{$Z_2$}\unboldmath{}} & \multicolumn{1}{c}{\boldmath{}\textbf{$Z_3$}\unboldmath{}} & \multicolumn{1}{c}{\boldmath{}\textbf{$Z_4 \dagger$}\unboldmath{}} &       & \multicolumn{1}{c}{\textbf{Case}} & \multicolumn{1}{c}{\textbf{Base}} & \multicolumn{1}{c}{\boldmath{}\textbf{$Z_1$}\unboldmath{}} & \multicolumn{1}{c}{\boldmath{}\textbf{$Z_2$}\unboldmath{}} & \multicolumn{1}{c}{\boldmath{}\textbf{$Z_3$}\unboldmath{}} & \multicolumn{1}{c}{\boldmath{}\textbf{$Z_4 \dagger$}\unboldmath{}} &       & \multicolumn{1}{c}{\textbf{Case}} & \multicolumn{1}{c}{\textbf{Base}} & \multicolumn{1}{c}{\boldmath{}\textbf{$Z_1$}\unboldmath{}} & \multicolumn{1}{c}{\boldmath{}\textbf{$Z_2$}\unboldmath{}} & \multicolumn{1}{c}{\boldmath{}\textbf{$Z_3$}\unboldmath{}} & \multicolumn{1}{c}{\boldmath{}\textbf{$Z_4 \dagger$}\unboldmath{}} \bigstrut\\
\cline{1-6}\cline{8-13}\cline{15-20}\multicolumn{1}{c}{1} & \multicolumn{1}{c}{A} & \multicolumn{1}{c}{A} & \multicolumn{1}{c}{B} & \multicolumn{1}{c}{C} & \multicolumn{1}{c}{V} &       & \multicolumn{1}{c}{1} & \multicolumn{1}{c}{A} & \multicolumn{1}{c}{A} & \multicolumn{1}{c}{B} & \multicolumn{1}{c}{C} & \multicolumn{1}{c}{V} &       & \multicolumn{1}{c}{1} & \multicolumn{1}{c}{A} & \multicolumn{1}{c}{A} & \multicolumn{1}{c}{B} & \multicolumn{1}{c}{C} & \multicolumn{1}{c}{V} \bigstrut[t]\\
\multicolumn{1}{c}{1} & \multicolumn{1}{c}{D} & \multicolumn{1}{c}{D} & \multicolumn{1}{c}{E} & \multicolumn{1}{c}{F} & \multicolumn{1}{c}{V} &       & \multicolumn{1}{c}{1} & \multicolumn{1}{c}{B} & \multicolumn{1}{c}{A} & \multicolumn{1}{c}{B} & \multicolumn{1}{c}{C} & \multicolumn{1}{c}{V} &       & \multicolumn{1}{c}{1} & \multicolumn{1}{c}{A} & \multicolumn{1}{c}{D} & \multicolumn{1}{c}{E} & \multicolumn{1}{c}{F} & \multicolumn{1}{c}{V} \\
\multicolumn{1}{c}{1} & \multicolumn{1}{c}{G} & \multicolumn{1}{c}{G} & \multicolumn{1}{c}{H} & \multicolumn{1}{c}{I} & \multicolumn{1}{c}{V} &       & \multicolumn{1}{c}{1} & \multicolumn{1}{c}{C} & \multicolumn{1}{c}{A} & \multicolumn{1}{c}{B} & \multicolumn{1}{c}{C} & \multicolumn{1}{c}{V} &       & \multicolumn{1}{c}{1} & \multicolumn{1}{c}{A} & \multicolumn{1}{c}{G} & \multicolumn{1}{c}{H} & \multicolumn{1}{c}{I} & \multicolumn{1}{c}{V} \bigstrut[b]\\
\cline{1-6}\cline{8-13}\cline{15-20}\multicolumn{1}{c}{2} & \multicolumn{1}{c}{A} & \multicolumn{1}{c}{A} & \multicolumn{1}{c}{B} & \multicolumn{1}{c}{C} & \multicolumn{1}{c}{V} &       & \multicolumn{1}{c}{2} & \multicolumn{1}{c}{A} & \multicolumn{1}{c}{A} & \multicolumn{1}{c}{B} & \multicolumn{1}{c}{C} & \multicolumn{1}{c}{V} &       & \multicolumn{1}{c}{2} & \multicolumn{1}{c}{A} & \multicolumn{1}{c}{A} & \multicolumn{1}{c}{B} & \multicolumn{1}{c}{C} & \multicolumn{1}{c}{V} \bigstrut[t]\\
\multicolumn{1}{c}{2} & \multicolumn{1}{c}{D} & \multicolumn{1}{c}{D} & \multicolumn{1}{c}{E} & \multicolumn{1}{c}{F} & \multicolumn{1}{c}{V} &       & \multicolumn{1}{c}{2} & \multicolumn{1}{c}{B} & \multicolumn{1}{c}{A} & \multicolumn{1}{c}{B} & \multicolumn{1}{c}{C} & \multicolumn{1}{c}{V} &       & \multicolumn{1}{c}{2} & \multicolumn{1}{c}{A} & \multicolumn{1}{c}{D} & \multicolumn{1}{c}{E} & \multicolumn{1}{c}{F} & \multicolumn{1}{c}{V} \\
\multicolumn{1}{c}{2} & \multicolumn{1}{c}{G} & \multicolumn{1}{c}{G} & \multicolumn{1}{c}{H} & \multicolumn{1}{c}{I} & \multicolumn{1}{c}{V} &       & \multicolumn{1}{c}{2} & \multicolumn{1}{c}{C} & \multicolumn{1}{c}{A} & \multicolumn{1}{c}{B} & \multicolumn{1}{c}{C} & \multicolumn{1}{c}{V} &       & \multicolumn{1}{c}{2} & \multicolumn{1}{c}{A} & \multicolumn{1}{c}{G} & \multicolumn{1}{c}{H} & \multicolumn{1}{c}{I} & \multicolumn{1}{c}{V} \bigstrut[b]\\
\cline{1-6}\cline{8-13}\cline{15-20}\multicolumn{1}{c}{3} & \multicolumn{1}{c}{A} & \multicolumn{1}{c}{A} & \multicolumn{1}{c}{B} & \multicolumn{1}{c}{C} & \multicolumn{1}{c}{V} &       & \multicolumn{1}{c}{3} & \multicolumn{1}{c}{A} & \multicolumn{1}{c}{A} & \multicolumn{1}{c}{B} & \multicolumn{1}{c}{C} & \multicolumn{1}{c}{V} &       & \multicolumn{1}{c}{3} & \multicolumn{1}{c}{A} & \multicolumn{1}{c}{A} & \multicolumn{1}{c}{B} & \multicolumn{1}{c}{C} & \multicolumn{1}{c}{V} \bigstrut[t]\\
\multicolumn{1}{c}{3} & \multicolumn{1}{c}{D} & \multicolumn{1}{c}{D} & \multicolumn{1}{c}{E} & \multicolumn{1}{c}{F} & \multicolumn{1}{c}{V} &       & \multicolumn{1}{c}{3} & \multicolumn{1}{c}{B} & \multicolumn{1}{c}{A} & \multicolumn{1}{c}{B} & \multicolumn{1}{c}{C} & \multicolumn{1}{c}{V} &       & \multicolumn{1}{c}{3} & \multicolumn{1}{c}{A} & \multicolumn{1}{c}{D} & \multicolumn{1}{c}{E} & \multicolumn{1}{c}{F} & \multicolumn{1}{c}{V} \\
\multicolumn{1}{c}{3} & \multicolumn{1}{c}{G} & \multicolumn{1}{c}{G} & \multicolumn{1}{c}{H} & \multicolumn{1}{c}{I} & \multicolumn{1}{c}{V} &       & \multicolumn{1}{c}{3} & \multicolumn{1}{c}{C} & \multicolumn{1}{c}{A} & \multicolumn{1}{c}{B} & \multicolumn{1}{c}{C} & \multicolumn{1}{c}{V} &       & \multicolumn{1}{c}{3} & \multicolumn{1}{c}{A} & \multicolumn{1}{c}{G} & \multicolumn{1}{c}{H} & \multicolumn{1}{c}{I} & \multicolumn{1}{c}{V} \bigstrut[b]\\
\cline{1-6}\cline{8-13}\cline{15-20}\multicolumn{1}{c}{4} & \multicolumn{1}{c}{A} & \multicolumn{1}{c}{A} & \multicolumn{1}{c}{B} & \multicolumn{1}{c}{C} & \multicolumn{1}{c}{V} &       & \multicolumn{1}{c}{4} & \multicolumn{1}{c}{A} & \multicolumn{1}{c}{A} & \multicolumn{1}{c}{B} & \multicolumn{1}{c}{C} & \multicolumn{1}{c}{V} &       & \multicolumn{1}{c}{4} & \multicolumn{1}{c}{A} & \multicolumn{1}{c}{A} & \multicolumn{1}{c}{B} & \multicolumn{1}{c}{C} & \multicolumn{1}{c}{V} \bigstrut[t]\\
\multicolumn{1}{c}{4} & \multicolumn{1}{c}{D} & \multicolumn{1}{c}{D} & \multicolumn{1}{c}{E} & \multicolumn{1}{c}{F} & \multicolumn{1}{c}{V} &       & \multicolumn{1}{c}{4} & \multicolumn{1}{c}{B} & \multicolumn{1}{c}{A} & \multicolumn{1}{c}{B} & \multicolumn{1}{c}{C} & \multicolumn{1}{c}{V} &       & \multicolumn{1}{c}{4} & \multicolumn{1}{c}{A} & \multicolumn{1}{c}{D} & \multicolumn{1}{c}{E} & \multicolumn{1}{c}{F} & \multicolumn{1}{c}{V} \\
\multicolumn{1}{c}{4} & \multicolumn{1}{c}{G} & \multicolumn{1}{c}{G} & \multicolumn{1}{c}{H} & \multicolumn{1}{c}{I} & \multicolumn{1}{c}{V} &       & \multicolumn{1}{c}{4} & \multicolumn{1}{c}{C} & \multicolumn{1}{c}{A} & \multicolumn{1}{c}{B} & \multicolumn{1}{c}{C} & \multicolumn{1}{c}{V} &       & \multicolumn{1}{c}{4} & \multicolumn{1}{c}{A} & \multicolumn{1}{c}{G} & \multicolumn{1}{c}{H} & \multicolumn{1}{c}{I} & \multicolumn{1}{c}{V} \bigstrut[b]\\
\cline{1-6}\cline{8-13}\cline{15-20}\multicolumn{1}{c}{5} & \multicolumn{1}{c}{A} & \multicolumn{1}{c}{A} & \multicolumn{1}{c}{B} & \multicolumn{1}{c}{C} & \multicolumn{1}{c}{V} &       & \multicolumn{1}{c}{5} & \multicolumn{1}{c}{A} & \multicolumn{1}{c}{A} & \multicolumn{1}{c}{B} & \multicolumn{1}{c}{C} & \multicolumn{1}{c}{V} &       & \multicolumn{1}{c}{5} & \multicolumn{1}{c}{A} & \multicolumn{1}{c}{A} & \multicolumn{1}{c}{B} & \multicolumn{1}{c}{C} & \multicolumn{1}{c}{V} \bigstrut[t]\\
\multicolumn{1}{c}{5} & \multicolumn{1}{c}{D} & \multicolumn{1}{c}{D} & \multicolumn{1}{c}{E} & \multicolumn{1}{c}{F} & \multicolumn{1}{c}{V} &       & \multicolumn{1}{c}{5} & \multicolumn{1}{c}{B} & \multicolumn{1}{c}{A} & \multicolumn{1}{c}{B} & \multicolumn{1}{c}{C} & \multicolumn{1}{c}{V} &       & \multicolumn{1}{c}{5} & \multicolumn{1}{c}{A} & \multicolumn{1}{c}{D} & \multicolumn{1}{c}{E} & \multicolumn{1}{c}{F} & \multicolumn{1}{c}{V} \\
\multicolumn{1}{c}{5} & \multicolumn{1}{c}{G} & \multicolumn{1}{c}{G} & \multicolumn{1}{c}{H} & \multicolumn{1}{c}{I} & \multicolumn{1}{c}{V} &       & \multicolumn{1}{c}{5} & \multicolumn{1}{c}{C} & \multicolumn{1}{c}{A} & \multicolumn{1}{c}{B} & \multicolumn{1}{c}{C} & \multicolumn{1}{c}{V} &       & \multicolumn{1}{c}{5} & \multicolumn{1}{c}{A} & \multicolumn{1}{c}{G} & \multicolumn{1}{c}{H} & \multicolumn{1}{c}{I} & \multicolumn{1}{c}{V} \bigstrut[b]\\
\cline{1-6}\cline{8-13}\cline{15-20}\multicolumn{1}{c}{6} & \multicolumn{1}{c}{A} & \multicolumn{1}{c}{A} & \multicolumn{1}{c}{B} & \multicolumn{1}{c}{C} & \multicolumn{1}{c}{V} &       & \multicolumn{1}{c}{6} & \multicolumn{1}{c}{A} & \multicolumn{1}{c}{A} & \multicolumn{1}{c}{B} & \multicolumn{1}{c}{C} & \multicolumn{1}{c}{V} &       & \multicolumn{1}{c}{6} & \multicolumn{1}{c}{A} & \multicolumn{1}{c}{A} & \multicolumn{1}{c}{B} & \multicolumn{1}{c}{C} & \multicolumn{1}{c}{V} \bigstrut[t]\\
\multicolumn{1}{c}{6} & \multicolumn{1}{c}{D} & \multicolumn{1}{c}{D} & \multicolumn{1}{c}{E} & \multicolumn{1}{c}{F} & \multicolumn{1}{c}{V} &       & \multicolumn{1}{c}{6} & \multicolumn{1}{c}{B} & \multicolumn{1}{c}{A} & \multicolumn{1}{c}{B} & \multicolumn{1}{c}{C} & \multicolumn{1}{c}{V} &       & \multicolumn{1}{c}{6} & \multicolumn{1}{c}{A} & \multicolumn{1}{c}{D} & \multicolumn{1}{c}{E} & \multicolumn{1}{c}{F} & \multicolumn{1}{c}{V} \\
\multicolumn{1}{c}{6} & \multicolumn{1}{c}{G} & \multicolumn{1}{c}{G} & \multicolumn{1}{c}{H} & \multicolumn{1}{c}{I} & \multicolumn{1}{c}{V} &       & \multicolumn{1}{c}{6} & \multicolumn{1}{c}{C} & \multicolumn{1}{c}{A} & \multicolumn{1}{c}{B} & \multicolumn{1}{c}{C} & \multicolumn{1}{c}{V} &       & \multicolumn{1}{c}{6} & \multicolumn{1}{c}{A} & \multicolumn{1}{c}{G} & \multicolumn{1}{c}{H} & \multicolumn{1}{c}{I} & \multicolumn{1}{c}{V} \bigstrut[b]\\
\cline{1-6}\cline{8-13}\cline{15-20}\multicolumn{1}{c}{7} & \multicolumn{1}{c}{A} & \multicolumn{1}{c}{A} & \multicolumn{1}{c}{B} & \multicolumn{1}{c}{C} & \multicolumn{1}{c}{V} &       & \multicolumn{1}{c}{7} & \multicolumn{1}{c}{A} & \multicolumn{1}{c}{A} & \multicolumn{1}{c}{B} & \multicolumn{1}{c}{C} & \multicolumn{1}{c}{V} &       & \multicolumn{1}{c}{7} & \multicolumn{1}{c}{A} & \multicolumn{1}{c}{A} & \multicolumn{1}{c}{B} & \multicolumn{1}{c}{C} & \multicolumn{1}{c}{V} \bigstrut[t]\\
\multicolumn{1}{c}{7} & \multicolumn{1}{c}{D} & \multicolumn{1}{c}{D} & \multicolumn{1}{c}{E} & \multicolumn{1}{c}{F} & \multicolumn{1}{c}{V} &       & \multicolumn{1}{c}{7} & \multicolumn{1}{c}{B} & \multicolumn{1}{c}{A} & \multicolumn{1}{c}{B} & \multicolumn{1}{c}{C} & \multicolumn{1}{c}{V} &       & \multicolumn{1}{c}{7} & \multicolumn{1}{c}{A} & \multicolumn{1}{c}{D} & \multicolumn{1}{c}{E} & \multicolumn{1}{c}{F} & \multicolumn{1}{c}{V} \\
\multicolumn{1}{c}{7} & \multicolumn{1}{c}{G} & \multicolumn{1}{c}{G} & \multicolumn{1}{c}{H} & \multicolumn{1}{c}{I} & \multicolumn{1}{c}{V} &       & \multicolumn{1}{c}{7} & \multicolumn{1}{c}{C} & \multicolumn{1}{c}{A} & \multicolumn{1}{c}{B} & \multicolumn{1}{c}{C} & \multicolumn{1}{c}{V} &       & \multicolumn{1}{c}{7} & \multicolumn{1}{c}{A} & \multicolumn{1}{c}{G} & \multicolumn{1}{c}{H} & \multicolumn{1}{c}{I} & \multicolumn{1}{c}{V} \bigstrut[b]\\
\cline{1-6}\cline{8-13}\cline{15-20}\multicolumn{1}{c}{8} & \multicolumn{1}{c}{A} & \multicolumn{1}{c}{A} & \multicolumn{1}{c}{B} & \multicolumn{1}{c}{C} & \multicolumn{1}{c}{V} &       & \multicolumn{1}{c}{8} & \multicolumn{1}{c}{A} & \multicolumn{1}{c}{A} & \multicolumn{1}{c}{B} & \multicolumn{1}{c}{C} & \multicolumn{1}{c}{V} &       & \multicolumn{1}{c}{8} & \multicolumn{1}{c}{A} & \multicolumn{1}{c}{A} & \multicolumn{1}{c}{B} & \multicolumn{1}{c}{C} & \multicolumn{1}{c}{V} \bigstrut[t]\\
\multicolumn{1}{c}{8} & \multicolumn{1}{c}{D} & \multicolumn{1}{c}{D} & \multicolumn{1}{c}{E} & \multicolumn{1}{c}{F} & \multicolumn{1}{c}{V} &       & \multicolumn{1}{c}{8} & \multicolumn{1}{c}{B} & \multicolumn{1}{c}{A} & \multicolumn{1}{c}{B} & \multicolumn{1}{c}{C} & \multicolumn{1}{c}{V} &       & \multicolumn{1}{c}{8} & \multicolumn{1}{c}{A} & \multicolumn{1}{c}{D} & \multicolumn{1}{c}{E} & \multicolumn{1}{c}{F} & \multicolumn{1}{c}{V} \\
\multicolumn{1}{c}{8} & \multicolumn{1}{c}{G} & \multicolumn{1}{c}{G} & \multicolumn{1}{c}{H} & \multicolumn{1}{c}{I} & \multicolumn{1}{c}{V} &       & \multicolumn{1}{c}{8} & \multicolumn{1}{c}{C} & \multicolumn{1}{c}{A} & \multicolumn{1}{c}{B} & \multicolumn{1}{c}{C} & \multicolumn{1}{c}{V} &       & \multicolumn{1}{c}{8} & \multicolumn{1}{c}{A} & \multicolumn{1}{c}{G} & \multicolumn{1}{c}{H} & \multicolumn{1}{c}{I} & \multicolumn{1}{c}{V} \bigstrut[b]\\
\cline{1-6}\cline{8-13}\cline{15-20}\multicolumn{20}{l}{$\dagger$ $Z_4$ is the number of soft constraint violations.} \bigstrut[t]\\
\end{tabular}%
}
  \label{tab:nsp}%
\end{table}%

%An attempt was made to create sets with different multiobjective natures by combining bases and case files and we use the analysis technique to assess if we achieved this goal. 
NSPLib is a library with problem instances for the NSP \citep{maenhout2005}. 
Two types of files are present: base files, which describe the preferences, and case files which specify the constraints. The data from that library was used in this work, with the addition of extra base files  to represent the new preference values. Four objectives are considered, so each instance is composed of the existing case file and base file, plus three new objective files. Table \ref{tab:nsp} presents the details for the different sets, where A to I are different base files (different preference values), $Z_1$ to $Z_3$ are the new preference objective values and $Z_4$ is the count of soft constraint violations. 
%Therefore, given two instances (\textit{1,A,A,B,C,V}) and (\textit{2,B,A,B,C,V}), 1 and 2 refer to the case files, A, B and C refers to three distinct base files to be used as base or as objectives. The files used for the first and the second instances are the same for the same letter, hence the file A in the first instance is the same file for A in the second instance. For each case file we generated three instances following the set rule, hence each set have 24 instances. Following we briefly describe each set:
\begin{itemize}
    \item \textbf{Independent Set (IS):} There is no correlation between base file and objectives files (all objective files are distinct). The aim is that the fitness landscape will be different for different instances.
    \item \textbf{Objective-Dependent Set (ODS):} This set uses the same objectives files for all instances, but the base files  change. This set assess whether keeping the objectives information across all instances will generate a standardised fitness landscape. 
% Rodrigo : Arrghhh - I don't get this. You said that the base was the objective too, and that changes, so the objectives do change. Please clarify this.
    \item \textbf{Requirements-Dependent Set (RDS):} This set employs a common base file for all instances but defines different objectives files for each instance. The goal of this scenario is to assess whether keeping a problem base and changing the objectives information has an impact on the fitness landscape of the instances in this set.
\end{itemize}

Weekly problems with 25, 50 and 100 nurses were generated using random base and objective files following the generational pattern presented. Hence, each set considered in this work is constituted of 75 instances totalling 225 problem instances.
% Rodrigo: How can this be random when you said which file (A to I) is used in each case? Please clarify

\subsection{Multiobjective Vehicle Routing Problem with Time Windows \\(MOVRPTW)}

A Multiobjective Vehicle Routing Problem with Time Windows (MOVRPTW) is defined on a graph $G=(V, E)$ where $V$ is the set of vertices representing the depot (vertex $0$) and the customers (vertices $1 \dots n$) where each customer has a demand $p_i$ ($i=1,\dots,n$). The depot has $h$ identical vehicles available, with capacity $Q$ that must satisfy all demands from all customers. The edge set $E$ denotes all possible connections between all vertices. Each edge (from vertex $i$ to vertex $j$) has a cost that represents the distance or time, denoted by $c_{ij}$. Each customer must be served during a time window $[a_{ij},b_{ij}]$. If a vehicle arrives early, it must wait until the beginning of the time window. Once the vehicle arrives at the customer, it stays there until the delivery is completed, for a duration known as the service time $s$.

\citet{Castro-Gutierrez2011} proposed a benchmark set for the MOVRPTW. They consider five objectives: number of vehicles ($Z_1$), total travel distance ($Z_2$), makespan ($Z_3$), total waiting time (in case of early arrival) ($Z_4$), and total delay time ($Z_5$). They designed their instances based on different characteristics of the problem and each instance is a combination of these features. The features that constitute an instance are:
\begin{itemize}
\item \textbf{Number of customers:} 50, 150 and 250 customers.
\item \textbf{Time window:} five different profiles \hl{\{$tw0$, $tw1$, $tw2$, $tw3$, $tw4$\} }of time windows across eight hours were used.
These are defined below in terms of minutes from the start of the working day (0 = 8:00am, 480 = 4:00pm, etc.). Customers are randomly assigned to one time-window:
\begin{itemize}
    \item{$tw0$:} [0,480].
    \item{$tw1$:} [0,160],[160,320],[320,480].
    \item{$tw2$:} [0,130],[175,305],[350,480].
    \item{$tw3$:} [0,100],[190,290],[350,480].
    \item{$tw4$:} all time-windows from $tw0, tw1, tw2$ and $tw3$.
\end{itemize}

\item \textbf{Demand types:} three values for the demandare used \hl{\{10, 20, 30\}}, uniformly distributed.
\item \textbf{Vehicle capacity:} the capacity of the vehicles are calculated according to a $\delta$ parameter such that $Q=\underline{D}+\delta/100(\overline{D}-\underline{D})$, where $\underline{D}= \smash{\displaystyle\max_{(i=1,\dots,n)}} p_i$ and $\overline{D}=\sum_{i=1}^{n}p_i$. The dataset considers three $\delta$ values ($\delta0=60$, $\delta1=20$, $\delta2=5$).
\item \textbf{Service time:} three values of service time \hl{\{10, 20, 30\}} are used, uniformly distributed.
\end{itemize}

%In total, there are 45 instances, and these can be downloaded from \citep{castro}.
In total, there are 45 instances, and these are publicly available\footnote{\url{https://github.com/psxjpc/} -- Acessed: 2016-09-22}.

\section{Application of the Proposed Technique}
\label{sResults}

To obtain the approximation sets, the MOEA/D \citep{Zhang2007} and NSGA-II \citep{Deb2002a} algorithms were chosen. Initial experiments showed that both algorithms struggled to produce good approximation sets due to the difficulty of tackling many-objective problems, hence to improve the obtained Pareto fronts the following mechanism was applied to all datasets: for each instance, a single-objective genetic algorithm was executed for each objective alone, then both NSGA-II and MOEA/D algorithms were executed on each pair and triplet of objectives. All of the non-dominated solutions obtained  were then combined into a single archive which was used to initialise further executions of the algorithms:
Three runs of MOEA/D and three runs of NSGA-II were then executed, considering all objectives, where half of the individuals for the initial population were randomly drawn from the archive and the other half were randomly generated. 
The final approximation set was formed from all of the non-dominated solutions which were found in the process. Both NSGA-II and MOEA/D used a population of 200 individuals, binary tournament selection and half uniform crossover \citep{j.eshelman:the} for $1.5\times 10^6$ function evaluations each run. 

\subsection{Multiobjective Multidimensional Knapsack Problem}

Using the process described above, non-dominated sets were obtained, with approximately 900 combined solutions for each instance in set $A$, 3 for each instance in set $B$, 550 for each instance in set $C$, 1500 for each instance in set $D$ and 2500 combined solutions for each instance in set $X$. Only a few non-dominated solutions were found for instances in set $B$ because the objectives there are strongly harmonious. Therefore, when maximising one of the objectives, the other objectives are also maximised, resulting in insufficient data for some of the analysis steps. However, the results are presented for completeness and illustrate that the number of solutions obtained can have a major impact on the analysis, and that it is important to have a comprehensive set, with enough well-spread solutions.

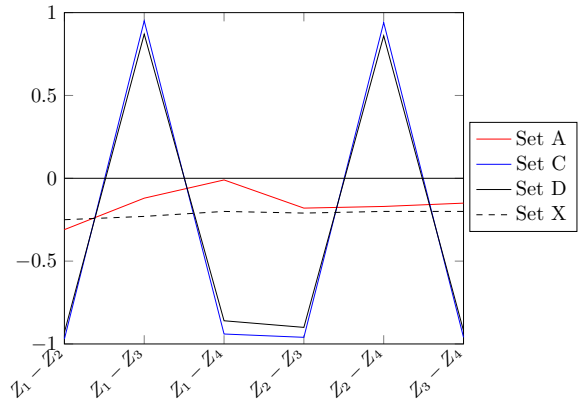
\begin{figure}[!h]
\centering
\resizebox{0.5\textwidth}{!}{%
\begin{tikzpicture}
\begin{axis}[
legend style={at={(1.15,0.67)},
	    	 anchor=north,legend columns=1},
xticklabels={{\small $Z_1-Z_2$},{\small $Z_1-Z_3$},{\small $Z_1-Z_4$},{\small $Z_2-Z_3$},{\small $Z_2-Z_4$},{\small $Z_3-Z_4$}},
xtick=data,
scaled y ticks = false,
x tick label style={rotate=45, anchor=east},
ymax=1,
ymin=-1,
xmin=0,
xmax=5,
cycle list name=color list,
]

\addplot coordinates
{
(0,-0.31)
(1,-0.12)
(2,-0.01)
(3,-0.18)
(4,-0.17)
(5,-0.15)
};

\addplot coordinates
{
(0,-0.97)
(1,0.95)
(2,-0.94)
(3,-0.96)
(4,0.94)
(5,-0.96)
};

\addplot coordinates
{
(0,-0.93)
(1,0.87)
(2,-0.86)
(3,-0.90)
(4,0.86)
(5,-0.92)
};

\addplot[color=black,dashed] coordinates
{
(0,-0.25)
(1,-0.23)
(2,-0.20)
(3,-0.21)
(4,-0.20)
(5,-0.20)
};

\addplot[] coordinates
{
(0,0)
(1,0)
(2,0)
(3,0)
(4,0)
(5,0)
(6,0)
(7,0)
(8,0)
(9,0)
};

\legend{Set A, Set C, Set D, Set X}
\end{axis}
\end{tikzpicture}
}
\caption{Pairwise correlations for each set of instances.}
\label{momkp:step1}
\end{figure}

\textbf{Step 1 - Global Pairwise Relationships Analysis.} The results of this step are in Figure \ref{momkp:step1}. Note that coefficient values for set $B$ are not provided for the reason given above. The figure presents the individual pairwise correlation value for each combination of objectives. As expected for fully independent objectives, set $A$ has values close to $0$. The values for sets $C$ and $D$ are also predictably close to either $1$ or $-1$ indicating global conflicting or harmonious relationships. Set $X$ has values similar to $A$ -- they do not reveal a strong global relationship between objectives as they are closer to $0$ than to $1$ or $-1$. Also, in set $X$ it is not possible to decompose the decision variables according to the objectives, as every item has all weights and values above zero. 

\begin{figure}[!h]
\centering
\begin{subfigure}[b]{0.28\textwidth}
\centering
\includegraphics[width=\linewidth]{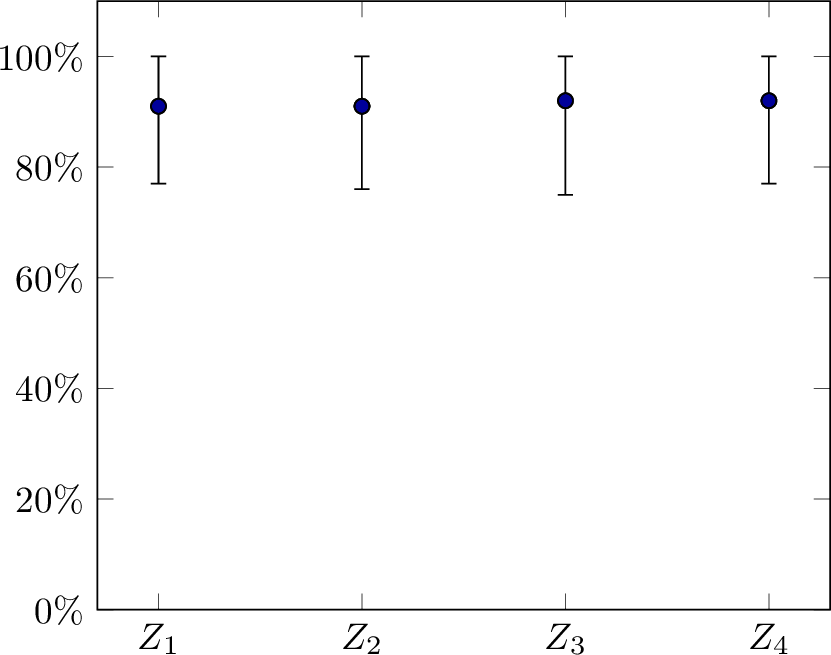}
\caption{Set A}
\label{momkp:step2a}
\end{subfigure}
\begin{subfigure}[b]{0.28\textwidth}
\centering
\includegraphics[width=\linewidth]{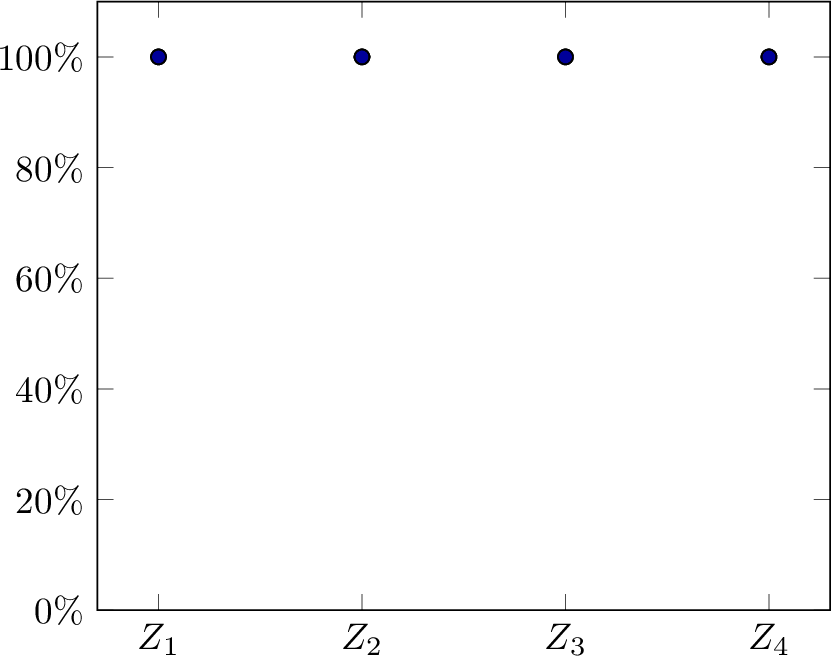}
\caption{Set B}
\label{momkp:step2b}
\end{subfigure}
\begin{subfigure}[b]{0.28\textwidth}
\centering
\includegraphics[width=\linewidth]{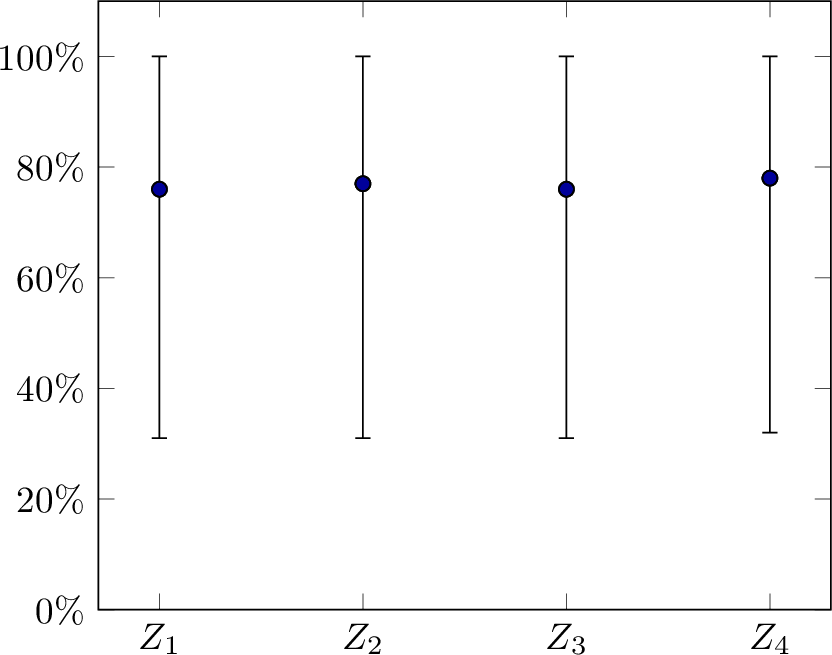}
\caption{Set C}
\label{momkp:step2c}
\end{subfigure}
\\
\begin{subfigure}[b]{0.28\textwidth}
\centering
\includegraphics[width=\linewidth]{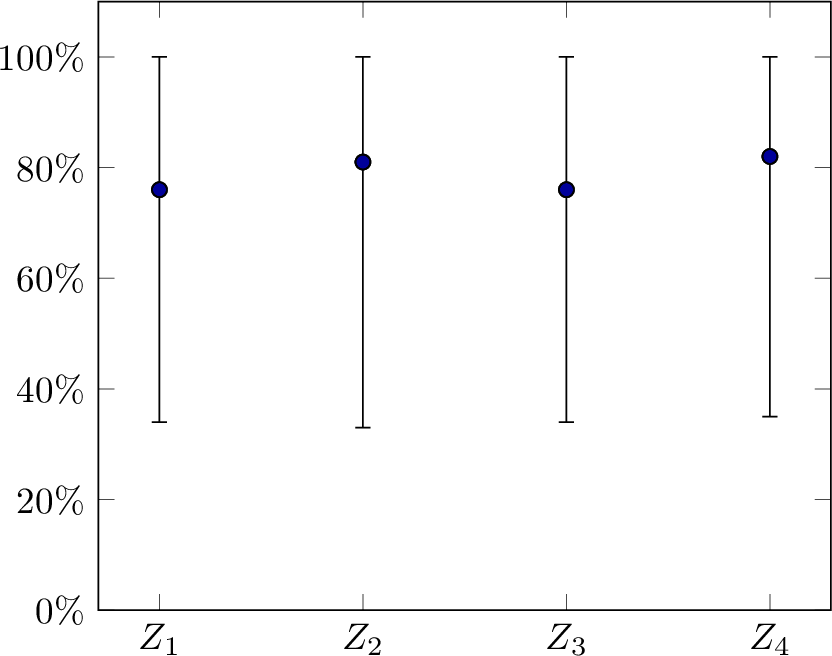}
\caption{Set D}
\label{momkp:step2d}
\end{subfigure}
\begin{subfigure}[b]{0.28\textwidth}
\centering
\includegraphics[width=\linewidth]{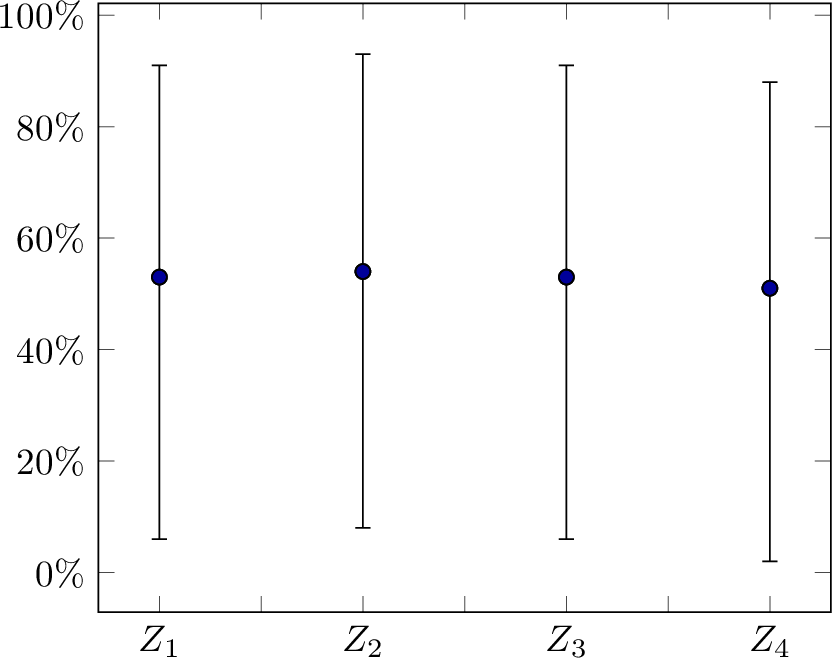}
\caption{Set X}
\label{momkp:step2x}
\end{subfigure}
\caption{Results for the objective ranges analysis. In each line, the central point is the average value for the set, with the top and bottom of the line representing the maximum and minimum values found, respectively.}
\label{momkp:step2}
\end{figure}

\textbf{Step 2 - Objective Range Analysis.} The results for this step are in Figure \ref{momkp:step2}. \hl{The $x$ axis presents the sets grouped by objective. The $y$ axis presents the minimum, maximum and average value of each objective as a percentage of the overall maximum value found for the respective objective} Considering set $B$, it can be seen, that the set presents small ranges of less than 0.3\% on average. In this dataset, all objectives are harmonious and the solutions found are all located in a small region of the objective space. These few solutions dominated all other solutions explored. Both sets $C$ and $D$ present similar results to each other with large ranges for each objective (over 60\%). This is expected since these are instances with conflicting objectives and present global trade-offs. The large ranges mean that while there are solutions with good values for a given objective, at least one other objective has a poor value.

While the global pairwise relationship analysis (step 1) hinted that sets $A$ and $X$ were similar, the difference between them now becomes clear with the results from step 2. In set $A$, each objective range is around 24.0\% of the maximum value -- the smallest ranges excluding the harmonious instances -- whilst in set $X$ the ranges go up to 84.9\%, the largest range found. Thus, it is apparent that the ranges for set $X$ are actually closer to those for set $D$, a conflicting scenario with global trade-offs.

%table2

\begin{figure}[!ht]
\centering
\resizebox{0.4\textwidth}{!}{%
\begin{tikzpicture}
	\begin{axis}[
		xlabel=Threshold,
		ylabel=\# Of Instances,
		xmin=0,
		xmax=1,
		legend pos= north east,	
		ymin=0,
		]
	\addplot[color=blue] coordinates {
(	1	,	0	)
(	0.98	,	0	)
(	0.96	,	0	)
(	0.94	,	0	)
(	0.92	,	0	)
(	0.9	,	0	)
(	0.88	,	0	)
(	0.86	,	0	)
(	0.84	,	0	)
(	0.82	,	0	)
(	0.8	,	0	)
(	0.78	,	0	)
(	0.76	,	0	)
(	0.74	,	0	)
(	0.72	,	0	)
(	0.7	,	0	)
(	0.68	,	2	)
(	0.66	,	3	)
(	0.64	,	5	)
(	0.62	,	5	)
(	0.6	,	5	)
(	0.58	,	5	)
(	0.56	,	5	)
(	0.54	,	5	)
(	0.52	,	5	)
(	0.5	,	5	)
(	0.48	,	5	)
(	0.46	,	5	)
(	0.44	,	5	)
(	0.42	,	5	)
(	0.4	,	5	)
(	0.38	,	5	)
(	0.36	,	5	)
(	0.34	,	5	)
(	0.32	,	5	)
(	0.3	,	5	)
(	0.28	,	5	)
(	0.26	,	5	)
(	0.24	,	5	)
(	0.22	,	5	)
(	0.2	,	5	)
(	0.18	,	5	)
(	0.16	,	5	)
(	0.14	,	5	)
(	0.12	,	5	)
(	0.1	,	5	)
(	0.08	,	5	)
(	0.06	,	5	)
(	0.04	,	5	)
(	0.02	,	5	)
	};	
	
		\addplot[color=black] coordinates {
(	1	,	0	)
(	0.98	,	0	)
(	0.96	,	0	)
(	0.94	,	0	)
(	0.92	,	0	)
(	0.9	,	0	)
(	0.88	,	0	)
(	0.86	,	0	)
(	0.84	,	0	)
(	0.82	,	0	)
(	0.8	,	0	)
(	0.78	,	0	)
(	0.76	,	0	)
(	0.74	,	0	)
(	0.72	,	0	)
(	0.7	,	0	)
(	0.68	,	0	)
(	0.66	,	2	)
(	0.64	,	4	)
(	0.62	,	5	)
(	0.6	,	5	)
(	0.58	,	5	)
(	0.56	,	5	)
(	0.54	,	5	)
(	0.52	,	5	)
(	0.5	,	5	)
(	0.48	,	5	)
(	0.46	,	5	)
(	0.44	,	5	)
(	0.42	,	5	)
(	0.4	,	5	)
(	0.38	,	5	)
(	0.36	,	5	)
(	0.34	,	5	)
(	0.32	,	5	)
(	0.3	,	5	)
(	0.28	,	5	)
(	0.26	,	5	)
(	0.24	,	5	)
(	0.22	,	5	)
(	0.2	,	5	)
(	0.18	,	5	)
(	0.16	,	5	)
(	0.14	,	5	)
(	0.12	,	5	)
(	0.1	,	5	)
(	0.08	,	5	)
(	0.06	,	5	)
(	0.04	,	5	)
(	0.02	,	5	)
	};	
	
		\addplot[color=red] coordinates {
(	1	,	0	)
(	0.98	,	0	)
(	0.96	,	0	)
(	0.94	,	0	)
(	0.92	,	0	)
(	0.9	,	0	)
(	0.88	,	0	)
(	0.86	,	0	)
(	0.84	,	0	)
(	0.82	,	0	)
(	0.8	,	0	)
(	0.78	,	0	)
(	0.76	,	0	)
(	0.74	,	0	)
(	0.72	,	0	)
(	0.7	,	0	)
(	0.68	,	2	)
(	0.66	,	5	)
(	0.64	,	5	)
(	0.62	,	5	)
(	0.6	,	5	)
(	0.58	,	5	)
(	0.56	,	5	)
(	0.54	,	5	)
(	0.52	,	5	)
(	0.5	,	5	)
(	0.48	,	5	)
(	0.46	,	5	)
(	0.44	,	5	)
(	0.42	,	5	)
(	0.4	,	5	)
(	0.38	,	5	)
(	0.36	,	5	)
(	0.34	,	5	)
(	0.32	,	5	)
(	0.3	,	5	)
(	0.28	,	5	)
(	0.26	,	5	)
(	0.24	,	5	)
(	0.22	,	5	)
(	0.2	,	5	)
(	0.18	,	5	)
(	0.16	,	5	)
(	0.14	,	5	)
(	0.12	,	5	)
(	0.1	,	5	)
(	0.08	,	5	)
(	0.06	,	5	)
(	0.04	,	5	)
(	0.02	,	5	)
	};	
	
		\addplot[color=black,dashed] coordinates {
(	1	,	0	)
(	0.98	,	0	)
(	0.96	,	0	)
(	0.94	,	0	)
(	0.92	,	0	)
(	0.9	,	0	)
(	0.88	,	0	)
(	0.86	,	0	)
(	0.84	,	0	)
(	0.82	,	0	)
(	0.8	,	0	)
(	0.78	,	0	)
(	0.76	,	0	)
(	0.74	,	0	)
(	0.72	,	0	)
(	0.7	,	0	)
(	0.68	,	0	)
(	0.66	,	0	)
(	0.64	,	1	)
(	0.62	,	1	)
(	0.6	,	1	)
(	0.58	,	1	)
(	0.56	,	1	)
(	0.54	,	2	)
(	0.52	,	4	)
(	0.5	,	5	)
(	0.48	,	5	)
(	0.46	,	5	)
(	0.44	,	5	)
(	0.42	,	5	)
(	0.4	,	5	)
(	0.38	,	5	)
(	0.36	,	5	)
(	0.34	,	5	)
(	0.32	,	5	)
(	0.3	,	5	)
(	0.28	,	5	)
(	0.26	,	5	)
(	0.24	,	5	)
(	0.22	,	5	)
(	0.2	,	5	)
(	0.18	,	5	)
(	0.16	,	5	)
(	0.14	,	5	)
(	0.12	,	5	)
(	0.1	,	5	)
(	0.08	,	5	)
(	0.06	,	5	)
(	0.04	,	5	)
(	0.02	,	5	)
	};	
	
\legend{A,C,D,X}
	\end{axis}
\end{tikzpicture}
}
\caption{Threshold analysis showing the number of scenarios with solutions found in region $r_0$ when the threshold decreases.}
\label{momkp:step3a}
\end{figure}
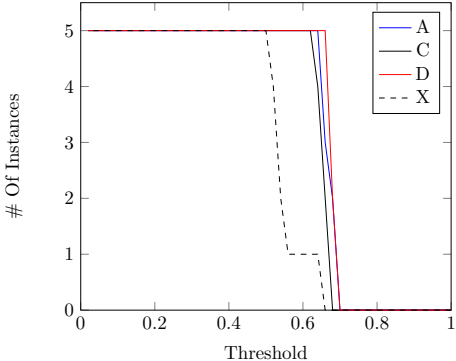

%attending to reviewer 4
\textbf{Step 3 - Trade-Off Regions Analysis.} The results for this step are in Figures \ref{momkp:step3a}--\ref{momkp:step3e}. Firstly, Figure~\ref{momkp:step3a} shows the number of instances with solutions in $r_0$ when the threshold is increased (normalised values). All instances present a similar threshold behaviour and with a threshold of roughly 0.7 there are no solutions with good values for all objectives. i.e. solutions with over 70\% quality in all objectives could not be found. Also, the steepness of the lines in the chart is very acute, indicating that all instances from a same set have very similar fitness landscapes.

\begin{figure}[!ht]
\centering
\begin{subfigure}{0.4\linewidth}
\resizebox{\linewidth}{!}{%
\begin{tabular}{r|c|c|c|c|r}
\multicolumn{1}{c}{} & \multicolumn{2}{c|}{$Z_2^\text{ \cmark}$} & \multicolumn{2}{c}{$Z_2^\text{ \xmark}$} &  \bigstrut[b]\\ \hhline{|~|-|-|-|-|}
\cline{2-5}\multicolumn{1}{c|}{\multirow{2}[4]{*}{$Z_4^\text{ \cmark}$}} & 0.0\% & {\cellcolor[HTML]{EFEFEF}{0.3\%}} & {\cellcolor[HTML]{C0C0C0}{0.9\%}} & {\cellcolor[HTML]{EFEFEF}{0.1\%}} & \multicolumn{1}{c}{$Z_3^\text{ \cmark}$} \bigstrut\\ \hhline{|~|-|-|-|-|}
\cline{2-6}\multicolumn{1}{c|}{} & {\cellcolor[HTML]{EFEFEF}{0.5\%}} & {\cellcolor[HTML]{C0C0C0}{5.3\%}} & {\cellcolor[HTML]{9B9B9B}{17.1\%}} & {\cellcolor[HTML]{C0C0C0}{4.6\%}} & \multicolumn{1}{c}{\multirow{2}[4]{*}{$Z_3^\text{ \xmark}$}} \bigstrut\\ \hhline{|~|-|-|-|-|}
\cline{1-5}\multicolumn{1}{c|}{\multirow{2}[4]{*}{$Z_4^\text{ \xmark}$}} & {\cellcolor[HTML]{C0C0C0}{8.7\%}} & {\cellcolor[HTML]{9B9B9B}{7.4\%}} & {\cellcolor[HTML]{656565}{\color[HTML]{FFFFFF}13.8\%}} & {\cellcolor[HTML]{9B9B9B}{11.6\%}} & \multicolumn{1}{c}{} \bigstrut\\ \hhline{|~|-|-|-|-|}
\cline{2-6}\multicolumn{1}{c|}{} & {\cellcolor[HTML]{EFEFEF}{4.4\%}} & {\cellcolor[HTML]{C0C0C0}{8.5\%}} & {\cellcolor[HTML]{9B9B9B}{10.5\%}} & {\cellcolor[HTML]{C0C0C0}{6.5\%}} & \multicolumn{1}{c}{$Z_3^\text{ \cmark}$} \bigstrut\\ \hhline{|~|-|-|-|-|}
\cline{2-5}\multicolumn{1}{r}{} & $Z_1^\text{ \cmark}$     & \multicolumn{2}{c|}{$Z_1^\text{ \xmark}$} & \multicolumn{1}{c}{$Z_1^\text{ \cmark}$} &  \bigstrut[t]\\
\end{tabular}%
}
\subcaption{Overall distribution of solutions.}
\end{subfigure}
\begin{subfigure}{0.4\linewidth}
\resizebox{\linewidth}{!}{%
\begin{tabular}{r|c|c|c|c|r}
\multicolumn{1}{c}{} & \multicolumn{2}{c|}{$Z_2^\text{ \cmark}$} & \multicolumn{2}{c}{$Z_2^\text{ \xmark}$} &  \bigstrut[b]\\ \hhline{|~|-|-|-|-|}
\cline{2-5}\multicolumn{1}{c|}{\multirow{2}[4]{*}{$Z_4^\text{ \cmark}$}} & 0\% & {\cellcolor[HTML]{EFEFEF}{20\%}} & {\cellcolor[HTML]{C0C0C0}{80\%}} & {\cellcolor[HTML]{EFEFEF}{20\%}} & \multicolumn{1}{c}{$Z_3^\text{ \cmark}$} \bigstrut\\ \hhline{|~|-|-|-|-|}
\cline{2-6}\multicolumn{1}{c|}{} & {\cellcolor[HTML]{EFEFEF}{40\%}} & {\cellcolor[HTML]{C0C0C0}{100\%}} & {\cellcolor[HTML]{9B9B9B}{100\%}} & {\cellcolor[HTML]{C0C0C0}{100\%}} & \multicolumn{1}{c}{\multirow{2}[4]{*}{$Z_3^\text{ \xmark}$}} \bigstrut\\ \hhline{|~|-|-|-|-|}
\cline{1-5}\multicolumn{1}{c|}{\multirow{2}[4]{*}{$Z_4^\text{ \xmark}$}} & {\cellcolor[HTML]{C0C0C0}{100\%}} & {\cellcolor[HTML]{9B9B9B}{100\%}} & {\cellcolor[HTML]{656565}{\color[HTML]{FFFFFF}100\%}} & {\cellcolor[HTML]{9B9B9B}{100\%}} & \multicolumn{1}{c}{} \bigstrut\\ \hhline{|~|-|-|-|-|}
\cline{2-6}\multicolumn{1}{c|}{} & {\cellcolor[HTML]{EFEFEF}{80\%}} & {\cellcolor[HTML]{C0C0C0}{100\%}} & {\cellcolor[HTML]{9B9B9B}{100\%}} & {\cellcolor[HTML]{C0C0C0}{100\%}} & \multicolumn{1}{c}{$Z_3^\text{ \cmark}$} \bigstrut\\ \hhline{|~|-|-|-|-|}
\cline{2-5}\multicolumn{1}{r}{} & $Z_1^\text{ \cmark}$     & \multicolumn{2}{c|}{$Z_1^\text{ \xmark}$} & \multicolumn{1}{c}{$Z_1^\text{ \cmark}$} &  \bigstrut[t]\\
\end{tabular}%
}
\subcaption{Frequency of instances.}
\end{subfigure}
\caption{Results for the trade-off regions analysis for Set A.}
\label{momkp:step3b}
\end{figure}

Region maps were calculated for each instance. The number of solutions in each region was computed and the map shows the average percentage of the solutions which lie in each region for each set. The range threshold was set to the minimum value such that there are no solutions in $r_0$. 

In addition to the distribution of solutions map, the frequency of instances region map is also introduced here, displaying the percentage of instances that contain at least one solution in a region. A value of 100\% in a region means that in all instances at least one solution can be found in that region. In set $A$ (Figure \ref{momkp:step3b}) the front is well distributed. Solutions can be found in all regions, scattered throughout the objective space, as a result of the independent objectives.

\begin{figure}[!ht]
\centering
\begin{subfigure}{0.4\linewidth}
\resizebox{\linewidth}{!}{%
\begin{tabular}{r|c|c|c|c|r}
\multicolumn{1}{c}{} & \multicolumn{2}{c|}{$Z_2^\text{ \cmark}$} & \multicolumn{2}{c}{$Z_2^\text{ \xmark}$} &  \bigstrut[b]\\ \hhline{|~|-|-|-|-|}
\cline{2-5}\multicolumn{1}{c|}{\multirow{2}[4]{*}{$Z_4^\text{ \cmark}$}} & 0.0\% & {\cellcolor[HTML]{EFEFEF}{0.0\%}} & {\cellcolor[HTML]{C0C0C0}{0.0\%}} & {\cellcolor[HTML]{EFEFEF}{0.0\%}} & \multicolumn{1}{c}{$Z_3^\text{ \cmark}$} \bigstrut\\ \hhline{|~|-|-|-|-|}
\cline{2-6}\multicolumn{1}{c|}{} & {\cellcolor[HTML]{EFEFEF}{0.0\%}} & {\cellcolor[HTML]{C0C0C0}{20.6\%}} & {\cellcolor[HTML]{9B9B9B}{2.9\%}} & {\cellcolor[HTML]{C0C0C0}{10.6\%}} & \multicolumn{1}{c}{\multirow{2}[4]{*}{$Z_3^\text{ \xmark}$}} \bigstrut\\ \hhline{|~|-|-|-|-|}
\cline{1-5}\multicolumn{1}{c|}{\multirow{2}[4]{*}{$Z_4^\text{ \xmark}$}} & {\cellcolor[HTML]{C0C0C0}{0.0\%}} & {\cellcolor[HTML]{9B9B9B}{8.7\%}} & {\cellcolor[HTML]{656565}{\color[HTML]{FFFFFF}6.6\%}} & {\cellcolor[HTML]{9B9B9B}{6.2\%}} & \multicolumn{1}{c}{} \bigstrut\\ \hhline{|~|-|-|-|-|}
\cline{2-6}\multicolumn{1}{c|}{} & {\cellcolor[HTML]{EFEFEF}{0.0\%}} & {\cellcolor[HTML]{C0C0C0}{13.3\%}} & {\cellcolor[HTML]{9B9B9B}{8.2\%}} & {\cellcolor[HTML]{C0C0C0}{22.9\%}} & \multicolumn{1}{c}{$Z_3^\text{ \cmark}$} \bigstrut\\ \hhline{|~|-|-|-|-|}
\cline{2-5}\multicolumn{1}{r}{} & $Z_1^\text{ \cmark}$     & \multicolumn{2}{c|}{$Z_1^\text{ \xmark}$} & \multicolumn{1}{c}{$Z_1^\text{ \cmark}$} &  \bigstrut[t]\\
\end{tabular}%
}
\subcaption{Overall distribution of solutions.}
\end{subfigure}
\begin{subfigure}{0.4\linewidth}
\resizebox{\linewidth}{!}{%
\begin{tabular}{r|c|c|c|c|r}
\multicolumn{1}{c}{} & \multicolumn{2}{c|}{$Z_2^\text{ \cmark}$} & \multicolumn{2}{c}{$Z_2^\text{ \xmark}$} &  \bigstrut[b]\\ \hhline{|~|-|-|-|-|}
\cline{2-5}\multicolumn{1}{c|}{\multirow{2}[4]{*}{$Z_4^\text{ \cmark}$}} & 0\% & {\cellcolor[HTML]{EFEFEF}{0\%}} & {\cellcolor[HTML]{C0C0C0}{0\%}} & {\cellcolor[HTML]{EFEFEF}{0\%}} & \multicolumn{1}{c}{$Z_3^\text{ \cmark}$} \bigstrut\\ \hhline{|~|-|-|-|-|}
\cline{2-6}\multicolumn{1}{c|}{} & {\cellcolor[HTML]{EFEFEF}{0\%}} & {\cellcolor[HTML]{C0C0C0}{100\%}} & {\cellcolor[HTML]{9B9B9B}{100\%}} & {\cellcolor[HTML]{C0C0C0}{100\%}} & \multicolumn{1}{c}{\multirow{2}[4]{*}{$Z_3^\text{ \xmark}$}} \bigstrut\\ \hhline{|~|-|-|-|-|}
\cline{1-5}\multicolumn{1}{c|}{\multirow{2}[4]{*}{$Z_4^\text{ \xmark}$}} & {\cellcolor[HTML]{C0C0C0}{0\%}} & {\cellcolor[HTML]{9B9B9B}{100\%}} & {\cellcolor[HTML]{656565}{\color[HTML]{FFFFFF}100\%}} & {\cellcolor[HTML]{9B9B9B}{100\%}} & \multicolumn{1}{c}{} \bigstrut\\ \hhline{|~|-|-|-|-|}
\cline{2-6}\multicolumn{1}{c|}{} & {\cellcolor[HTML]{EFEFEF}{0\%}} & {\cellcolor[HTML]{C0C0C0}{100\%}} & {\cellcolor[HTML]{9B9B9B}{100\%}} & {\cellcolor[HTML]{C0C0C0}{100\%}} & \multicolumn{1}{c}{$Z_3^\text{ \cmark}$} \bigstrut\\ \hhline{|~|-|-|-|-|}
\cline{2-5}\multicolumn{1}{r}{} & $Z_1^\text{ \cmark}$     & \multicolumn{2}{c|}{$Z_1^\text{ \xmark}$} & \multicolumn{1}{c}{$Z_1^\text{ \cmark}$} &  \bigstrut[t]\\
\end{tabular}%
}
\subcaption{Frequency of instances.}
\end{subfigure}
\caption{Results for the trade-off regions analysis for Set C.}
\label{momkp:step3c}
\end{figure}

\begin{figure}[!ht]
\centering
\begin{subfigure}{0.4\linewidth}
\resizebox{\linewidth}{!}{%
\begin{tabular}{r|c|c|c|c|r}
\multicolumn{1}{c}{} & \multicolumn{2}{c|}{$Z_2^\text{ \cmark}$} & \multicolumn{2}{c}{$Z_2^\text{ \xmark}$} &  \bigstrut[b]\\ \hhline{|~|-|-|-|-|}
\cline{2-5}\multicolumn{1}{c|}{\multirow{2}[4]{*}{$Z_4^\text{ \cmark}$}} & 0.0\% & {\cellcolor[HTML]{EFEFEF}{0.0\%}} & {\cellcolor[HTML]{C0C0C0}{0.0\%}} & {\cellcolor[HTML]{EFEFEF}{0.0\%}} & \multicolumn{1}{c}{$Z_3^\text{ \cmark}$} \bigstrut\\ \hhline{|~|-|-|-|-|}
\cline{2-6}\multicolumn{1}{c|}{} & {\cellcolor[HTML]{EFEFEF}{0.0\%}} & {\cellcolor[HTML]{C0C0C0}{54.1\%}} & {\cellcolor[HTML]{9B9B9B}{3.6\%}} & {\cellcolor[HTML]{C0C0C0}{5.5\%}} & \multicolumn{1}{c}{\multirow{2}[4]{*}{$Z_3^\text{ \xmark}$}} \bigstrut\\ \hhline{|~|-|-|-|-|}
\cline{1-5}\multicolumn{1}{c|}{\multirow{2}[4]{*}{$Z_4^\text{ \xmark}$}} & {\cellcolor[HTML]{C0C0C0}{0.0\%}} & {\cellcolor[HTML]{9B9B9B}{6.1\%}} & {\cellcolor[HTML]{656565}{\color[HTML]{FFFFFF}11.0\%}} & {\cellcolor[HTML]{9B9B9B}{2.0\%}} & \multicolumn{1}{c}{} \bigstrut\\ \hhline{|~|-|-|-|-|}
\cline{2-6}\multicolumn{1}{c|}{} & {\cellcolor[HTML]{EFEFEF}{0.0\%}} & {\cellcolor[HTML]{C0C0C0}{5.8\%}} & {\cellcolor[HTML]{9B9B9B}{2.0\%}} & {\cellcolor[HTML]{C0C0C0}{10.0\%}} & \multicolumn{1}{c}{$Z_3^\text{ \cmark}$} \bigstrut\\ \hhline{|~|-|-|-|-|}
\cline{2-5}\multicolumn{1}{r}{} & $Z_1^\text{ \cmark}$     & \multicolumn{2}{c|}{$Z_1^\text{ \xmark}$} & \multicolumn{1}{c}{$Z_1^\text{ \cmark}$} &  \bigstrut[t]\\
\end{tabular}%
}
\subcaption{Overall distribution of solutions.}
\end{subfigure}
\begin{subfigure}{0.4\linewidth}
\resizebox{\linewidth}{!}{%
\begin{tabular}{r|c|c|c|c|r}
\multicolumn{1}{c}{} & \multicolumn{2}{c|}{$Z_2^\text{ \cmark}$} & \multicolumn{2}{c}{$Z_2^\text{ \xmark}$} &  \bigstrut[b]\\ \hhline{|~|-|-|-|-|}
\cline{2-5}\multicolumn{1}{c|}{\multirow{2}[4]{*}{$Z_4^\text{ \cmark}$}} & 0\% & {\cellcolor[HTML]{EFEFEF}{0\%}} & {\cellcolor[HTML]{C0C0C0}{0\%}} & {\cellcolor[HTML]{EFEFEF}{0\%}} & \multicolumn{1}{c}{$Z_3^\text{ \cmark}$} \bigstrut\\ \hhline{|~|-|-|-|-|}
\cline{2-6}\multicolumn{1}{c|}{} & {\cellcolor[HTML]{EFEFEF}{0\%}} & {\cellcolor[HTML]{C0C0C0}{100\%}} & {\cellcolor[HTML]{9B9B9B}{100\%}} & {\cellcolor[HTML]{C0C0C0}{100\%}} & \multicolumn{1}{c}{\multirow{2}[4]{*}{$Z_3^\text{ \xmark}$}} \bigstrut\\ \hhline{|~|-|-|-|-|}
\cline{1-5}\multicolumn{1}{c|}{\multirow{2}[4]{*}{$Z_4^\text{ \xmark}$}} & {\cellcolor[HTML]{C0C0C0}{0\%}} & {\cellcolor[HTML]{9B9B9B}{100\%}} & {\cellcolor[HTML]{656565}{\color[HTML]{FFFFFF}100\%}} & {\cellcolor[HTML]{9B9B9B}{100\%}} & \multicolumn{1}{c}{} \bigstrut\\ \hhline{|~|-|-|-|-|}
\cline{2-6}\multicolumn{1}{c|}{} & {\cellcolor[HTML]{EFEFEF}{0\%}} & {\cellcolor[HTML]{C0C0C0}{100\%}} & {\cellcolor[HTML]{9B9B9B}{100\%}} & {\cellcolor[HTML]{C0C0C0}{100\%}} & \multicolumn{1}{c}{$Z_3^\text{ \cmark}$} \bigstrut\\ \hhline{|~|-|-|-|-|}
\cline{2-5}\multicolumn{1}{r}{} & $Z_1^\text{ \cmark}$     & \multicolumn{2}{c|}{$Z_1^\text{ \xmark}$} & \multicolumn{1}{c}{$Z_1^\text{ \cmark}$} &  \bigstrut[t]\\
\end{tabular}%
}
\subcaption{Frequency of instances.}
\end{subfigure}
\caption{Results for the trade-off regions analysis for Set D.}
\label{momkp:step3d}
\end{figure}

The global relationships are clear for sets $C$ (Figure \ref{momkp:step3c}) and $D$ (Figure \ref{momkp:step3d}).
There are no solutions with good values in all objectives and most instances present no solution with good values in three of the objectives. The majority of the solutions are situated where $Z_1$ and $Z_3$ alone have good values or where $Z_2$ and $Z_4$ alone have good values, as these are the harmonious pairs. Additionally, it can be observed that almost no solutions are present in conflicting areas. For instance, where $Z_1$ and $Z_2$ present good values simultaneously. 
Moreover, solutions in conflicting areas should be close to the chosen threshold.

\begin{figure}[!h]
\centering
\begin{subfigure}{0.4\linewidth}
\resizebox{\linewidth}{!}{%
\begin{tabular}{r|c|c|c|c|r}
\multicolumn{1}{c}{} & \multicolumn{2}{c|}{$Z_2^\text{ \cmark}$} & \multicolumn{2}{c}{$Z_2^\text{ \xmark}$} &  \bigstrut[b]\\ \hhline{|~|-|-|-|-|}
\cline{2-5}\multicolumn{1}{c|}{\multirow{2}[4]{*}{$Z_4^\text{ \cmark}$}} & 0.0\% & {\cellcolor[HTML]{EFEFEF}{0.0\%}} & {\cellcolor[HTML]{C0C0C0}{0.7\%}} & {\cellcolor[HTML]{EFEFEF}{0.0\%}} & \multicolumn{1}{c}{$Z_3^\text{ \cmark}$} \bigstrut\\ \hhline{|~|-|-|-|-|}
\cline{2-6}\multicolumn{1}{c|}{} & {\cellcolor[HTML]{EFEFEF}{0.0\%}} & {\cellcolor[HTML]{C0C0C0}{1.5\%}} & {\cellcolor[HTML]{9B9B9B}{5.3\%}} & {\cellcolor[HTML]{C0C0C0}{1.7\%}} & \multicolumn{1}{c}{\multirow{2}[4]{*}{$Z_3^\text{ \xmark}$}} \bigstrut\\ \hhline{|~|-|-|-|-|}
\cline{1-5}\multicolumn{1}{c|}{\multirow{2}[4]{*}{$Z_4^\text{ \xmark}$}} & {\cellcolor[HTML]{C0C0C0}{0.7\%}} & {\cellcolor[HTML]{9B9B9B}{3.9\%}} & {\cellcolor[HTML]{656565}{\color[HTML]{FFFFFF}70.5\%}} & {\cellcolor[HTML]{9B9B9B}{4.9\%}} & \multicolumn{1}{c}{} \bigstrut\\ \hhline{|~|-|-|-|-|}
\cline{2-6}\multicolumn{1}{c|}{} & {\cellcolor[HTML]{EFEFEF}{0.0\%}} & {\cellcolor[HTML]{C0C0C0}{1.6\%}} & {\cellcolor[HTML]{9B9B9B}{7.4\%}} & {\cellcolor[HTML]{C0C0C0}{1.7\%}} & \multicolumn{1}{c}{$Z_3^\text{ \cmark}$} \bigstrut\\ \hhline{|~|-|-|-|-|}
\cline{2-5}\multicolumn{1}{r}{} & $Z_1^\text{ \cmark}$     & \multicolumn{2}{c|}{$Z_1^\text{ \xmark}$} & \multicolumn{1}{c}{$Z_1^\text{ \cmark}$} &  \bigstrut[t]\\
\end{tabular}%
}
\subcaption{Overall distribution of solutions.}
\end{subfigure}
\begin{subfigure}{0.4\linewidth}
\resizebox{\linewidth}{!}{%
\begin{tabular}{r|c|c|c|c|r}
\multicolumn{1}{c}{} & \multicolumn{2}{c|}{$Z_2^\text{ \cmark}$} & \multicolumn{2}{c}{$Z_2^\text{ \xmark}$} &  \bigstrut[b]\\ \hhline{|~|-|-|-|-|}
\cline{2-5}\multicolumn{1}{c|}{\multirow{2}[4]{*}{$Z_4^\text{ \cmark}$}} & 0\% & {\cellcolor[HTML]{EFEFEF}{0\%}} & {\cellcolor[HTML]{C0C0C0}{80\%}} & {\cellcolor[HTML]{EFEFEF}{0\%}} & \multicolumn{1}{c}{$Z_3^\text{ \cmark}$} \bigstrut\\ \hhline{|~|-|-|-|-|}
\cline{2-6}\multicolumn{1}{c|}{} & {\cellcolor[HTML]{EFEFEF}{0\%}} & {\cellcolor[HTML]{C0C0C0}{100\%}} & {\cellcolor[HTML]{9B9B9B}{100\%}} & {\cellcolor[HTML]{C0C0C0}{100\%}} & \multicolumn{1}{c}{\multirow{2}[4]{*}{$Z_3^\text{ \xmark}$}} \bigstrut\\ \hhline{|~|-|-|-|-|}
\cline{1-5}\multicolumn{1}{c|}{\multirow{2}[4]{*}{$Z_4^\text{ \xmark}$}} & {\cellcolor[HTML]{C0C0C0}{100\%}} & {\cellcolor[HTML]{9B9B9B}{100\%}} & {\cellcolor[HTML]{656565}{\color[HTML]{FFFFFF}100\%}} & {\cellcolor[HTML]{9B9B9B}{100\%}} & \multicolumn{1}{c}{} \bigstrut\\ \hhline{|~|-|-|-|-|}
\cline{2-6}\multicolumn{1}{c|}{} & {\cellcolor[HTML]{EFEFEF}{0\%}} & {\cellcolor[HTML]{C0C0C0}{100\%}} & {\cellcolor[HTML]{9B9B9B}{100\%}} & {\cellcolor[HTML]{C0C0C0}{100\%}} & \multicolumn{1}{c}{$Z_3^\text{ \cmark}$} \bigstrut\\ \hhline{|~|-|-|-|-|}
\cline{2-5}\multicolumn{1}{r}{} & $Z_1^\text{ \cmark}$     & \multicolumn{2}{c|}{$Z_1^\text{ \xmark}$} & \multicolumn{1}{c}{$Z_1^\text{ \cmark}$} &  \bigstrut[t]\\
\end{tabular}%
}
\subcaption{Frequency of instances.}
\end{subfigure}
\caption{Results for the trade-off regions analysis for Set X.}
\label{momkp:step3e}
\end{figure}

The set $X$ (Figure \ref{momkp:step3e}) does not contain solutions in $r_0$ and there are no solutions in regions $r_1$, $r_2$, $r_4$ and $r_8$, meaning that no good values can be simultaneously found for three or more objectives. Good values can only be found simultaneously for up to two objectives. The map for set $X$ resembles the ones for sets $C$ and $D$ in the sense that it is clear that there are several regions without solutions. Thus, there are trade-offs to present to the decision-maker. This means that the decision-maker has to choose between up to two good objective values to the detriment of the remaining objectives, since all of the regions containing solutions have at most two simultaneously good values.

\begin{figure*}[!h]
\centering
\begin{subfigure}[b]{0.31\textwidth}
\centering
\includegraphics[width=\linewidth]{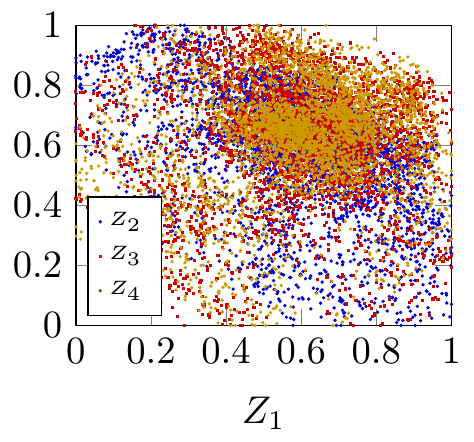}
\caption{Set A}
\label{momkp:step4a}
\end{subfigure}
\begin{subfigure}[b]{0.31\textwidth}
\centering
\includegraphics[width=\linewidth]{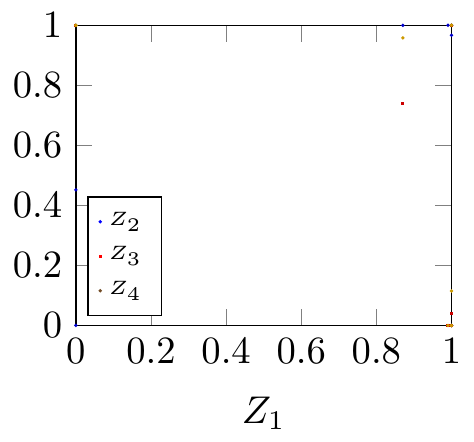}
\caption{Set B}
\label{momkp:step4b}
\end{subfigure}
\begin{subfigure}[b]{0.31\textwidth}
\centering
\includegraphics[width=\linewidth]{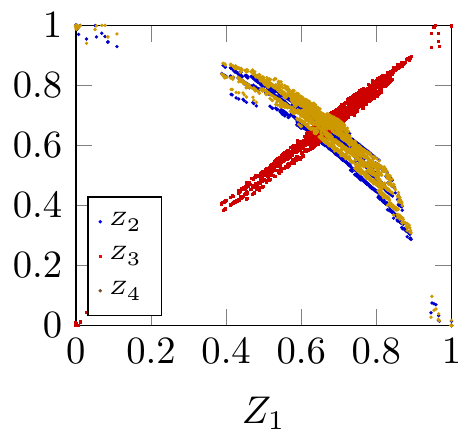}
\caption{Set C}
\label{momkp:step4c}
\end{subfigure}
\begin{subfigure}[b]{0.31\textwidth}
\centering
\includegraphics[width=\linewidth]{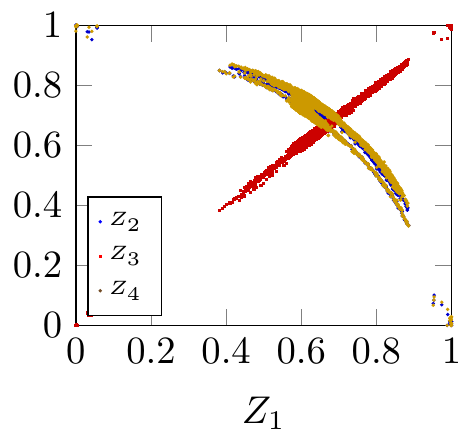}
\caption{Set D}
\label{momkp:step4d}
\end{subfigure}
\begin{subfigure}[b]{0.31\textwidth}
\centering
\includegraphics[width=\linewidth]{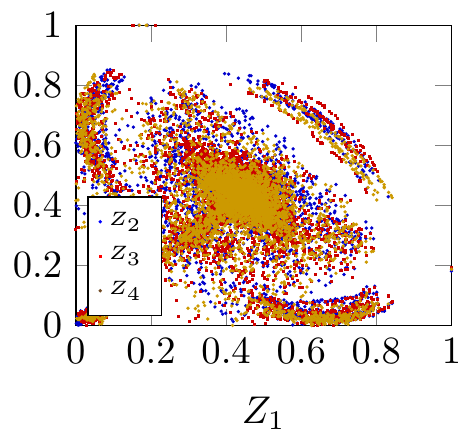}
\caption{Set X}
\label{momkp:step4x}
\end{subfigure}
\caption{\hl{Scatter plots of each of the MOMKP sets, showing the relationship between objective $Z_1$ and each of the other three objectives. In each case, the $x$ axis shows the value of objective $Z_1$ and the $y$ axis shows the values of each of the other objectives in different colours.}}
\label{momkp:step4}
\end{figure*}

\textbf{Step 4 - Multiobjective Scatter Plot Analysis.} This analysis was performed for each instance and Figure \ref{momkp:step4} presents the results for all instances of each set combined. It can be seen that, although instances in dataset $A$ are completely random, they show similar landscapes with a high concentration of solutions towards the $(1,1)$ corner. Moreover, no local relationships can be identified, which is expected as the data is completely random.

The trade-off regions are clear for sets $C$ and $D$. There is also a noticeable gap in the objective space when $Z_1$ is in the range from 0 to 0.5 and when the remaining objectives are in the range from 0 to 0.4, approximately. Moreover, the landscape of the objective space appears to be similar for all instances of each of the sets $C$ and $D$.

Since the data was uniformly generated, these gaps are unlikely to arise from the data itself, and could represent limitations in the solution algorithms, indicating that they did not explore the entire front. It is well known that the performance of some MOEAs is limited when the number of objectives is more than three \citep{Giagkiozis2012}.

Set $X$ presents a unique scenario and patterns and gaps can be identified in the objective space. 
There is a lack of solutions with values within $[0.85,1]$. This is again likely to be due to limitations of the solution algorithms. However, it can be seen that the size of the gap is small, confirming that instances with strongly conflicting objectives present a bigger challenge for these algorithms. 
Several local relationships can also be identified. When $Z_1$ ranges from 0.5 to 0.8, the three remaining objectives simultaneously conflict and harmonise. Knowing that only two simultaneous objectives present high values (from the region map analysis), it can be concluded that whenever $Z_1$ increases, only one of the other objectives simultaneously increases too.

\subsubsection{Discussion}
\label{discussion}

%cut candidate
The proposed analysis and visualisation technique can aid in understanding the multiobjective nature of the problem instances considered. Looking at only the correlation coefficients, it could be concluded that: sets $A$ and $X$ do not present interesting multiobjective traits, that set $B$ is inconclusive and that sets $C$ and $D$ present conflicting and harmonious objectives. However, by applying this analysis and visualisation technique,  a more comprehensive understanding of these instance sets can be gained.

As fully random instances, dataset $A$ does not present relevant global or local pairwise relationships according to the global pairwise analysis (step 1) and the multiobjective scatter plot analysis (step 4). Additionally, the objective range analysis (step 2) shows that even though there is a large set of non-dominated solutions, these are concentrated in a reasonably small area of the search space. For this dataset, the information from the trade-off region maps can be used to interact with the decision-maker to identify which regions are of more interest and then use single-objective optimisation algorithms to find solutions in that region. Since there are solutions in all of the regions of the map, any objective vector could provide an adequate solution.

Set $B$ presents a completely harmonious case and by analysing the ranges and bearing in mind that the algorithms found just a handful of solutions, it can be concluded that a single-objective algorithm aiming to maximise any of the objectives could provide a good solution.

Sets $C$ and $D$ present similar scenarios, hence the correlation between weights and coefficients does not impact on the nature of the problem. The entire solution set represents a huge trade-off. The algorithms were not successful in expanding along the front and they mainly explored the region surrounding the intersection of the trade-off. Nonetheless, by perceiving that all instances in these sets have similar landscapes and by knowing the approximate boundaries of each objective (by applying single-objective algorithms to each objective alone), the landscape of solutions could be estimated for other instances in those sets. The search could then be directed to the regions of interest after presenting the expected trade-offs to the decision-maker. However, if it is imperative to use an \textit{a posteriori} approach, the global pairwise analysis and the scatter plots provide sufficient information to make feasible the grouping of harmonious objectives.

Finally, set $X$ presents a quite different picture. By only evaluating the global pairwise analysis (step 1) it could be concluded that there is no strong pairwise relationship between objectives. However, the objectives range analysis (step 2) shows that in fact there are non-dominated solutions that vary greatly in quality. This is an indication of the existence of trade-offs (as can be seen by comparing this set with sets $C$ and $D$). The trade-off region analysis (step 3) showed the existence of overall trade-offs as it is not possible to have solutions with good values in more than two objectives simultaneously. 
Finally, the multiobjective scatter plot analysis (step 4) identified local relationships between objectives and gaps in the objective space, pointing to the existence of local conflicts. Therefore, instances in dataset $X$ exhibit a distinctive multiobjective nature perhaps with interesting options for a decision-maker. A sound possibility to tackle this problem would be to use the region map to identify the regions of interest and then locate those regions in the scatter plot. In cases where a selected region contains a local conflict, the algorithm proposed by \citep{Knowles2002} could be used to reach the trade-off front and then expand through it.

\subsection{Multiobjective Nurse Scheduling Problem}

To assess the MONSP, the instances were initially grouped into the three sets described above: IS, ODS and RDS. 
However, meaningful dissimilarities were not obvious between the sets, because they presented similar fitness landscapes, meaning that the process to generate the instances did not create sufficiently different scenarios. However, after evaluating the data, a different grouping of instances was identified according to case files:
\begin{itemize}
\item \textbf{Set A:} Instances built using the case files 1, 2, 3, 4 and 6.
\item \textbf{Set B:} Instances built using the case files 5, 7 and 8.
\end{itemize}
% Rodrigo - I changes cases to group, since cases sounds odd, and you seem to refer to these as cases or sets interchangably later.

\begin{figure}[!ht]
\centering
\resizebox{0.4\textwidth}{!}{%
\begin{tikzpicture}
\begin{axis}[
legend style={at={(0.1,1.01)},anchor=south west,legend columns=3},
xticklabels={{\small $Z_1-Z_2$},{\small $Z_1-Z_3$},{\small $Z_1-Z_4$},{\small $Z_2-Z_3$},{\small $Z_2-Z_4$},{\small $Z_3-Z_4$}},
xtick=data,
legend pos=north west,
scaled y ticks = false,
x tick label style={rotate=45, anchor=east},
ymax=1,
ymin=-1,
xmin=0,
xmax=5,
cycle list name=color list,
]

\addplot coordinates
{
(0,0.013)
(1,-0.146)
(2,-0.286)
(3,-0.053)
(4,-0.250)
(5,-0.464)
};

\addplot coordinates
{
(0,-0.032)
(1,0.081)
(2,-0.405)
(3,0.045)
(4,-0.339)
(5,-0.492)
};

\addplot coordinates
{
(0,-0.006)
(1,-0.037)
(2,-0.350)
(3,0.002)
(4,-0.321)
(5,-0.420)
};

\addplot[color=black,dashed] coordinates
{
(0,-0.014)
(1,-0.244)
(2,-0.173)
(3,-0.131)
(4,-0.311)
(5,-0.479)
};

\addplot[color=red,dashed] coordinates
{
(0,0.051)
(1,-0.065)
(2,-0.373)
(3,0.000)
(4,-0.403)
(5,-0.319)
};

\addplot[] coordinates
{
(0,0)
(1,0)
(2,0)
(3,0)
(4,0)
(5,0)
(6,0)
(7,0)
(8,0)
(9,0)
};

\legend{IS,ODS,RDS,Cases A, Cases B}
\end{axis}
\end{tikzpicture}
}
\caption{Pairwise correlations for each set of instances.}
\label{nsp:step1}
\end{figure}
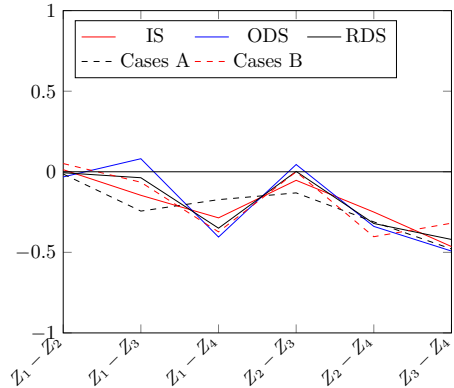

The results of the analysis for these new sets A and B are presented here, along with the results for the original sets (IS, ODS and RDS) to validate the claim regarding their similarities.

\textbf{Step 1 - Global Pairwise Relationships Analysis.} Figure \ref{nsp:step1} presents the results for the pairwise relationships analysis. Note that all solid lines representing the generated sets have similar values on all pairs of objectives. Now it is visible that set B approximately follows the curve of the generational sets, however, set A slightly diverges for ($Z_1-Z_3$), ($Z_1-Z_4$) and ($Z_2-Z_3$). Nonetheless, it is clear that the differences are minimal and that no set presents strong global pairwise relationships.

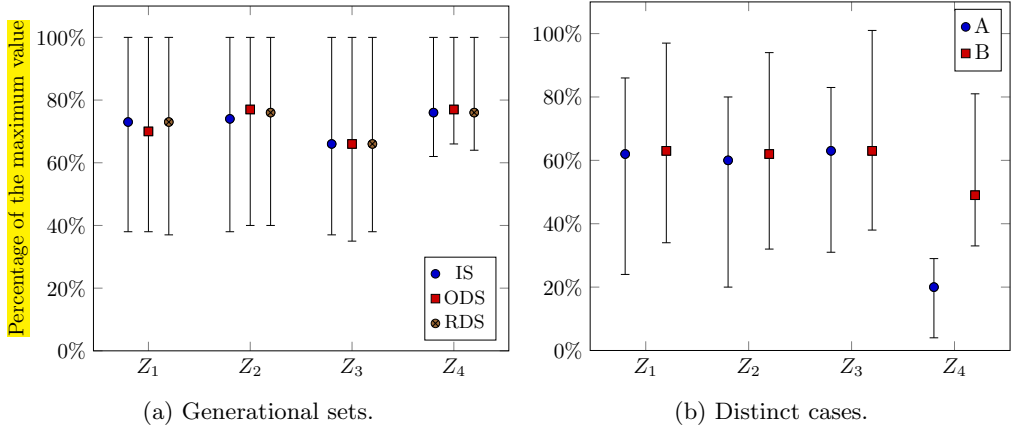
\begin{figure}[!h]
\centering
\begin{subfigure}[b]{0.44\textwidth}
\centering
\resizebox{\textwidth}{!}{%
\begin{tikzpicture}
\begin{axis} [symbolic x coords={0,$Z_1$,1,2,2.5,3,$Z_2$,4,5,5.5,6,$Z_3$,7,8,8.5,9,$Z_4$,10},
legend pos=south east,
yticklabel={\pgfmathparse{\tick*100}\pgfmathprintnumber{\pgfmathresult}\%},
    point meta={x*100},ymin=0,ymax=1.1,xtick=data,
scaled ticks=false, tick label style={/pgf/number format/fixed},
xtick={$Z_1$,$Z_2$,$Z_3$,$Z_4$},
ylabel=\hl{Percentage of the maximum value}
]
\addplot+[forget plot,only marks] plot[color=black,error bars/.cd, y dir=plus, y explicit]
coordinates{
(0,0.73)+-(0.27,0.27)
(3,0.74)+-(0.26,0.26)
(6,0.66)+-(0.34,0.34)
(9,0.76)+-(0.24,0.24)
};
\addplot+[only marks] plot[color=black,error bars/.cd, y dir=minus, y explicit]
coordinates{
(0,0.73)+-(0.35,0.35)
(3,0.74)+-(0.36,0.36)
(6,0.66)+-(0.29,0.29)
(9,0.76)+-(0.14,0.14)
};
\addplot+[color=black,forget plot,only marks] plot[color=black,error bars/.cd, y dir=plus, y explicit]
coordinates{
($Z_1$,0.70)+-(0.30,0.30)
($Z_2$,0.77)+-(0.23,0.23)
($Z_3$,0.66)+-(0.34,0.34)
($Z_4$,0.77)+-(0.23,0.23)
};
\addplot+[only marks] plot[color=black,error bars/.cd, y dir=minus, y explicit]
coordinates{
($Z_1$,0.70)+-(0.32,0.32)
($Z_2$,0.77)+-(0.37,0.37)
($Z_3$,0.66)+-(0.31,0.31)
($Z_4$,0.77)+-(0.11,0.11)
};
\addplot+[forget plot,only marks] plot[color=black,error bars/.cd, y dir=plus, y explicit]
coordinates{
(1,0.73)+-(0.27,0.27)
(4,0.76)+-(0.24,0.24)
(7,0.66)+-(0.34,0.34)
(10,0.76)+-(0.24,0.24)
};
\addplot+[only marks] plot[color=black,error bars/.cd, y dir=minus, y explicit]
coordinates{
(1,0.73)+-(0.36,0.36)
(4,0.76)+-(0.36,0.36)
(7,0.66)+-(0.28,0.28)
(10,0.76)+-(0.12,0.12)
};
\legend{{\small IS},{\small ODS},{\small RDS}}
\end{axis} 
\end{tikzpicture}
}
\caption{Generational sets.}
\label{nsp:step2a}
\end{subfigure}
\begin{subfigure}[b]{0.42\textwidth}
\centering
\resizebox{\textwidth}{!}{%
\begin{tikzpicture}
\begin{axis} [symbolic x coords={0,$Z_1$,1,2,2.5,3,$Z_2$,4,5,5.5,6,$Z_3$,7,8,8.5,9,$Z_4$,10},
yticklabel={\pgfmathparse{\tick*100}\pgfmathprintnumber{\pgfmathresult}\%},
    point meta={x*100},ymin=0,ymax=1.1,xtick=data,
scaled ticks=false, tick label style={/pgf/number format/fixed},
xtick={$Z_1$,$Z_2$,$Z_3$,$Z_4$},
]
\addplot+[forget plot,only marks] plot[color=black,error bars/.cd, y dir=plus, y explicit]
coordinates{
(0,0.62)+-(0.24,0.24)
(3,0.60)+-(0.20,0.20)
(6,0.63)+-(0.20,0.20)
(9,0.20)+-(0.09,0.09)
};
\addplot+[only marks] plot[color=black,error bars/.cd, y dir=minus, y explicit]
coordinates{
(0,0.62)+-(0.38,0.38)
(3,0.60)+-(0.40,0.40)
(6,0.63)+-(0.32,0.32)
(9,0.20)+-(0.16,0.16)
};
\addplot+[forget plot,only marks] plot[color=black,error bars/.cd, y dir=plus, y explicit]
coordinates{
(1,0.63)+-(0.34,0.34)
(4,0.62)+-(0.32,0.32)
(7,0.63)+-(0.38,0.38)
(10,0.49)+-(0.32,0.32)
};
\addplot+[only marks] plot[color=black,error bars/.cd, y dir=minus, y explicit]
coordinates{
(1,0.63)+-(0.29,0.29)
(4,0.62)+-(0.30,0.30)
(7,0.63)+-(0.25,0.25)
(10,0.49)+-(0.16,0.16)
};
\legend{A,B}
\end{axis} 
\end{tikzpicture}
}
\caption{Distinct cases.}
\label{nsp:step2b}
\end{subfigure}
\caption{Range analysis results for both the generational sets and the distinct cases. In each line, the central point is the average value for the set, with the top and bottom of the line representing the maximum and minimum values found, respectively.}
\label{nsp:step2}
\end{figure}

\textbf{Step 2 - Objective Range Analysis.} In the objective range analysis the differences between the generated sets and the new sets A and B becomes evident. Figure \ref{nsp:step2a} compares the ranges between the generated sets. \hl{The $x$ axis presents the sets grouped by objective. The $y$ axis presents the minimum, maximum and average value of each objective as a percentage of the overall maximum value found for the respective objective}. The similarity across all sets in the ranges and average values for all objectives is apparent. However, Figure \ref{nsp:step2b} shows that although the average values and ranges are equivalent between sets A and B on $Z_1$, $Z_2$ and $Z_3$, the maximum and minimum values found for set A are roughly 10\% lower than for set B. Additionally, for objective $Z_4$ both ranges and average values are noticeably smaller for set A. It can be concluded that grouping the instances based on the generated sets did not present any meaningful differences. However, considering the new sets A and B, meaningful differences are apparent. Therefore, from this point on, the analysis considers these case sets.  

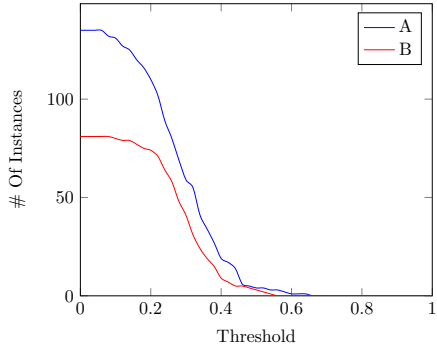
\begin{figure}[!ht]
\centering
\resizebox{0.38\textwidth}{!}{%
\begin{tikzpicture}
	\begin{axis}[
		xlabel=Threshold,
		ylabel=\# Of Instances,
		xmin=0,
		xmax=1,
		legend pos= north east,	
		ymin=0,
		]
	\addplot[color=blue,smooth] coordinates {
(	0	,	135	)
(	0.02	,	135	)
(	0.04	,	135	)
(	0.06	,	135	)
(	0.08	,	132	)
(	0.1	,	131	)
(	0.12	,	127	)
(	0.14	,	125	)
(	0.16	,	120	)
(	0.18	,	116	)
(	0.2	,	110	)
(	0.22	,	102	)
(	0.24	,	89	)
(	0.26	,	80	)
(	0.28	,	69	)
(	0.3	,	59	)
(	0.32	,	55	)
(	0.34	,	41	)
(	0.36	,	34	)
(	0.38	,	27	)
(	0.4	,	19	)
(	0.42	,	17	)
(	0.44	,	14	)
(	0.46	,	6	)
(	0.48	,	5	)
(	0.5	,	4	)
(	0.52	,	4	)
(	0.54	,	3	)
(	0.56	,	3	)
(	0.58	,	2	)
(	0.6	,	1	)
(	0.62	,	1	)
(	0.64	,	1	)
(	0.66	,	0	)
(	0.68	,	0	)
(	0.7	,	0	)
(	0.72	,	0	)
(	0.74	,	0	)
(	0.76	,	0	)
(	0.78	,	0	)
(	0.8	,	0	)
(	0.82	,	0	)
(	0.84	,	0	)
(	0.86	,	0	)
(	0.88	,	0	)
(	0.9	,	0	)
(	0.92	,	0	)
(	0.94	,	0	)
(	0.96	,	0	)
(	0.98	,	0	)
(	1	,	0	)
	};	
	
	\addplot[color=red,smooth] coordinates {(	0	,	81	)
(	0.02	,	81	)
(	0.04	,	81	)
(	0.06	,	81	)
(	0.08	,	81	)
(	0.1	,	80	)
(	0.12	,	79	)
(	0.14	,	79	)
(	0.16	,	77	)
(	0.18	,	75	)
(	0.2	,	74	)
(	0.22	,	71	)
(	0.24	,	64	)
(	0.26	,	58	)
(	0.28	,	48	)
(	0.3	,	41	)
(	0.32	,	31	)
(	0.34	,	24	)
(	0.36	,	19	)
(	0.38	,	15	)
(	0.4	,	9	)
(	0.42	,	7	)
(	0.44	,	5	)
(	0.46	,	5	)
(	0.48	,	4	)
(	0.5	,	3	)
(	0.52	,	2	)
(	0.54	,	1	)
(	0.56	,	0	)
(	0.58	,	0	)
(	0.6	,	0	)
(	0.62	,	0	)
(	0.64	,	0	)
(	0.66	,	0	)
(	0.68	,	0	)
(	0.7	,	0	)
(	0.72	,	0	)
(	0.74	,	0	)
(	0.76	,	0	)
(	0.78	,	0	)
(	0.8	,	0	)
(	0.82	,	0	)
(	0.84	,	0	)
(	0.86	,	0	)
(	0.88	,	0	)
(	0.9	,	0	)
(	0.92	,	0	)
(	0.94	,	0	)
(	0.96	,	0	)
(	0.98	,	0	)
(	1	,	0	)
};	
\legend{A,B}
	\end{axis}
\end{tikzpicture}
}
\caption{Threshold analysis showing the number of scenarios with solutions found in region $r_0$ when the threshold decreases.}
\label{nsp:step3a}
\end{figure}

\textbf{Step 3 - Trade-Off Region Analysis.} The results of the trade-off region analysis for sets A and B are now presented. The threshold analysis is shown in Figure \ref{nsp:step3a}, where both curves can be seen to be similar and to converge to roughly the same point, showing that in both cases no solution in $r_0$ can be found with more than 60\% of the maximum values. This means that to obtain an objective with more than 60\% quality, at least one other objective will present an inferior value. 

\begin{figure}[!ht]
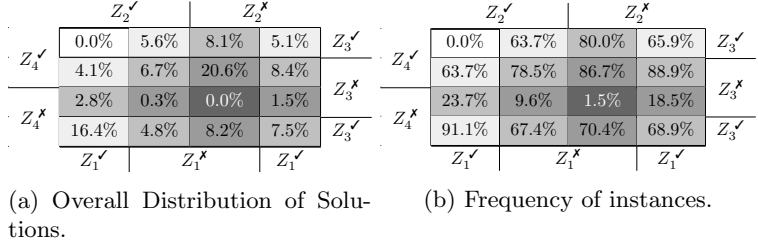

\centering
\begin{subfigure}[t]{0.31\linewidth}
\resizebox{\linewidth}{!}{%
\begin{tabular}{r|c|c|c|c|r}
\multicolumn{1}{c}{} & \multicolumn{2}{c|}{$Z_2^\text{ \cmark}$} & \multicolumn{2}{c}{$Z_2^\text{ \xmark}$} &  \bigstrut[b]\\ \hhline{|~|-|-|-|-|}
\cline{2-5}\multicolumn{1}{c|}{\multirow{2}[4]{*}{$Z_4^\text{ \cmark}$}} & 0.0\% & {\cellcolor[HTML]{EFEFEF}{5.6\%}} & {\cellcolor[HTML]{C0C0C0}{8.1\%}} & {\cellcolor[HTML]{EFEFEF}{5.1\%}} & \multicolumn{1}{c}{$Z_3^\text{ \cmark}$} \bigstrut\\ \hhline{|~|-|-|-|-|}
\cline{2-6}\multicolumn{1}{c|}{} & {\cellcolor[HTML]{EFEFEF}{4.1\%}} & {\cellcolor[HTML]{C0C0C0}{6.7\%}} & {\cellcolor[HTML]{9B9B9B}{20.6\%}} & {\cellcolor[HTML]{C0C0C0}{8.4\%}} & \multicolumn{1}{c}{\multirow{2}[4]{*}{$Z_3^\text{ \xmark}$}} \bigstrut\\ \hhline{|~|-|-|-|-|}
\cline{1-5}\multicolumn{1}{c|}{\multirow{2}[4]{*}{$Z_4^\text{ \xmark}$}} & {\cellcolor[HTML]{C0C0C0}{2.8\%}} & {\cellcolor[HTML]{9B9B9B}{0.3\%}} & {\cellcolor[HTML]{656565}{\color[HTML]{FFFFFF}0.0\%}} & {\cellcolor[HTML]{9B9B9B}{1.5\%}} & \multicolumn{1}{c}{} \bigstrut\\ \hhline{|~|-|-|-|-|}
\cline{2-6}\multicolumn{1}{c|}{} & {\cellcolor[HTML]{EFEFEF}{16.4\%}} & {\cellcolor[HTML]{C0C0C0}{4.8\%}} & {\cellcolor[HTML]{9B9B9B}{8.2\%}} & {\cellcolor[HTML]{C0C0C0}{7.5\%}} & \multicolumn{1}{c}{$Z_3^\text{ \cmark}$} \bigstrut\\ \hhline{|~|-|-|-|-|}
\cline{2-5}\multicolumn{1}{r}{} & $Z_1^\text{ \cmark}$     & \multicolumn{2}{c|}{$Z_1^\text{ \xmark}$} & \multicolumn{1}{c}{$Z_1^\text{ \cmark}$} &  \bigstrut[t]\\
\end{tabular}%
}
\subcaption{Overall Distribution of Solutions.}
\end{subfigure}
\begin{subfigure}[t]{0.32\linewidth}
\resizebox{\linewidth}{!}{%
\begin{tabular}{r|c|c|c|c|r}
\multicolumn{1}{c}{} & \multicolumn{2}{c|}{$Z_2^\text{ \cmark}$} & \multicolumn{2}{c}{$Z_2^\text{ \xmark}$} &  \bigstrut[b]\\ \hhline{|~|-|-|-|-|}
\cline{2-5}\multicolumn{1}{c|}{\multirow{2}[4]{*}{$Z_4^\text{ \cmark}$}} & 0.0\% & {\cellcolor[HTML]{EFEFEF}{63.7\%}} & {\cellcolor[HTML]{C0C0C0}{80.0\%}} & {\cellcolor[HTML]{EFEFEF}{65.9\%}} & \multicolumn{1}{c}{$Z_3^\text{ \cmark}$} \bigstrut\\ \hhline{|~|-|-|-|-|}
\cline{2-6}\multicolumn{1}{c|}{} & {\cellcolor[HTML]{EFEFEF}{63.7\%}} & {\cellcolor[HTML]{C0C0C0}{78.5\%}} & {\cellcolor[HTML]{9B9B9B}{86.7\%}} & {\cellcolor[HTML]{C0C0C0}{88.9\%}} & \multicolumn{1}{c}{\multirow{2}[4]{*}{$Z_3^\text{ \xmark}$}} \bigstrut\\ \hhline{|~|-|-|-|-|}
\cline{1-5}\multicolumn{1}{c|}{\multirow{2}[4]{*}{$Z_4^\text{ \xmark}$}} & {\cellcolor[HTML]{C0C0C0}{23.7\%}} & {\cellcolor[HTML]{9B9B9B}{9.6\%}} & {\cellcolor[HTML]{656565}{\color[HTML]{FFFFFF}1.5\%}} & {\cellcolor[HTML]{9B9B9B}{18.5\%}} & \multicolumn{1}{c}{} \bigstrut\\ \hhline{|~|-|-|-|-|}
\cline{2-6}\multicolumn{1}{c|}{} & {\cellcolor[HTML]{EFEFEF}{91.1\%}} & {\cellcolor[HTML]{C0C0C0}{67.4\%}} & {\cellcolor[HTML]{9B9B9B}{70.4\%}} & {\cellcolor[HTML]{C0C0C0}{68.9\%}} & \multicolumn{1}{c}{$Z_3^\text{ \cmark}$} \bigstrut\\ \hhline{|~|-|-|-|-|}
\cline{2-5}\multicolumn{1}{r}{} & $Z_1^\text{ \cmark}$     & \multicolumn{2}{c|}{$Z_1^\text{ \xmark}$} & \multicolumn{1}{c}{$Z_1^\text{ \cmark}$} &  \bigstrut[t]\\
\end{tabular}%
}
\subcaption{Frequency of instances.}
\end{subfigure}
\caption{Region maps for Cases A.}
\label{nsp:step3b}
\end{figure}

\begin{figure}[!h]
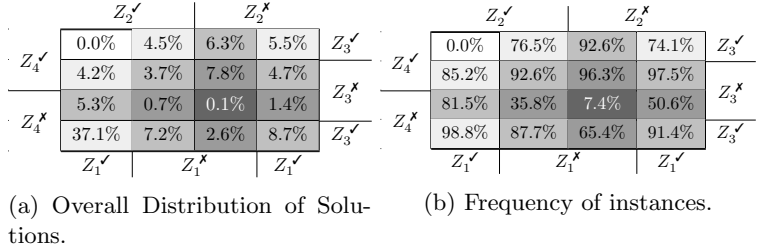

\centering
\begin{subfigure}[t]{0.31\linewidth}
\resizebox{\linewidth}{!}{%
\begin{tabular}{r|c|c|c|c|r}
\multicolumn{1}{c}{} & \multicolumn{2}{c|}{$Z_2^\text{ \cmark}$} & \multicolumn{2}{c}{$Z_2^\text{ \xmark}$} &  \bigstrut[b]\\ \hhline{|~|-|-|-|-|}
\cline{2-5}\multicolumn{1}{c|}{\multirow{2}[4]{*}{$Z_4^\text{ \cmark}$}} & 0.0\% & {\cellcolor[HTML]{EFEFEF}{4.5\%}} & {\cellcolor[HTML]{C0C0C0}{6.3\%}} & {\cellcolor[HTML]{EFEFEF}{5.5\%}} & \multicolumn{1}{c}{$Z_3^\text{ \cmark}$} \bigstrut\\ \hhline{|~|-|-|-|-|}
\cline{2-6}\multicolumn{1}{c|}{} & {\cellcolor[HTML]{EFEFEF}{4.2\%}} & {\cellcolor[HTML]{C0C0C0}{3.7\%}} & {\cellcolor[HTML]{9B9B9B}{7.8\%}} & {\cellcolor[HTML]{C0C0C0}{4.7\%}} & \multicolumn{1}{c}{\multirow{2}[4]{*}{$Z_3^\text{ \xmark}$}} \bigstrut\\ \hhline{|~|-|-|-|-|}
\cline{1-5}\multicolumn{1}{c|}{\multirow{2}[4]{*}{$Z_4^\text{ \xmark}$}} & {\cellcolor[HTML]{C0C0C0}{5.3\%}} & {\cellcolor[HTML]{9B9B9B}{0.7\%}} & {\cellcolor[HTML]{656565}{\color[HTML]{FFFFFF}0.1\%}} & {\cellcolor[HTML]{9B9B9B}{1.4\%}} & \multicolumn{1}{c}{} \bigstrut\\ \hhline{|~|-|-|-|-|}
\cline{2-6}\multicolumn{1}{c|}{} & {\cellcolor[HTML]{EFEFEF}{37.1\%}} & {\cellcolor[HTML]{C0C0C0}{7.2\%}} & {\cellcolor[HTML]{9B9B9B}{2.6\%}} & {\cellcolor[HTML]{C0C0C0}{8.7\%}} & \multicolumn{1}{c}{$Z_3^\text{ \cmark}$} \bigstrut\\ \hhline{|~|-|-|-|-|}
\cline{2-5}\multicolumn{1}{r}{} & $Z_1^\text{ \cmark}$     & \multicolumn{2}{c|}{$Z_1^\text{ \xmark}$} & \multicolumn{1}{c}{$Z_1^\text{ \cmark}$} &  \bigstrut[t]\\
\end{tabular}%
}
\subcaption{Overall Distribution of Solutions.}
\end{subfigure}
\begin{subfigure}[t]{0.32\linewidth}
\resizebox{\linewidth}{!}{%
\begin{tabular}{r|c|c|c|c|r}
\multicolumn{1}{c}{} & \multicolumn{2}{c|}{$Z_2^\text{ \cmark}$} & \multicolumn{2}{c}{$Z_2^\text{ \xmark}$} &  \bigstrut[b]\\ \hhline{|~|-|-|-|-|}
\cline{2-5}\multicolumn{1}{c|}{\multirow{2}[4]{*}{$Z_4^\text{ \cmark}$}} & 0.0\% & {\cellcolor[HTML]{EFEFEF}{76.5\%}} & {\cellcolor[HTML]{C0C0C0}{92.6\%}} & {\cellcolor[HTML]{EFEFEF}{74.1\%}} & \multicolumn{1}{c}{$Z_3^\text{ \cmark}$} \bigstrut\\ \hhline{|~|-|-|-|-|}
\cline{2-6}\multicolumn{1}{c|}{} & {\cellcolor[HTML]{EFEFEF}{85.2\%}} & {\cellcolor[HTML]{C0C0C0}{92.6\%}} & {\cellcolor[HTML]{9B9B9B}{96.3\%}} & {\cellcolor[HTML]{C0C0C0}{97.5\%}} & \multicolumn{1}{c}{\multirow{2}[4]{*}{$Z_3^\text{ \xmark}$}} \bigstrut\\ \hhline{|~|-|-|-|-|}
\cline{1-5}\multicolumn{1}{c|}{\multirow{2}[4]{*}{$Z_4^\text{ \xmark}$}} & {\cellcolor[HTML]{C0C0C0}{81.5\%}} & {\cellcolor[HTML]{9B9B9B}{35.8\%}} & {\cellcolor[HTML]{656565}{\color[HTML]{FFFFFF}7.4\%}} & {\cellcolor[HTML]{9B9B9B}{50.6\%}} & \multicolumn{1}{c}{} \bigstrut\\ \hhline{|~|-|-|-|-|}
\cline{2-6}\multicolumn{1}{c|}{} & {\cellcolor[HTML]{EFEFEF}{98.8\%}} & {\cellcolor[HTML]{C0C0C0}{87.7\%}} & {\cellcolor[HTML]{9B9B9B}{65.4\%}} & {\cellcolor[HTML]{C0C0C0}{91.4\%}} & \multicolumn{1}{c}{$Z_3^\text{ \cmark}$} \bigstrut\\ \hhline{|~|-|-|-|-|}
\cline{2-5}\multicolumn{1}{r}{} & $Z_1^\text{ \cmark}$     & \multicolumn{2}{c|}{$Z_1^\text{ \xmark}$} & \multicolumn{1}{c}{$Z_1^\text{ \cmark}$} &  \bigstrut[t]\\
\end{tabular}%
}
\subcaption{Frequency of instances.}
\end{subfigure}
\caption{Region maps for Cases B.}
\label{nsp:step3c}
\end{figure}

% Rodrigo: What threshold values are these shown for? I think it matters doesn't it?
The region maps for sets A and B are shown in Figures \ref{nsp:step3b} and \ref{nsp:step3c}.
Sets A and B are similar regarding the regions where solutions can be found, but some differences can be observed. The first difference is that the frequency of solutions is higher in most regions for set B, meaning that the fitness landscapes of instances in B are more alike than the instances in A. In both cases, solutions can be found in all regions of the map, as for the MOMKP independent set (Figure \ref{momkp:step3b}). The difference, however, is that in the independent MOMKP set, there were 100\% frequencies in most regions while here the values vary greatly. This means that the instances share less in common between themselves than in the independent MOMKP case.

\begin{figure}[!h]
\centering
\begin{subfigure}[b]{0.24\textwidth}
\centering
\includegraphics[width=\linewidth]{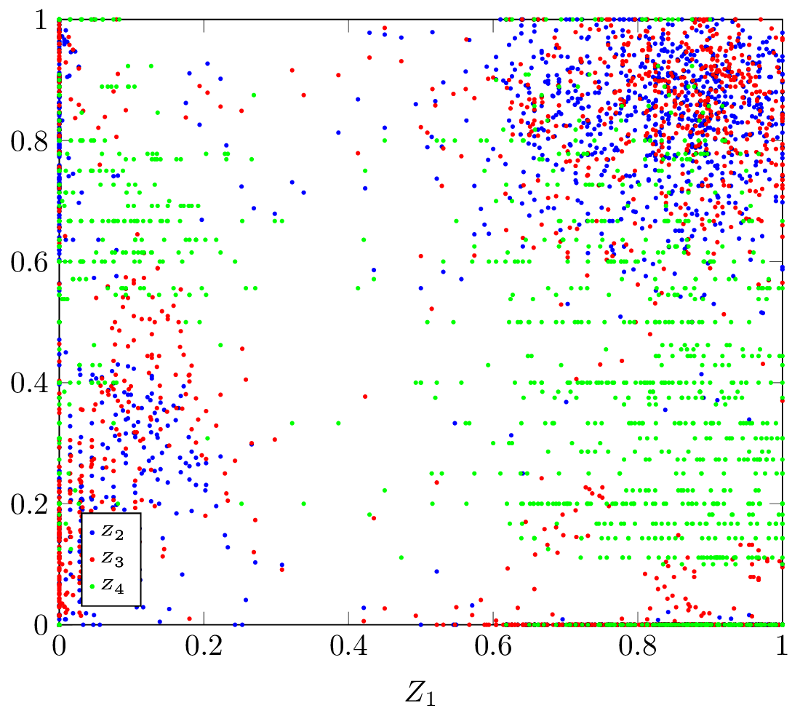}
\caption{Case file 1}
\label{nsp:step41}
\end{subfigure}
\begin{subfigure}[b]{0.24\textwidth}
\centering
\includegraphics[width=\linewidth]{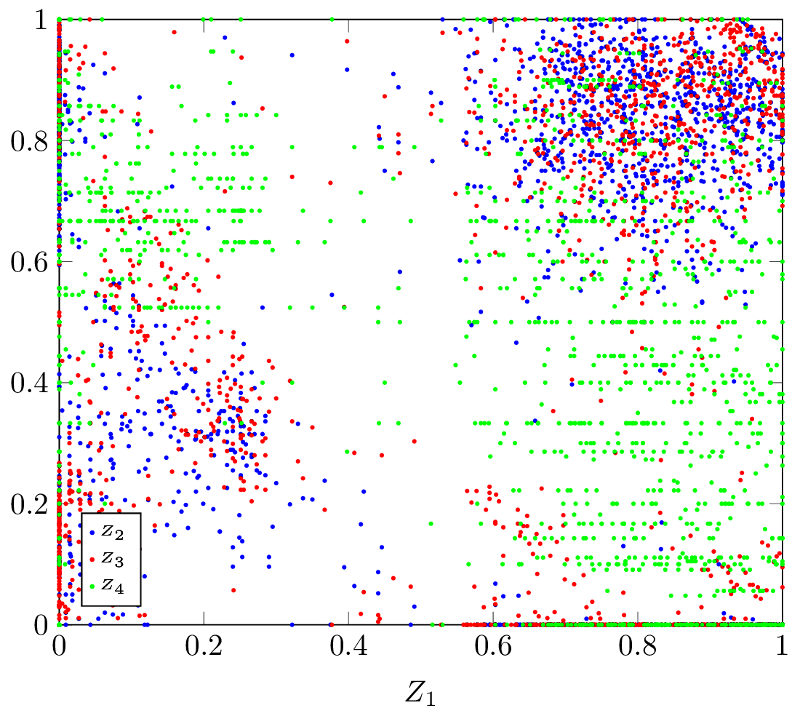}
\caption{Case file 2}
\label{nsp:step42}
\end{subfigure}
\begin{subfigure}[b]{0.24\textwidth}
\centering
\includegraphics[width=\linewidth]{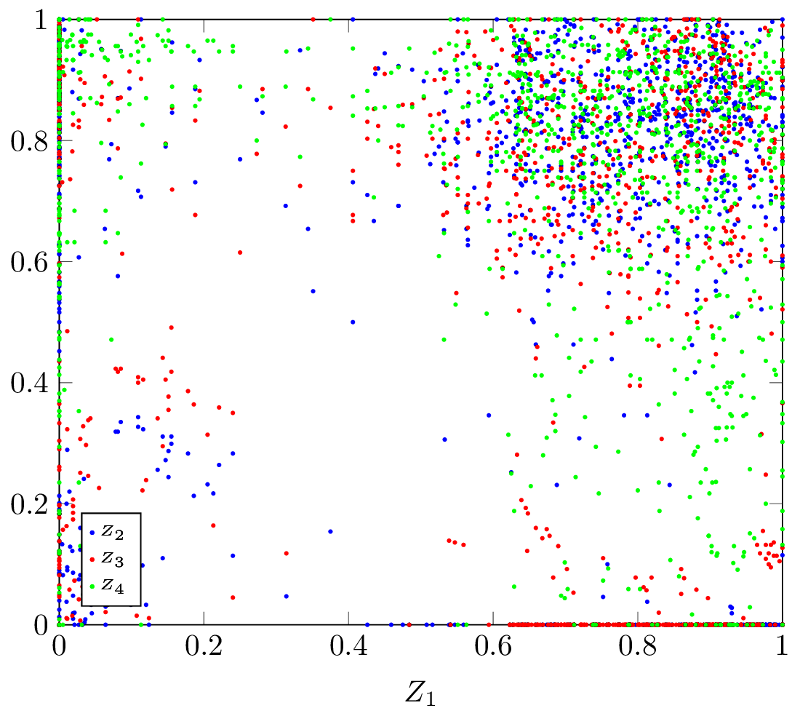}
\caption{Case file 3}
\label{nsp:step43}
\end{subfigure}
\begin{subfigure}[b]{0.24\textwidth}
\centering
\includegraphics[width=\linewidth]{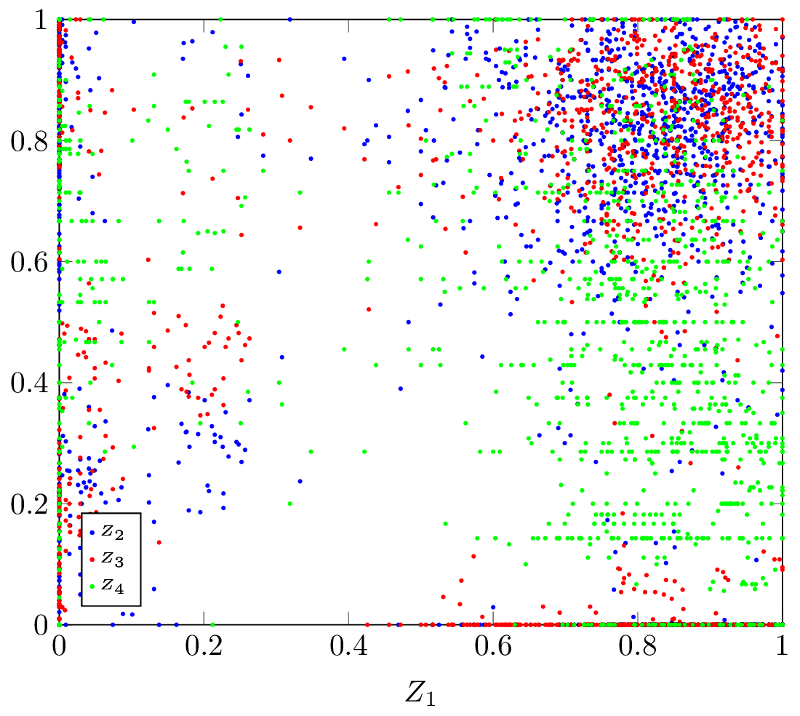}
\caption{Case file 4}
\label{nsp:step44}
\end{subfigure}
\begin{subfigure}[b]{0.24\textwidth}
\centering
\includegraphics[width=\linewidth]{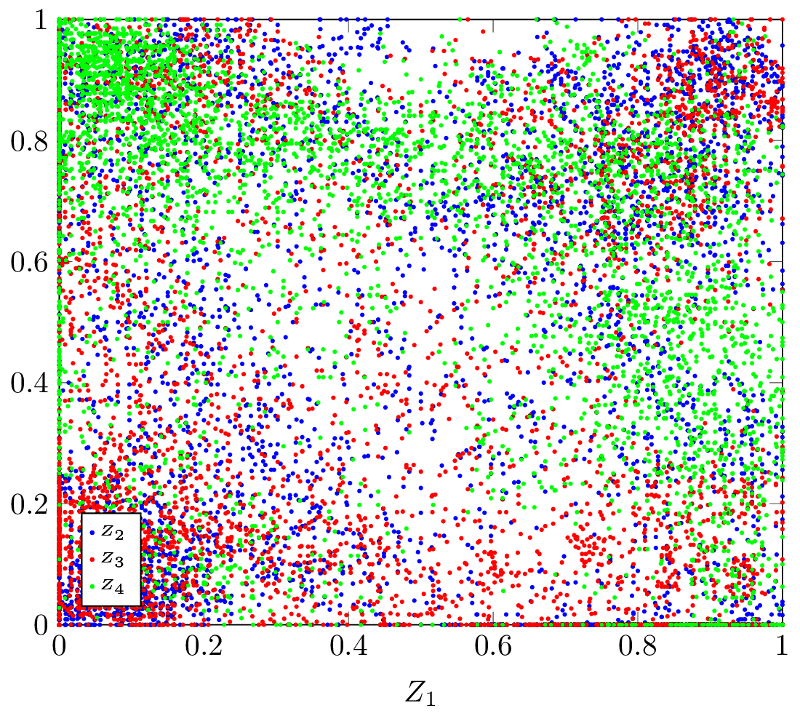}
\caption{Case file 5}
\label{nsp:step45}
\end{subfigure}
\begin{subfigure}[b]{0.24\textwidth}
\centering
\includegraphics[width=\linewidth]{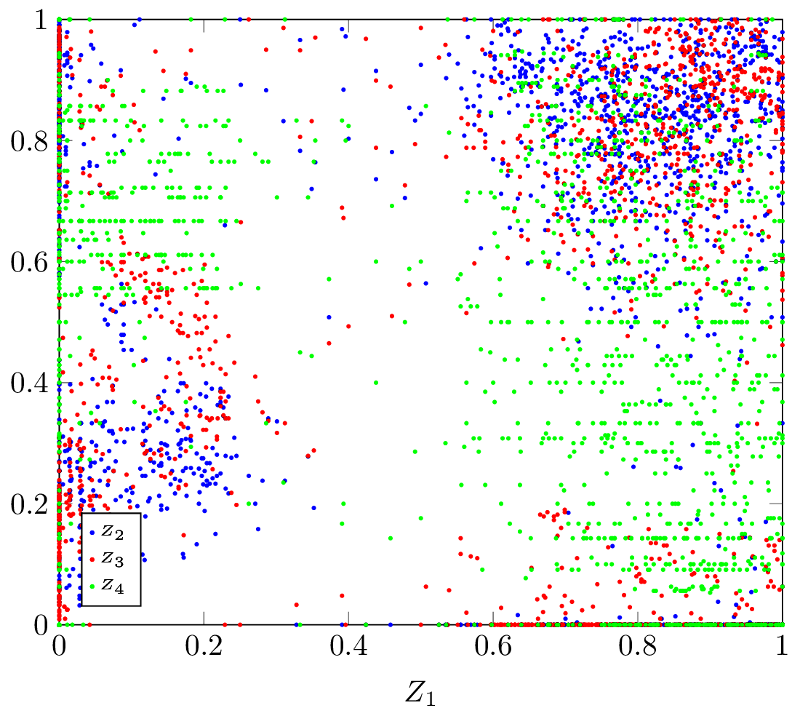}
\caption{Case file 6}
\label{nsp:step46}
\end{subfigure}
\begin{subfigure}[b]{0.24\textwidth}
\centering
\includegraphics[width=\linewidth]{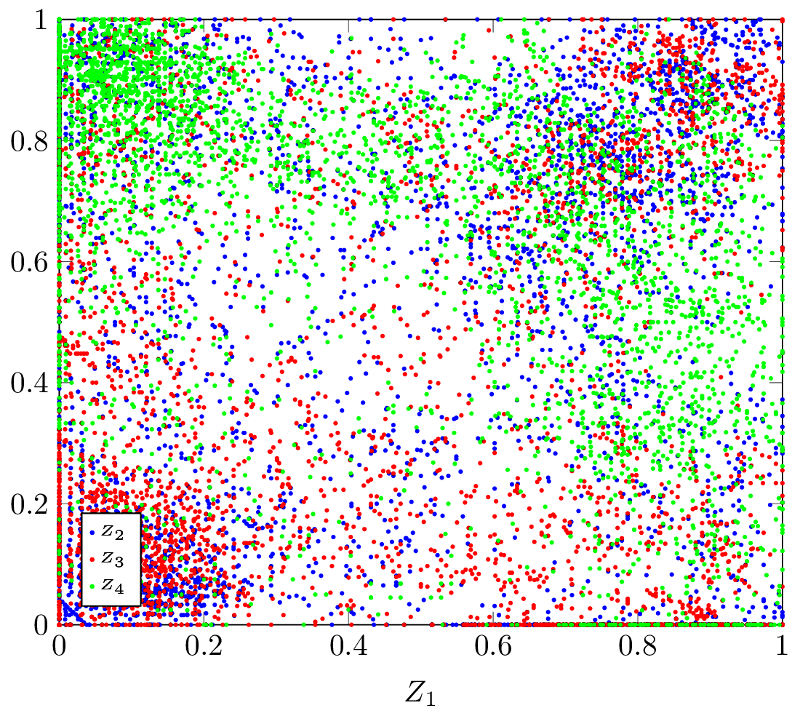}
\caption{Case file 7}
\label{nsp:step47}
\end{subfigure}
\begin{subfigure}[b]{0.24\textwidth}
\centering
\includegraphics[width=\linewidth]{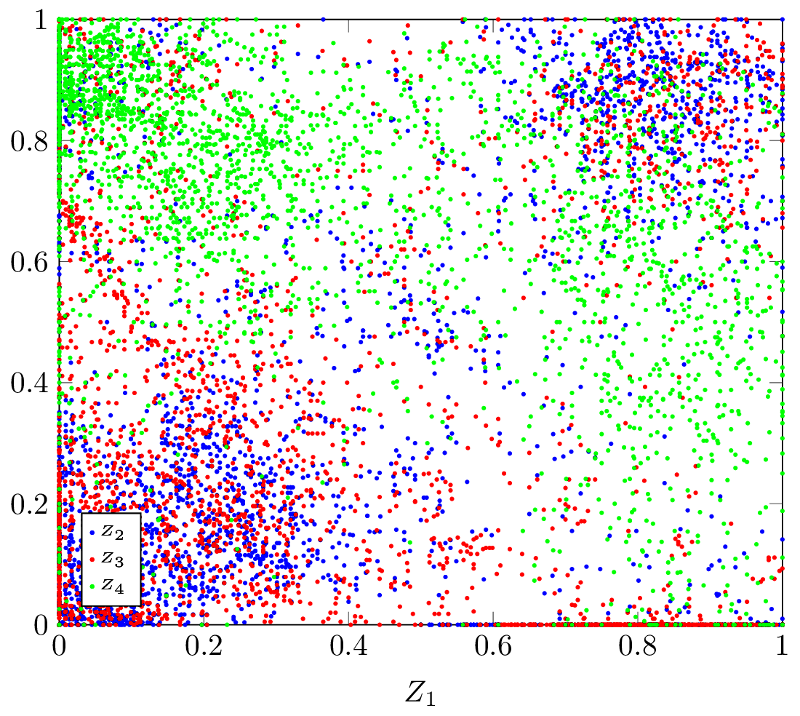}
\caption{Case file 8}
\label{nsp:step48}
\end{subfigure}
\begin{subfigure}[b]{0.45\textwidth}
\centering
\includegraphics[width=\linewidth]{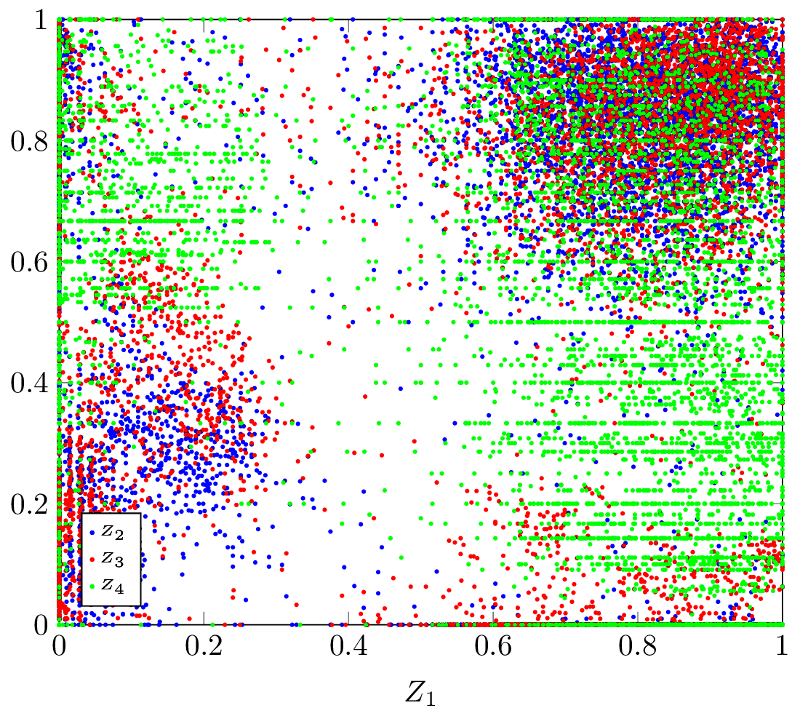}
\caption{Cases files A}
\label{nsp:step4a}
\end{subfigure}
\begin{subfigure}[b]{0.45\textwidth}
\centering
\includegraphics[width=\linewidth]{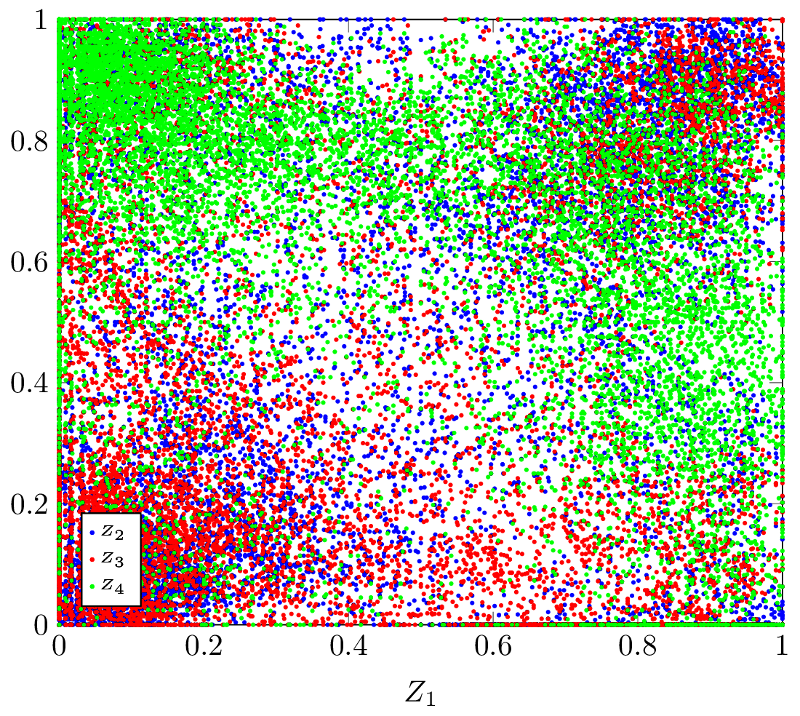}
\caption{Cases files B}
\label{nsp:step4b}
\end{subfigure}
\caption{\hl{Scatter plots of all MONSP sets where the objective $Z_1$  is represented on the $x$ axis and the remaining objectives as data points.}}
\label{nsp:step4}
\end{figure}

\textbf{Step 4 - Multiobjective Scatter Plot Analysis.} Finally, Figure \ref{nsp:step4} presents the scatter plot graphs of all individual cases (\ref{nsp:step41}--\ref{nsp:step48}), for set A (\ref{nsp:step4a}) and for set B (\ref{nsp:step4b}). The Pareto front looks similar in each case, with only minor differences, which are highlighted for sets A and B. A higher concentration of solutions with low $Z_3$ values when $Z_1 \lessapprox 0.2$ can be observed for set B than for set A, along with a higher concentration of large $Z_4$ values for the same interval. However, there are no clear trade-off regions and the overall appearance of both sets is closer to the independent MOMKP set  than to a scenario where global and local relationships and trade-offs are evident. 
% Rodrigo - what other scenarios? I changed this to 'a' because I couldn't see this, but feel free to refer to the scenarios.

\subsubsection{Discussion}

The NSPLib is a relatively well-known single-objective NSP problem database, that was used here to generate scenarios for the MONSP. A straightforward process to create different sets of instances was followed here, maintaining either the objectives information or the base problem definition across all instances. Then, the proposed analysis technique was applied to assess whether these new instances present significant multiobjective traits or not.

The initial two steps of the analysis revealed that the generated sets do not present meaningful differences between themselves. Hence, by grouping the instances using objective files or base file, generates no dissimilarities in fitness landscapes across different instances. The reason for that behaviour is that the data is based on a uniform distribution, hence there are no patterns that could result in different landscapes. When grouping the instances by the case files (which describes the soft constraints), slight dissimilarities were noted between two sets, A and B, namely in $Z_4$ ranges, which presented much smaller values for set A.

Additionally, the fairly high ranges presented in step two showed that it is not possible to have near-optimal values in all objectives simultaneously. However, the threshold analysis and the region maps showed that the distribution of solutions is well spread throughout the Pareto front, meaning that solutions can be found in all regions of the solution space and hence the decision maker can decide about the desired quality of each individual objective. Finally, the scatter plot analysis resulted in graphs resembling the independent MOMKP set, with little difference between sets A and B.

It is then concluded that due to the uniform distribution of the data of the NSPLib, the generated sets did not pose any relevant multiobjective difference between themselves. However, some distinct features were found due to different constraint set ups, a scenario which is completely different from that of the MOMKP instances, where all of the differences were due to the data itself while the constraints remained unchanged. What the instances of set B (cases 5, 7 and 8) have in common is that the minimum number of consecutive working shifts is set to two (set A is zero), which results in considerably more combinations of solutions with violations of that constraint. With a higher range of $Z4$, there is a higher number of solutions in the front, hence the scatter plots look denser.

Regarding the multiobjective nature of these instances, the existing trade-off relates only to the quality of each objective and the fitness landscape is similar for all of the instances. This means that although the problem is multiobjective, it may not pose any additional difficulties to search algorithms as there are no clear local relationships and non-dominated solutions can easily be found in all regions.

Importantly, the uniformity of the data may not reflect real-world problems and algorithms tailored for these scenarios may lack the ability to perform well on more complex fitness landscapes.

\subsection{Multiobjective Vehicle Routing Problem with Time Windows}

\citet{Castro-Gutierrez2011} proposed a new set of benchmark instances for the MOVRPTW after identifying that some instances provided in the literature do not present an interesting multiobjective benchmarking scenario. They proposed different sets based on different time windows ($tw0, \dots, tw4$) and capacity definitions ($\delta0$, $\delta1$, $\delta2$). These are now considered using the proposed analysis technique, to assess whether changing time windows and capacities impacts on the fitness landscape of the MOVRPTW.

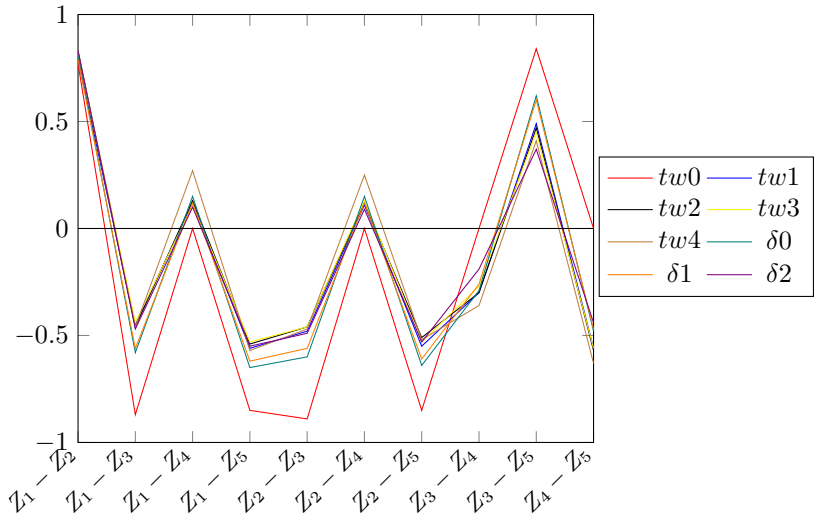
\begin{figure}[!ht]
\centering
\begin{tikzpicture}
\begin{axis}[
legend style={at={(1.01,0.34)},anchor=south west,legend columns=2},
xticklabels={{\small $Z_1-Z_2$},{\small $Z_1-Z_3$},{\small $Z_1-Z_4$},{\small $Z_1-Z_5$},{\small $Z_2-Z_3$},{\small $Z_2-Z_4$},{\small $Z_2-Z_5$},{\small $Z_3-Z_4$},{\small $Z_3-Z_5$},{\small $Z_4-Z_5$}},
xtick=data,
scaled y ticks = false,
x tick label style={rotate=45, anchor=east},
ymax=1,
ymin=-1,
xmin=0,
xmax=9,
cycle list name=color list,
]

\addplot coordinates
{
(0,0.77)
(1,-0.87)
(2,0.00)
(3,-0.85)
(4,-0.89)
(5,0.00)
(6,-0.85)
(7,0.00)
(8,0.84)
(9,0.00)
};

\addplot coordinates
{
(0,0.79)
(1,-0.45)
(2,0.12)
(3,-0.56)
(4,-0.48)
(5,0.11)
(6,-0.55)
(7,-0.30)
(8,0.49)
(9,-0.56)
};

\addplot coordinates
{
(0,0.83)
(1,-0.46)
(2,0.13)
(3,-0.54)
(4,-0.46)
(5,0.12)
(6,-0.51)
(7,-0.30)
(8,0.47)
(9,-0.57)
};

\addplot coordinates
{
(0,0.82)
(1,-0.44)
(2,0.11)
(3,-0.53)
(4,-0.46)
(5,0.13)
(6,-0.53)
(7,-0.27)
(8,0.45)
(9,-0.56)
};

\addplot coordinates
{
(0,0.83)
(1,-0.46)
(2,0.27)
(3,-0.57)
(4,-0.47)
(5,0.25)
(6,-0.52)
(7,-0.36)
(8,0.41)
(9,-0.63)
};

\addplot coordinates
{
(0,0.81)
(1,-0.58)
(2,0.15)
(3,-0.65)
(4,-0.60)
(5,0.15)
(6,-0.64)
(7,-0.29)
(8,0.62)
(9,-0.48)
};

\addplot coordinates
{
(0,0.79)
(1,-0.56)
(2,0.12)
(3,-0.62)
(4,-0.56)
(5,0.12)
(6,-0.61)
(7,-0.26)
(8,0.60)
(9,-0.47)
};

\addplot coordinates
{
(0,0.83)
(1,-0.47)
(2,0.10)
(3,-0.55)
(4,-0.49)
(5,0.09)
(6,-0.53)
(7,-0.19)
(8,0.37)
(9,-0.44)
};

\addplot[] coordinates
{
(0,0)
(1,0)
(2,0)
(3,0)
(4,0)
(5,0)
(6,0)
(7,0)
(8,0)
(9,0)
};

\legend{$tw0$,$tw1$,$tw2$,$tw3$,$tw4$,$\delta0$,$\delta1$,$\delta2$}
\end{axis}
\end{tikzpicture}
\caption{Pairwise correlation values ($y$ axis) for each pair of objectives ($x$ axis). The results for each set of instances are shown in a different colour.}
\label{vrp:step1}
\end{figure}

\textbf{Step 1 - Global Pairwise Relationships Analysis.}
The global pairwise relationship analysis (Figure \ref{vrp:step1}) corroborates the claim that these instances provide interesting multiobjective challenges, as some pairs of objectives have either high harmonious or conflicting relationships. It is, however, noticeable that changing the time windows and $\delta$ does not greatly impact on the nature of the pairwise global relationships. The only exception is with $tw0$ which slightly deviates from the other configurations by presenting stronger (conflicting and harmonious) relationships.

\begin{figure}[!ht]
\centering
\begin{subfigure}[t]{0.31\textwidth}
\centering
\resizebox{\textwidth}{!}{%
\begin{tikzpicture}
\begin{axis} [symbolic x coords={$tw0$,$tw1$,$tw2$,$tw3$,$tw4$,$\delta0$,$\delta1$,$\delta2$},
xtick={$tw0$,$tw1$,$tw2$,$tw3$,$tw4$,$\delta0$,$\delta1$,$\delta2$},
]
\addplot+[forget plot,only marks] plot[color=black,error bars/.cd, y dir=plus, y explicit]
coordinates{
($tw0$,77.68)+-(16.32,16.32)
($tw1$,75.08)+-(18.25,18.25)
($tw2$,72.11)+-(23.89,23.89)
($tw3$,74.56)+-(22.50,22.50)
($tw4$,69.22)+-(21.84,21.84)
($\delta0$,73.74)+-(19.92,19.92)
($\delta1$,74.59)+-(19.34,19.34)
($\delta2$,72.38)+-(22.42,22.42)
};
\addplot+[only marks] plot[color=black,error bars/.cd, y dir=minus, y explicit]
coordinates{
($tw0$,77.68)+-(48.12,48.12)
($tw1$,75.08)+-(52.41,52.41)
($tw2$,72.11)+-(50.44,50.44)
($tw3$,74.56)+-(52.05,52.05)
($tw4$,69.22)+-(47.38,47.38)
($\delta0$,73.74)+-(52.74,52.74)
($\delta1$,74.59)+-(52.26,52.26)
($\delta2$,72.38)+-(45.25,45.25)

};
\end{axis} 
\end{tikzpicture}
}
\caption{$Z_1$ -- Number of Vehicles}
\end{subfigure}
\begin{subfigure}[t]{0.32\textwidth}
\centering
\resizebox{\textwidth}{!}{%
\begin{tikzpicture}
\begin{axis} [symbolic x coords={$tw0$,$tw1$,$tw2$,$tw3$,$tw4$,$\delta0$,$\delta1$,$\delta2$},
xtick={$tw0$,$tw1$,$tw2$,$tw3$,$tw4$,$\delta0$,$\delta1$,$\delta2$},
]
\addplot+[forget plot,only marks] plot[color=black,error bars/.cd, y dir=plus, y explicit]
coordinates{
($tw0$,5891.65)+-(1866.94,1866.94)
($tw1$,5651.89)+-(1698.58,1698.58)
($tw2$,5366.89)+-(2141.26,2141.26)
($tw3$,5420.89)+-(2122.65,2122.65)
($tw4$,5166.82)+-(2081.67,2081.67)
($\delta0$,5511.00)+-(1912.54,1912.54)
($\delta1$,5586.34)+-(1934.52,1934.52)
($\delta2$,5401.55)+-(2099.60,2099.60)
};
\addplot+[only marks] plot[color=black,error bars/.cd, y dir=minus, y explicit]
coordinates{
($tw0$,5891.65)+-(4715.94,4715.94)
($tw1$,5651.89)+-(4752.45,4752.45)
($tw2$,5366.89)+-(4716.95,4716.95)
($tw3$,5420.89)+-(4805.19,4805.19)
($tw4$,5166.82)+-(4511.59,4511.59)
($\delta0$,5511.00)+-(4876.02,4876.02)
($\delta1$,5586.34)+-(4851.14,4851.14)
($\delta2$,5401.55)+-(4374.12,4374.12)
};
\end{axis} 
\end{tikzpicture}
}
\caption{$Z_2$ -- Total Travel Distances}
\end{subfigure}
\begin{subfigure}[t]{0.32\textwidth}
\centering
\resizebox{\textwidth}{!}{%
\begin{tikzpicture}
\begin{axis} [symbolic x coords={$tw0$,$tw1$,$tw2$,$tw3$,$tw4$,$\delta0$,$\delta1$,$\delta2$},
xtick={$tw0$,$tw1$,$tw2$,$tw3$,$tw4$,$\delta0$,$\delta1$,$\delta2$},
scaled y ticks = false,
]
\addplot+[forget plot,only marks] plot[color=black,error bars/.cd, y dir=plus, y explicit]
coordinates{
($tw0$,25533.58)+-(16679.75,16679.75)
($tw1$,33732.90)+-(20280.44,20280.44)
($tw2$,35741.37)+-(20351.96,20351.96)
($tw3$,37095.45)+-(18271.21,18271.21)
($tw4$,34924.51)+-(17042.15,17042.15)
($\delta0$,34149.16)+-(23214.84,23214.84)
($\delta1$,33910.73)+-(20893.27,20893.27)
($\delta2$,32156.80)+-(11467.20,11467.20)
};
\addplot+[only marks] plot[color=black,error bars/.cd, y dir=minus, y explicit]
coordinates{
($tw0$,25533.58)+-(15560.25,15560.25)
($tw1$,33732.90)+-(23732.90,23732.90)
($tw2$,35741.37)+-(25741.37,25741.37)
($tw3$,37095.45)+-(27095.45,27095.45)
($tw4$,34924.51)+-(24924.51,24924.51)
($\delta0$,34149.16)+-(24154.49,24154.49)
($\delta1$,33910.73)+-(23916.07,23916.07)
($\delta2$,32156.80)+-(22162.13,22162.13)
};
\end{axis} 
\end{tikzpicture}
}
\caption{$Z_3$ -- Makespan}
\end{subfigure}
\\
\begin{subfigure}[t]{0.32\textwidth}
\centering
\resizebox{\textwidth}{!}{%
\begin{tikzpicture}
\begin{axis} [symbolic x coords={$tw0$,$tw1$,$tw2$,$tw3$,$tw4$,$\delta0$,$\delta1$,$\delta2$},
xtick={$tw0$,$tw1$,$tw2$,$tw3$,$tw4$,$\delta0$,$\delta1$,$\delta2$},
scaled y ticks = false,
]
\addplot+[forget plot,only marks] plot[color=black,error bars/.cd, y dir=plus, y explicit]
coordinates{
($tw0$,0.00)+-(0.00,0.00)
($tw1$,180121.75)+-(201551.58,201551.58)
($tw2$,175940.54)+-(260006.13,260006.13)
($tw3$,185528.85)+-(336697.81,336697.81)
($tw4$,190741.99)+-(288238.01,288238.01)
($\delta0$,147610.99)+-(215205.01,215205.01)
($\delta1$,154290.84)+-(210501.16,210501.16)
($\delta2$,137498.06)+-(226189.94,226189.94)
};
\addplot+[only marks] plot[color=black,error bars/.cd, y dir=minus, y explicit]
coordinates{
($tw0$,0.00)+-(0.00,0.00)
($tw1$,180121.75)+-(180121.75,180121.75)
($tw2$,175940.54)+-(175940.54,175940.54)
($tw3$,185528.85)+-(185528.85,185528.85)
($tw4$,190741.99)+-(190741.99,190741.99)
($\delta0$,147610.99)+-(147610.99,147610.99)
($\delta1$,154290.84)+-(154290.84,154290.84)
($\delta2$,137498.06)+-(137498.06,137498.06)
};
\end{axis} 
\end{tikzpicture}
}
\caption{$Z_4$ -- Total Waiting Time}
\end{subfigure}
\begin{subfigure}[t]{0.32\textwidth}
\centering
\resizebox{\textwidth}{!}{%
\begin{tikzpicture}
\begin{axis} [symbolic x coords={$tw0$,$tw1$,$tw2$,$tw3$,$tw4$,$\delta0$,$\delta1$,$\delta2$},
xtick={$tw0$,$tw1$,$tw2$,$tw3$,$tw4$,$\delta0$,$\delta1$,$\delta2$},
scaled y ticks = false,
]
\addplot+[forget plot,only marks] plot[color=black,error bars/.cd, y dir=plus, y explicit]
coordinates{
($tw0$,3408.85)+-(30931.15,30931.15)
($tw1$,21896.79)+-(202169.88,202169.88)
($tw2$,47100.42)+-(291226.25,291226.25)
($tw3$,70178.76)+-(408954.58,408954.58)
($tw4$,47156.61)+-(237583.39,237583.39)
($\delta0$,38390.45)+-(269833.55,269833.55)
($\delta1$,38031.21)+-(262424.79,262424.79)
($\delta2$,37423.20)+-(170260.80,170260.80)
};
\addplot+[only marks] plot[color=black,error bars/.cd, y dir=minus, y explicit]
coordinates{
($tw0$,3408.85)+-(3408.85,3408.85)
($tw1$,21896.79)+-(21896.79,21896.79)
($tw2$,47100.42)+-(47100.42,47100.42)
($tw3$,70178.76)+-(70178.76,70178.76)
($tw4$,47156.61)+-(47156.61,47156.61)
($\delta0$,38390.45)+-(38390.45,38390.45)
($\delta1$,38031.21)+-(38031.21,38031.21)
($\delta2$,37423.20)+-(37423.20,37423.20)
};
\end{axis} 
\end{tikzpicture}
}
\caption{$Z_5$ -- Total Delay Time}
\label{vrp:step2e}
\end{subfigure}
\caption{Range analysis results for each objective considering instances with different time windows and delta set-ups. In each line, the central point is the average value
for the set, with the top and bottom of the line representing the maximum and minimum values found,
respectively.}
\label{vrp:step2}
\end{figure}
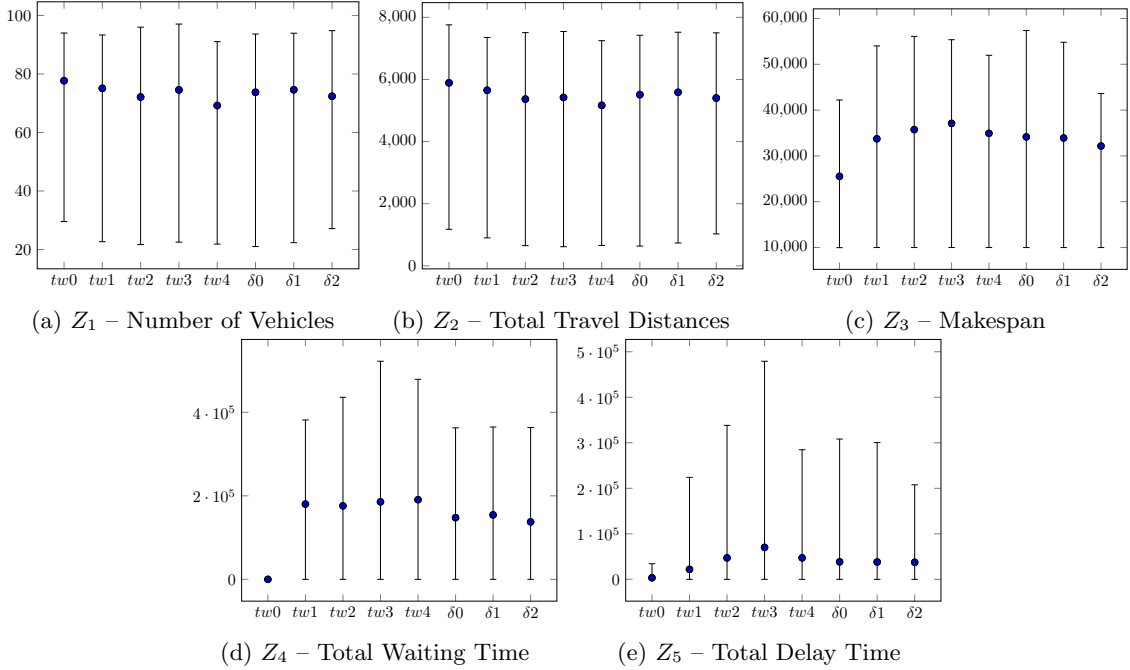

\textbf{Step 2 - Objective Range Analysis.}
On the objective range analysis (Figure \ref{vrp:step2}), because the values of individual objectives are similar, the actual values are presented here rather than the relative margin, with a chart for each objective. 
Excluding $Z_4$ and $Z_5$ for $tw0$, all objectives present high ranges, again supporting the claim of interesting multiobjective scenarios. The average values found for all setups are similar, but some ranges, namely $Z_4$ and $Z_5$ present high variance. On $Z_4$ and $Z_5$, it is clear that $tw0$ is smaller than the other scenarios. $Z_5$ (\ref{vrp:step2e}) is also the objective with the larger discrepancy between sets.

\begin{figure}[!ht]
\centering
\resizebox{0.4\textwidth}{!}{%
\begin{tikzpicture}
	\begin{axis}[
		xlabel=Threshold,
		ylabel=\# Of Instances,
		xmin=0,
		xmax=1,
		legend pos= north west,	
		ymin=0,]
		]
	\addplot[color=blue,smooth] coordinates {
(	1	,	0	)
(	0.98	,	0	)
(	0.96	,	0	)
(	0.94	,	0	)
(	0.92	,	0	)
(	0.9	,	0	)
(	0.88	,	0	)
(	0.86	,	0	)
(	0.84	,	0	)
(	0.82	,	0	)
(	0.8	,	0	)
(	0.78	,	0	)
(	0.76	,	0	)
(	0.74	,	0	)
(	0.72	,	0	)
(	0.7	,	0	)
(	0.68	,	0	)
(	0.66	,	0	)
(	0.64	,	0	)
(	0.62	,	0	)
(	0.6	,	1	)
(	0.58	,	1	)
(	0.56	,	1	)
(	0.54	,	1	)
(	0.52	,	1	)
(	0.5	,	2	)
(	0.48	,	2	)
(	0.46	,	3	)
(	0.44	,	3	)
(	0.42	,	3	)
(	0.4	,	5	)
(	0.38	,	9	)
(	0.36	,	10	)
(	0.34	,	11	)
(	0.32	,	16	)
(	0.3	,	20	)
(	0.28	,	22	)
(	0.26	,	30	)
(	0.24	,	32	)
(	0.22	,	37	)
(	0.2	,	41	)
(	0.18	,	44	)
(	0.16	,	45	)
(	0.14	,	45	)
(	0.12	,	45	)
(	0.1	,	45	)
(	0.08	,	45	)
(	0.06	,	45	)
(	0.04	,	45	)
(	0.02	,	45	)

	};	
	\end{axis}
\end{tikzpicture}
}
\caption{Threshold analysis showing the number of scenarios with solutions found in region $r_0$ when the threshold decreases.}
\label{vrp:step3a}
\end{figure}
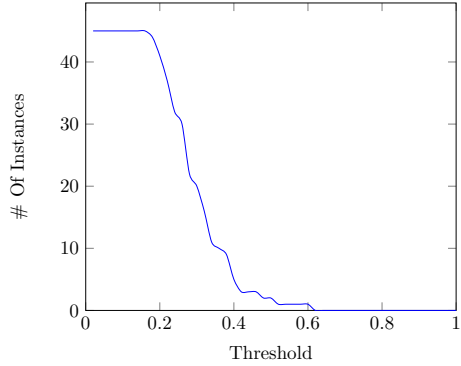

\textbf{Step 3 - Trade-Off Regions Analysis.}
The threshold analysis, presented in Figure \ref{vrp:step3a}, reveals that there are no instances with solutions in $r_0$ when the threshold is roughly 60\% or higher, meaning that if a solution with over 60\% quality in at least one criterion is required, then one or more different criteria will present inferior quality.  
The steepness of the chart is also accentuated if compared to the MONSP, meaning that the instances in the MOVRPTW are more alike than the MONSP instances. 

\begin{figure}[!ht]
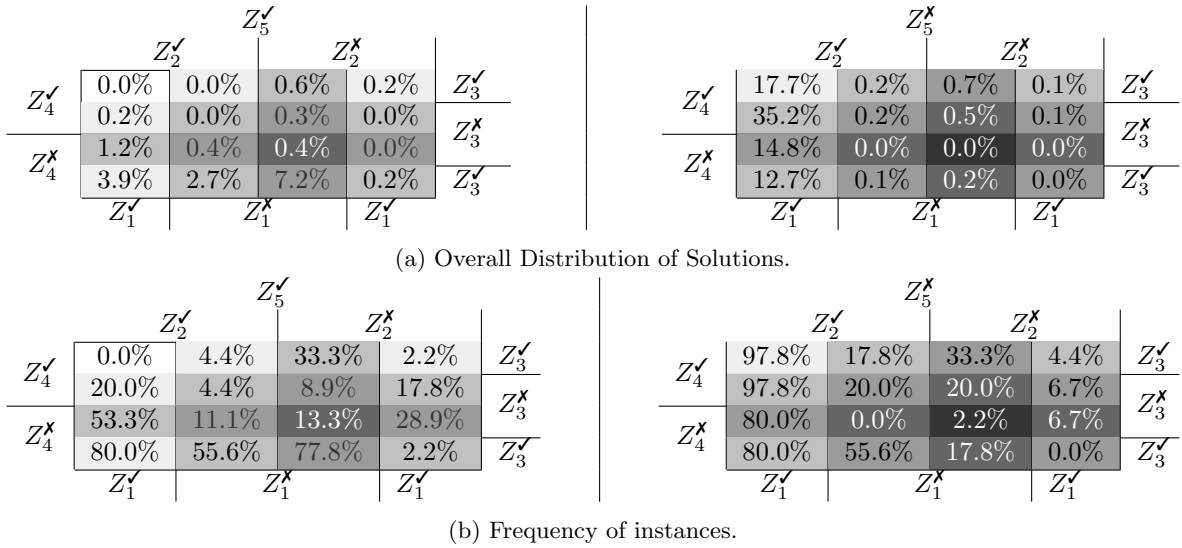

\centering
\begin{subfigure}{\linewidth}
\resizebox{\linewidth}{!}{%
\begin{tabularx}{\textwidth}{XX|X|X|XXX|XXX|X|X|XX}
\multicolumn{6}{c}{$Z_5^\text{\cmark}$} &  & \multicolumn{1}{c}{} & \multicolumn{6}{c}{$Z_5^\text{\xmark}$} \\
 & \multicolumn{2}{c|}{$Z_2^\text{\cmark}$} & \multicolumn{2}{c|}{$Z_2^\text{\xmark}$} &  &  &  &  & \multicolumn{2}{c|}{$Z_2^\text{\cmark}$} & \multicolumn{2}{c|}{$Z_2^\text{\xmark}$} &  \\ \cline{2-5} \cline{10-13}
\multicolumn{1}{c|}{} & \multicolumn{1}{c|}{0.0\%} & \multicolumn{1}{c|}{\cellcolor[HTML]{EFEFEF}0.0\%} & \multicolumn{1}{c|}{\cellcolor[HTML]{C0C0C0}0.6\%} & \multicolumn{1}{c|}{\cellcolor[HTML]{EFEFEF}0.2\%} & $Z_3^\text{\cmark}$ &  &  & \multicolumn{1}{c|}{} & \multicolumn{1}{c|}{\cellcolor[HTML]{EFEFEF}17.7\%} & \multicolumn{1}{c|}{\cellcolor[HTML]{C0C0C0}0.2\%} & \multicolumn{1}{c|}{\cellcolor[HTML]{9B9B9B}0.7\%} & \multicolumn{1}{c|}{\cellcolor[HTML]{C0C0C0}0.1\%} & $Z_3^\text{\cmark}$ \\ \cline{2-6} \cline{10-14} 
\multicolumn{1}{c|}{\multirow{-2}{*}{$Z_4^\text{\cmark}$}} & \multicolumn{1}{c|}{\cellcolor[HTML]{EFEFEF}0.2\%} & \multicolumn{1}{c|}{\cellcolor[HTML]{C0C0C0}0.0\%} & \multicolumn{1}{c|}{\cellcolor[HTML]{9B9B9B}{\color[HTML]{333333} 0.3\%}} & \multicolumn{1}{c|}{\cellcolor[HTML]{C0C0C0}0.0\%} &  &  &  & \multicolumn{1}{c|}{\multirow{-2}{*}{$Z_4^\text{\cmark}$}} & \multicolumn{1}{c|}{\cellcolor[HTML]{C0C0C0}35.2\%} & \multicolumn{1}{c|}{\cellcolor[HTML]{9B9B9B}0.2\%} & \multicolumn{1}{c|}{\cellcolor[HTML]{656565}{\color[HTML]{FFFFFF} 0.5\%}} & \multicolumn{1}{c|}{\cellcolor[HTML]{9B9B9B}0.1\%} &  \\ \cline{1-5} \cline{9-13}
\multicolumn{1}{c|}{} & \multicolumn{1}{c|}{\cellcolor[HTML]{C0C0C0}1.2\%} & \multicolumn{1}{c|}{\cellcolor[HTML]{9B9B9B}{\color[HTML]{333333} 0.4\%}} & \multicolumn{1}{c|}{\cellcolor[HTML]{656565}{\color[HTML]{FFFFFF} 0.4\%}} & \multicolumn{1}{c|}{\cellcolor[HTML]{9B9B9B}{\color[HTML]{333333} 0.0\%}} & \multirow{-2}{*}{$Z_3^\text{\xmark}$} &  &  & \multicolumn{1}{c|}{} & \multicolumn{1}{c|}{\cellcolor[HTML]{9B9B9B}14.8\%} & \multicolumn{1}{c|}{\cellcolor[HTML]{656565}{\color[HTML]{FFFFFF} 0.0\%}} & \multicolumn{1}{c|}{\cellcolor[HTML]{343434}{\color[HTML]{FFFFFF} 0.0\%}} & \multicolumn{1}{c|}{\cellcolor[HTML]{656565}{\color[HTML]{FFFFFF} 0.0\%}} & \multirow{-2}{*}{$Z_3^\text{\xmark}$} \\ \cline{2-6} \cline{10-14} 
\multicolumn{1}{c|}{\multirow{-2}{*}{$Z_4^\text{\xmark}$}} & \multicolumn{1}{c|}{\cellcolor[HTML]{EFEFEF}3.9\%} & \multicolumn{1}{c|}{\cellcolor[HTML]{C0C0C0}2.7\%} & \multicolumn{1}{c|}{\cellcolor[HTML]{9B9B9B}{\color[HTML]{333333} 7.2\%}} & \multicolumn{1}{c|}{\cellcolor[HTML]{C0C0C0}0.2\%} & $Z_3^\text{\cmark}$ &  &  & \multicolumn{1}{c|}{\multirow{-2}{*}{$Z_4^\text{\xmark}$}} & \multicolumn{1}{c|}{\cellcolor[HTML]{C0C0C0}12.7\%} & \multicolumn{1}{c|}{\cellcolor[HTML]{9B9B9B}0.1\%} & \multicolumn{1}{c|}{\cellcolor[HTML]{656565}{\color[HTML]{FFFFFF} 0.2\%}} & \multicolumn{1}{c|}{\cellcolor[HTML]{9B9B9B}0.0\%} & $Z_3^\text{\cmark}$ \\ \cline{2-5} \cline{10-13}
 & \multicolumn{1}{c|}{$Z_1^\text{\cmark}$} & \multicolumn{2}{c|}{$Z_1^\text{\xmark}$} & $Z_1^\text{\cmark}$ &  &  &  &  & \multicolumn{1}{c|}{$Z_1^\text{\cmark}$} & \multicolumn{2}{c|}{$Z_1^\text{\xmark}$} & $Z_1^\text{\cmark}$ & 
\end{tabularx}%
}
\subcaption{Overall Distribution of Solutions.}
\end{subfigure}
\begin{subfigure}{\linewidth}
\resizebox{\linewidth}{!}{%
\begin{tabularx}{\textwidth}{XX|X|X|XXX|XXX|X|X|XX}
\multicolumn{6}{c}{$Z_5^\text{\cmark}$} &  & \multicolumn{1}{c}{} & \multicolumn{6}{c}{$Z_5^\text{\xmark}$} \\
 & \multicolumn{2}{c|}{$Z_2^\text{\cmark}$} & \multicolumn{2}{c|}{$Z_2^\text{\xmark}$} &  &  &  &  & \multicolumn{2}{c|}{$Z_2^\text{\cmark}$} & \multicolumn{2}{c|}{$Z_2^\text{\xmark}$} &  \\ \cline{2-5} \cline{10-13}
\multicolumn{1}{c|}{} & \multicolumn{1}{c|}{0.0\%} & \multicolumn{1}{c|}{\cellcolor[HTML]{EFEFEF}4.4\%} & \multicolumn{1}{c|}{\cellcolor[HTML]{C0C0C0}33.3\%} & \multicolumn{1}{c|}{\cellcolor[HTML]{EFEFEF}2.2\%} & $Z_3^\text{\cmark}$ &  &  & \multicolumn{1}{c|}{} & \multicolumn{1}{c|}{\cellcolor[HTML]{EFEFEF}97.8\%} & \multicolumn{1}{c|}{\cellcolor[HTML]{C0C0C0}17.8\%} & \multicolumn{1}{c|}{\cellcolor[HTML]{9B9B9B}33.3\%} & \multicolumn{1}{c|}{\cellcolor[HTML]{C0C0C0}4.4\%} & $Z_3^\text{\cmark}$ \\ \cline{2-6} \cline{10-14} 
\multicolumn{1}{c|}{\multirow{-2}{*}{$Z_4^\text{\cmark}$}} & \multicolumn{1}{c|}{\cellcolor[HTML]{EFEFEF}20.0\%} & \multicolumn{1}{c|}{\cellcolor[HTML]{C0C0C0}4.4\%} & \multicolumn{1}{c|}{\cellcolor[HTML]{9B9B9B}{\color[HTML]{333333} 8.9\%}} & \multicolumn{1}{c|}{\cellcolor[HTML]{C0C0C0}17.8\%} &  &  &  & \multicolumn{1}{c|}{\multirow{-2}{*}{$Z_4^\text{\cmark}$}} & \multicolumn{1}{c|}{\cellcolor[HTML]{C0C0C0}97.8\%} & \multicolumn{1}{c|}{\cellcolor[HTML]{9B9B9B}20.0\%} & \multicolumn{1}{c|}{\cellcolor[HTML]{656565}{\color[HTML]{FFFFFF} 20.0\%}} & \multicolumn{1}{c|}{\cellcolor[HTML]{9B9B9B}6.7\%} &  \\ \cline{1-5} \cline{9-13}
\multicolumn{1}{c|}{} & \multicolumn{1}{c|}{\cellcolor[HTML]{C0C0C0}53.3\%} & \multicolumn{1}{c|}{\cellcolor[HTML]{9B9B9B}{\color[HTML]{333333} 11.1\%}} & \multicolumn{1}{c|}{\cellcolor[HTML]{656565}{\color[HTML]{FFFFFF} 13.3\%}} & \multicolumn{1}{c|}{\cellcolor[HTML]{9B9B9B}{\color[HTML]{333333} 28.9\%}} & \multirow{-2}{*}{$Z_3^\text{\xmark}$} &  &  & \multicolumn{1}{c|}{} & \multicolumn{1}{c|}{\cellcolor[HTML]{9B9B9B}80.0\%} & \multicolumn{1}{c|}{\cellcolor[HTML]{656565}{\color[HTML]{FFFFFF} 0.0\%}} & \multicolumn{1}{c|}{\cellcolor[HTML]{343434}{\color[HTML]{FFFFFF} 2.2\%}} & \multicolumn{1}{c|}{\cellcolor[HTML]{656565}{\color[HTML]{FFFFFF} 6.7\%}} & \multirow{-2}{*}{$Z_3^\text{\xmark}$} \\ \cline{2-6} \cline{10-14} 
\multicolumn{1}{c|}{\multirow{-2}{*}{$Z_4^\text{\xmark}$}} & \multicolumn{1}{c|}{\cellcolor[HTML]{EFEFEF}80.0\%} & \multicolumn{1}{c|}{\cellcolor[HTML]{C0C0C0}55.6\%} & \multicolumn{1}{c|}{\cellcolor[HTML]{9B9B9B}{\color[HTML]{333333} 77.8\%}} & \multicolumn{1}{c|}{\cellcolor[HTML]{C0C0C0}2.2\%} & $Z_3^\text{\cmark}$ &  &  & \multicolumn{1}{c|}{\multirow{-2}{*}{$Z_4^\text{\xmark}$}} & \multicolumn{1}{c|}{\cellcolor[HTML]{C0C0C0}80.0\%} & \multicolumn{1}{c|}{\cellcolor[HTML]{9B9B9B}55.6\%} & \multicolumn{1}{c|}{\cellcolor[HTML]{656565}{\color[HTML]{FFFFFF} 17.8\%}} & \multicolumn{1}{c|}{\cellcolor[HTML]{9B9B9B}0.0\%} & $Z_3^\text{\cmark}$ \\ \cline{2-5} \cline{10-13}
 & \multicolumn{1}{c|}{$Z_1^\text{\cmark}$} & \multicolumn{2}{c|}{$Z_1^\text{\xmark}$} & $Z_1^\text{\cmark}$ &  &  &  &  & \multicolumn{1}{c|}{$Z_1^\text{\cmark}$} & \multicolumn{2}{c|}{$Z_1^\text{\xmark}$} & $Z_1^\text{\cmark}$ & 
\end{tabularx}%
}
\subcaption{Frequency of instances.}
\end{subfigure}
\caption{Overall region maps.}
\label{vrp:step3b}
\end{figure}

Figure \ref{vrp:step3b} presents the region map and the frequency map using the threshold found. 
Many regions without solutions are identified, representing trade-offs for the decision-maker. 
Some global relationships can also be observed, for instance, because $Z_1$ and $Z_2$ are harmonious, few solutions can be found when $Z_1 = \text{\cmark}$ and $Z_2= \text{\xmark}$ and vice-versa.

Interestingly, because  $Z_3$ and $Z_5$ are fairly globally harmonious, a relationship similar to $Z_1$ and $Z_2$ would be expected, however over 17\% solutions are in regions where  $Z_3 = \text{\cmark}$ and  $Z_5 = \text{\xmark}$ or  $Z_3 = \text{\xmark}$ and $Z_5 = \text{\cmark}$. 
% Rodrigo: I changtes whether to where here, because I think this is what you meant.

\begin{figure}[!ht]
\centering
\begin{subfigure}[b]{0.24\textwidth}
\centering
\includegraphics[width=\linewidth]{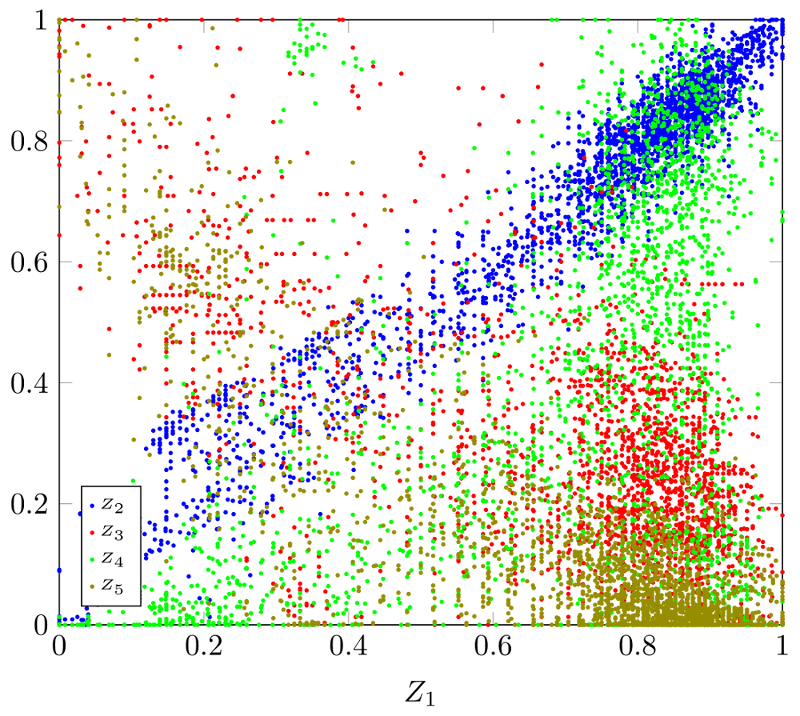}
\caption{Delta $\delta0$}
\label{vrp:step4d0}
\end{subfigure}
\begin{subfigure}[b]{0.24\textwidth}
\centering
\includegraphics[width=\linewidth]{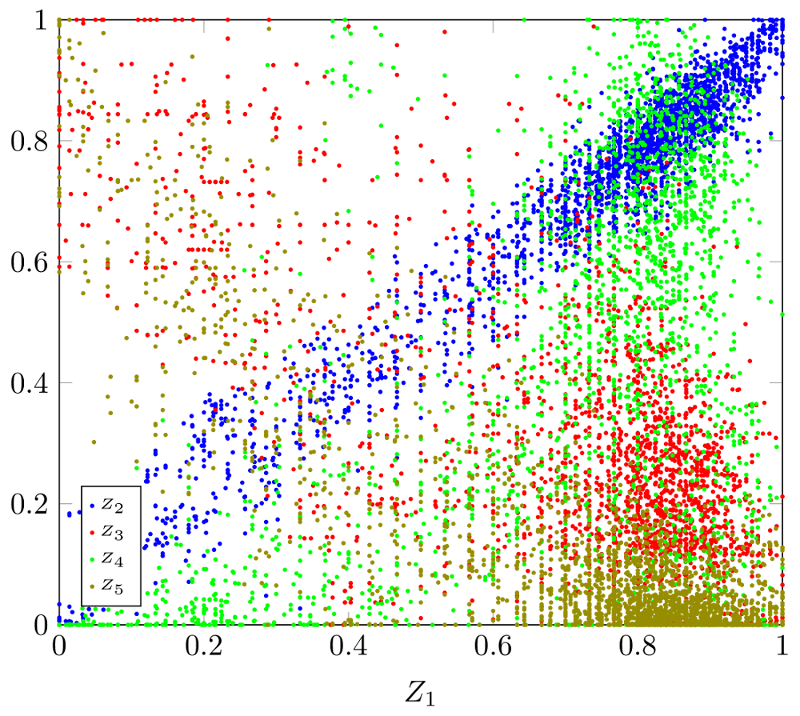}
\caption{Delta $\delta1$}
\label{vrp:step4d1}
\end{subfigure}
\begin{subfigure}[b]{0.24\textwidth}
\centering
\includegraphics[width=\linewidth]{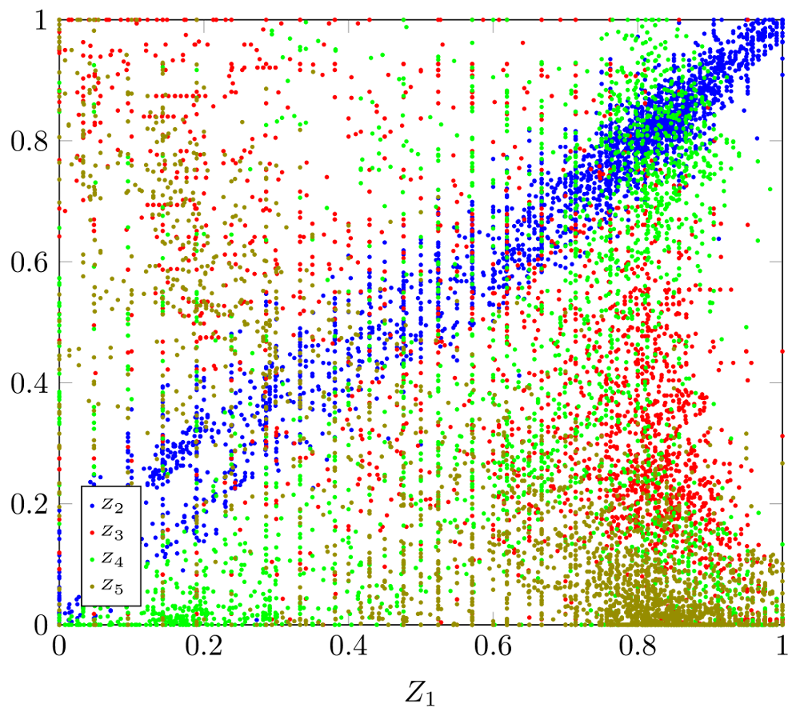}
\caption{Delta $\delta2$}
\label{vrp:step4d2}
\end{subfigure}
\begin{subfigure}[b]{0.24\textwidth}
\centering
\includegraphics[width=\linewidth]{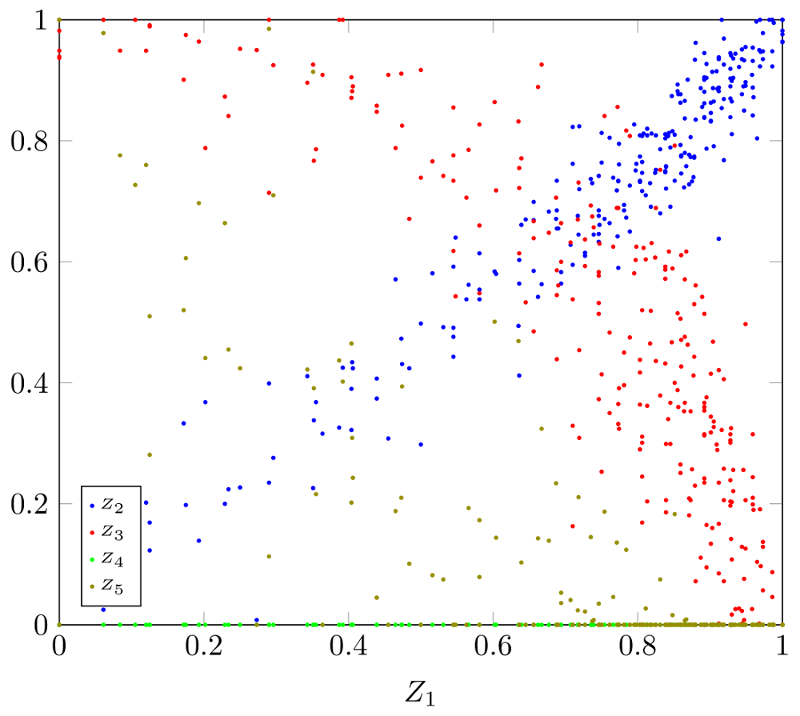}
\caption{Time window $tw0$}
\label{vrp:step4tw0}
\end{subfigure}
\begin{subfigure}[b]{0.24\textwidth}
\centering
\includegraphics[width=\linewidth]{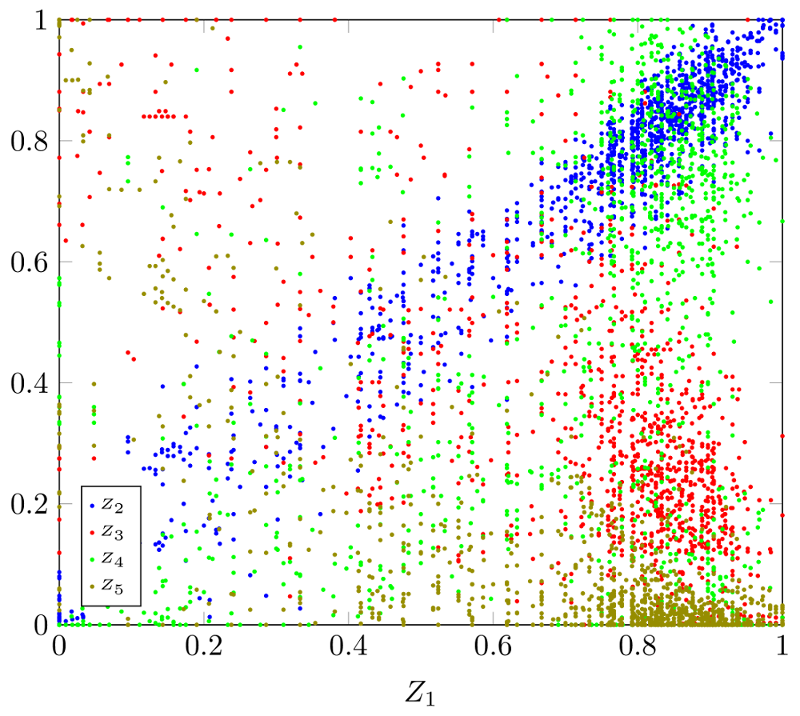}
\caption{Time window $tw1$}
\label{vrp:steptw1}
\end{subfigure}
\begin{subfigure}[b]{0.24\textwidth}
\centering
\includegraphics[width=\linewidth]{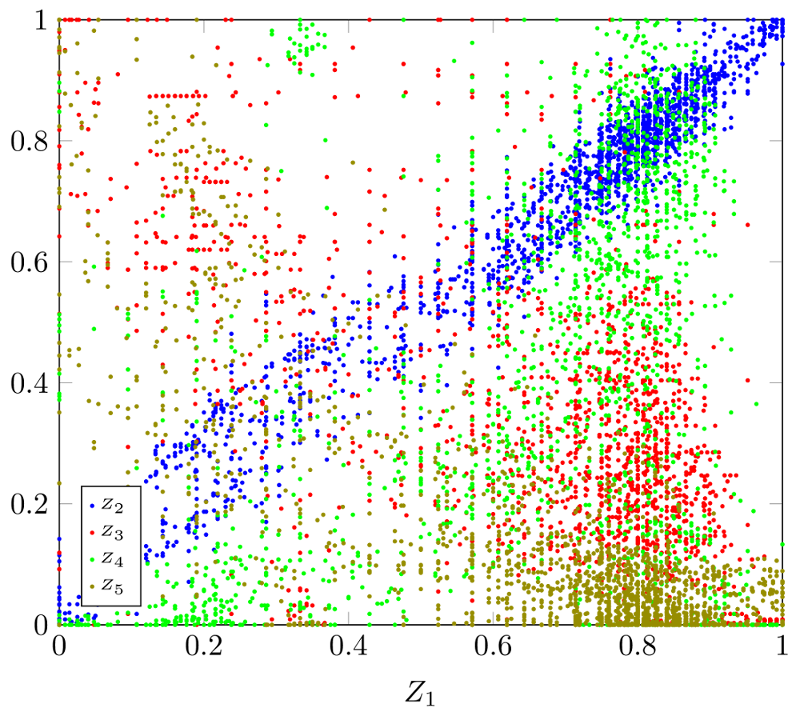}
\caption{Time window $tw2$}
\label{vrp:step4tw2}
\end{subfigure}
\begin{subfigure}[b]{0.24\textwidth}
\centering
\includegraphics[width=\linewidth]{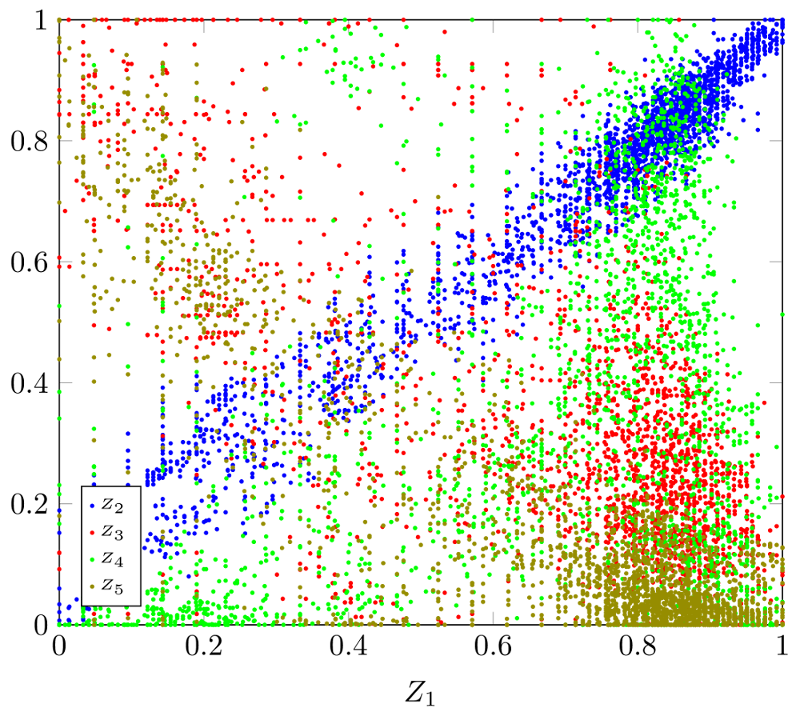}
\caption{Time window $tw3$}
\label{vrp:step4tw3}
\end{subfigure}
\begin{subfigure}[b]{0.24\textwidth}
\centering
\includegraphics[width=\linewidth]{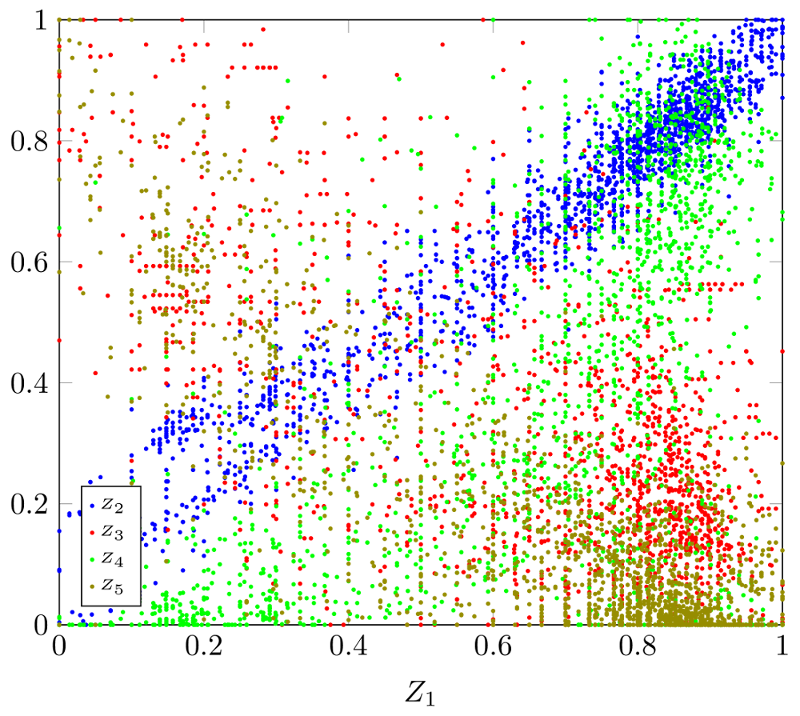}
\caption{Time window $tw4$}
\label{vrp:step4tw4}
\end{subfigure}
\begin{subfigure}[b]{0.4\textwidth}
\centering
\includegraphics[width=\linewidth]{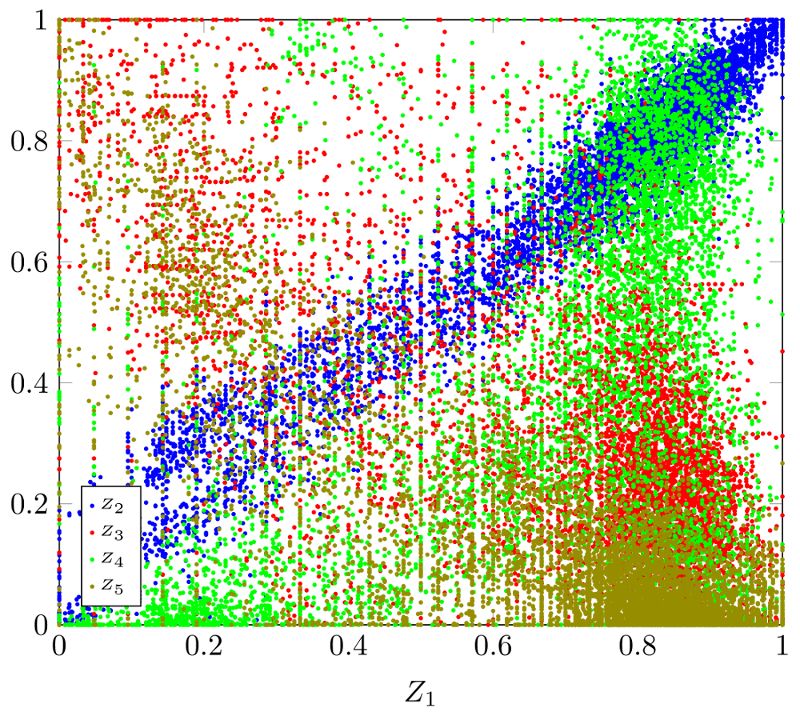}
\caption{Overall scatter-plot.}
\label{vrp:step4o}
\end{subfigure}
\caption{\hl{Scatter plots of all MOVRPTW sets where the objective $Z_1$  is represented on the $x$ axis and the remaining objectives as data points.}}
\label{vrp:step4}
\end{figure}

\textbf{Step 4 - Multiobjective Scatter Plot Analysis.}
Figure \ref{vrp:step4} presents the scatter plots of all sets and the combined overall scatter plot. The first relevant information shown is that the number of solutions for $tw0$ is noticeably smaller than for the other time windows. 
This happens because each customer in $tw0$ must be served in a single time window consisting of the entire day. Hence, there is no waiting time (because there is never an early arrival) and there will rarely be delays. Therefore, taking out two objectives drastically reduces the number of non-dominated solutions. 

Additionally, the fitness landscapes of all configurations are alike, meaning that varying the delta and time windows does not impose scenarios with distinct multiobjective natures. 
%Rodrigo: what does "impose scenarios with distinct multiobjective natures." mean?
Finally, the global pairwise harmonious relationship between $Z_1$ and $Z_2$ is evident in all graphs.

\subsubsection{Discussion}

The analysis performed on the benchmark instances of the MOVRPTW \citep{Castro-Gutierrez2011} corroborates their claim of a challenging multiobjective dataset. The global pairwise relationship analysis clearly showed fairly strong relationships while the range analysis presented high variance in all objectives, meaning that it is not possible to find good solutions in all areas. The region maps presented many areas where no solutions were found, hence the decision-maker must pick the regions of interest. Finally, the scatter plots showed similarities between the setups and where the global relationships stand.

The parametrisation proposed by the authors of those instances was observed to have little to no impact on the fitness landscape, meaning that as a benchmark for the problem it fails to provide diverse scenarios. 
An algorithm that can successfully explore $\delta0$ will most likely find no difficulties in exploring $\delta1$ or $\delta2$ due to the similarities of the objective space. Just like with the generational sets of the MONSP, the parameters used to generate the scenarios (delta and time windows) were not sufficient to provide instances with distinct fitness landscapes.
% Rodrigo: do we actually have references to show that if the Pareto fronts are similar, algorithms which perform well on one will most likely have no difficulties on others? If so then could add cite to support the argument here.

When comparing these scenarios with the previous problems, the fitness landscape resembles the MONSP and the independent set of the MOMKP, with no clear local relationships and a high density of solutions in a region: in this case when $Z_1 \gtrapprox 0.7$. However, the global relationships can be clearly observed, especially the harmonious $Z_1-Z_2$ and $Z_3-Z_5$ and the conflicting $Z_1-Z_3$ and $Z_1-Z_5$. 
These scenarios should represent challenging MOVRPTW instances because of the global relationships, high ranges and regions with no solutions.
% Rodrigo: if you have evidence of this then you can change should back to definitely. I am still concerned that the structure is as important in some cases as the Pareto front.

It is useful to note that other visualisation techniques might provide additional information. In particular, heatmaps \citep{Korhonen2008,Walker2013} might prove to be a useful tool and the inclusion of heatmaps into the proposed analysis technique is a subject that will be investigated in future work.
These were not considered here since the tools in hand were already able to provide sufficient information and the added complexity may be detrimental to the clarity of the analysis. For example the colour scheme of the heatmaps could cause confusion with the colour scheme used in the scatter-plots.

\section{Conclusion}
\label{sConclusion}

When working with real-world problems, many instances will often exhibit similar structures and characteristics.
Even where structures differ between instances, it is common to have only a few different structures.
Similarly, for many benchmark problems, the structure will often be similar within a set of instances which were created in the same way, unless measures were actively taken to avoid this.
For this reason, understanding the structure of some of the instances from a particular problem can give insight into the structure of other instances. Similarly, solution methods which work well for some instances of a problem can often work well for others, where the structure is similar. On the other hand, where the structure is different, an alternative solution method may be preferable, or at least an adaptation to handle that structural change. For example, some problems may be better suited for alternative techniques such as decomposition or weighted sum approaches. However assessing the nature of a MOP is itself, to make this determination, is a non-trivial task.

This paper presented a method for analysing and visualising the structure of the solutions for many-objective problems, providing the researcher with additional tools to aid the understanding of the structure and trade-offs in a set of solutions which approximate the Pareto optimal front. It has been identified that a specific issue is the locality of relationships between solutions, such that these may not hold globally. But identifying the fact that such relationships are present in specific areas of the objective space can be valuable for the development of solution methods that could find solutions that are acceptable to the practitioners. Indeed, it may be the case that a solution method which attempts to find a more complete approximation to the Pareto optimal front may benefit from a design which takes into account the fact that objectives may be positively or negatively correlated in different areas of the objective space, perhaps even changing its objective function accordingly.

%mean of the It is beneficial to assess an optimisation problem before deciding how to tackle it. Even though a problem may be multiobjective by definition, specific scenarios may be better suited for alternative techniques such as decomposition or weighted sum. Nonetheless, assessing the nature of a MOP is itself a non-trivial task. Common analysis techniques found in the literature does not consider complex relationship between objectives, furthermore, they were primarily conceived for two or three objectives, hence when applied to many-objective problems the information they provide may not be sufficient.

% JA got here.... (but having dinner now)
The proposed technique has four steps. The first evaluates the global correlation values to identify global relationships. The second identifies the most meaningful objectives. The third step uses trade-off region maps to highlight composite relationships. Finally, the fourth step uses scatter plots to identify local relationships. 

%In this work, we presented tools for the analysis and visualisation of objectives' relationships in MOPs. The proposed technique consists of fours steps. The first evaluates the global correlation values to identify relationships that happens throughout the fitness landscape, the second compute the ranges of values for all objectives -- to separate objectives that are meaningful from objectives that are not. The third step computes the distribution and frequency of solutions to draw trade-off region maps that highlight composite relationships. Finally, the fourth step consists of drawing scatter plots to identify local relationships. This technique is aimed to help assess MOPs, particularly combinatorial MOPs, where the fitness landscape is irregular and difficult to assess beforehand.

The proposed technique was applied to sets of different multiobjective problems:
\begin{enumerate}
\item Firstly, five sets of instances of a multiobjective multidimensional knapsack problem were considered and it was shown that the proposed technique can provide accurate information about different problem specifications. 
A misleading scenario ($X$) was considered, where correlation values pointed towards independence, but the proposed analysis technique identified composite and local relationships. 
%Also, we conjectured on how the new information can be used to design improved solutions. % needed?
\item Secondly, the well known NSPLib \citep{maenhout2005} was used to generate multiobjective scenarios for the nurse scheduling problem. This analysis technique showed that the generational patterns were not sufficient to create diversity in the fitness landscapes and to generate interesting multiobjective traits. However, the analysis identified the difference between fitness landscapes when the minimum number of consecutive days constraint was set to a more restrictive value.
\item Finally the technique was applied to a well known dataset for a multiobjective vehicle routing problem with time windows \citep{Castro-Gutierrez2011}. This analysis technique corroborated the authors' claim of interesting multiobjective scenarios due to the high pairwise correlation values. Nonetheless, the analysis technique also showed that the different parameters used to generate the instances (time windows and $\delta$ setups) were not sufficient to provide instances with different fitness landscapes.
\end{enumerate}

In addition to a demonstration of the capabilities of the proposed analysis technique, this paper gave insights into the generation of MOP benchmark scenarios. It was seen that constraint setup variance, commonly used in the literature, has limited impact on the fitness landscapes, hence relying solely on that may not be sufficient to generate varied scenarios. 
On the other hand, using data dependency to generate the instances was able to achieve higher diversity and obtain truly unique fitness landscapes, as was shown with the MOMKP. 

Future work will include the tailoring of algorithmic components based on information obtained from the analysis technique and its assessment. \hl{Moreover}, the use of components of the analysis technique, such as the region maps, during the optimisation process to help guide the search towards regions of interest, may also be considered. Finally, the investigation of the use of heatmaps to assess whether using these in addition to the current methods could provide \hl{additional} information on the multiobjective nature of the problem in hand, is also being considered as \hl{further} future work.

\section*{References}
\bibliographystyle{plainnat}
\bibliography{references}

\iffalse
%FIGURES AND TABLES SHOULD BE AT THE END
\newpage
\input{figures/complexrelationship.tex}
\clearpage
\input{figures/conflictingobj}
\clearpage
\input{figures/regionmap}
\clearpage
\input{figures/mokp/step1}%table1
\clearpage
\input{figures/mokp/step2}%table2
\clearpage
\input{figures/mokp/step3a}%table2
\clearpage
\input{figures/mokp/step3b}%table2
\clearpage
\input{figures/mokp/step3c}%table2
\clearpage
\input{figures/mokp/step3d}%table2
\clearpage
\input{figures/mokp/step3e}%table2
\clearpage
\input{figures/step4}%table4
\clearpage
\input{figures/nsp/scenarios}%table4
\clearpage
\input{figures/nsp/step1}%table4
\clearpage
\input{figures/nsp/step2}%table4
\clearpage
\input{figures/nsp/step3a}%table4
\clearpage
\input{figures/nsp/step3b}%table4
\clearpage
\input{figures/nsp/step3c}%table4
\clearpage
\input{figures/nsp/step4}%SLOW, WONT COMPILE IN SHARELATEX
\clearpage
\input{figures/vns/step1}%table4
\clearpage
\input{figures/vns/step2}%table4
\clearpage
\input{figures/vns/step3a}%table4
\clearpage
\input{figures/vns/step3b}%table4
\clearpage
%\input{figures/vns/step4}%SLOW, WONT COMPILE IN SHARELATEX
\clearpage
\fi
\end{document}